\documentclass[preprint,12pt]{elsarticle} 

\usepackage{amssymb}

\usepackage{amsmath,amsfonts,bm}

\def\eqref#1{equation~\ref{#1}}

\def\1{\bm{1}}

\def\ra{{\textnormal{a}}}

\def\rs{{\textnormal{s}}}

\def\rva{{\mathbf{a}}}

\def\va{{\bm{a}}}

\DeclareMathAlphabet{\mathsfit}{\encodingdefault}{\sfdefault}{m}{sl}
\SetMathAlphabet{\mathsfit}{bold}{\encodingdefault}{\sfdefault}{bx}{n}

\newcommand{\E}{\mathbb{E}}

\newcommand{\R}{\mathbb{R}}

\makeatletter
\newcommand{\removelatexerror}{\let\@latex@error\@gobble}
\makeatother

\usepackage{hyperref}
\usepackage{url}
\usepackage{subcaption}
\usepackage{amssymb}
\usepackage{amsmath}
\usepackage{amsthm,amsmath,amssymb}
\usepackage{mathrsfs}

\usepackage[flushleft]{threeparttable}
\usepackage{threeparttable}
\usepackage{booktabs}
\usepackage{comment}
\usepackage{amsthm}
\usepackage{amsmath}
\usepackage{apxproof}
\usepackage{thmtools, thm-restate}
\usepackage{wrapfig}
\usepackage{adjustbox}
\usepackage[flushleft]{threeparttable}
\usepackage{graphicx}

\usepackage{dashrule}

\usepackage[noamssymbols,upint]{newtxmath} 

\usepackage[ruled,linesnumbered]{algorithm2e}
\usepackage{algorithmic}
\usepackage{float}
\usepackage{ulem}
\usepackage{cancel}

\declaretheorem{theorem}

\declaretheorem{assumption}

\DeclareMathAlphabet\mathbfcal{OMS}{cmsy}{b}{n}
\newenvironment{smalleralign}[1][\small]
 {\par\nopagebreak\leavevmode\vspace*{-\baselineskip}%
  \skip0=\abovedisplayskip
  #1%
  \def\maketag@@@##1{\hbox{\m@th\normalfont\normalsize##1}}%
  \abovedisplayskip=\skip0
  \align}
 {\endalign\ignorespacesafterend}

\usepackage{color, xcolor}
\usepackage{soul}
\soulregister{\cite}7
\soulregister{\ref}7
\soulregister{\url}7
\soulregister{\footnote}7
\renewcommand{\hl}[1]{#1}

\usepackage{microtype}
\usepackage{graphicx}

\usepackage{booktabs} 

\usepackage{hyperref}
\usepackage{amssymb}
\usepackage{amsmath}
\usepackage{amsthm}

\usepackage{adjustbox}

\usepackage{cancel}

\usepackage[inline]{enumitem}

\usepackage{bm}
\usepackage{makecell}
\usepackage[figuresright]{rotating}

\newcommand{\policy}{\pi_{{\bm\theta}}}

\newcommand{\pis}{\Pi_{\mathtt{S}}}

\newcommand{\calA}{\mathcal{A}}

\newcommand{\calC}{\mathcal{C}}

\newcommand{\calO}{\mathcal{O}}
\newcommand{\calS}{\mathcal{S}}
\newcommand{\calM}{\mathcal{M}}

\newcommand{\bb}{\mathbf{b}}
\newcommand{\bc}{\mathbf{c}}

\usepackage{adjustbox}

\usepackage{enumitem}

\usepackage{graphicx}
\usepackage{array}

\usepackage{extarrows}
\usepackage{caption}
\usepackage{graphicx}
\usepackage{sidecap}
\usepackage{float}
\usepackage{framed}
\usepackage{tabularx}
\usepackage{twoopt}
\usepackage{adjustbox}

\usepackage{lscape}

\usepackage{float}

\usepackage{tikz}
\usetikzlibrary{fit}

\usepackage{booktabs,threeparttable}

\usepackage{lineno}

\newcommandtwoopt\Textbox[5][7.2cm][2cm]{%
\begin{tikzpicture}[remember picture,overlay]
  \coordinate (aux) at ([xshift=#1]#4);
  \node[inner ysep=3pt,yshift=1ex,draw=pink,thick,
    fit=(#3) (aux),baseline] 
    (box) {};
  \node[text width=#2,anchor=north east,
    font=\sffamily\footnotesize,
  align=right
    ] 
    at (box.north east) {#5};
\end{tikzpicture}%
}

\hypersetup{
colorlinks=true,
citecolor=red,
    linkcolor=red,
    filecolor=magenta,      
    urlcolor=red,
linktocpage}

\usepackage{tablefootnote}

\usepackage{lipsum}


\begin{document}

\begin{frontmatter}



\title{A Review of Safe Reinforcement Learning: Methods, Theories and Applications}




\author{Shangding Gu$^{a*}$, Long Yang$^{b}$,  Yali Du$^c$, Guang Chen$^d$, Florian Walter$^a$, Jun Wang$^e$, Alois Knoll$^a$}



\cortext[cor1]{This manuscript is under actively development. We appreciate any constructive comments and
suggestions corresponding to   \textit{shangding.gu@tum.de}.} 



\affiliation{organization={Department of Computer Science, Technical University of Munich},
            country={Germany}}
            
 \affiliation{organization={Institute
for AI, Peking University \& BIGAI},
            country={China}}
            
 \affiliation{organization={Department of Informatics, King's College London},
            country={UK}}
            
\affiliation{organization={College of Electronic and Information Engineering, Tongji University},
            country={China}}
            
 \affiliation{organization={Department of Computer Science, University College London},
            country={UK}}

\begin{abstract}

Reinforcement Learning (RL) has achieved tremendous success in many complex decision-making tasks. However, \hl{safety concerns are raised during deploying RL in real-world applications}, leading to a growing demand for safe RL algorithms, such as in autonomous driving and robotics scenarios. 
 While safe control has a long history, the study of safe RL algorithms is still in the early stages. 
To establish a good foundation for future safe RL research, in this paper, we provide a review of safe RL from the perspectives of methods, \hl{theories}, and applications. 
Firstly, we review the progress of safe RL from five dimensions and come up with five crucial problems for safe RL being deployed in real-world applications, coined as \textbf{``2H3W"}. Secondly,  we analyze the algorithm and theory progress from the perspectives of answering the \textbf{``2H3W"} problems. \hl{Particularly, the sample complexity of safe RL algorithms is reviewed and discussed}, followed by an introduction to the applications and benchmarks of safe RL algorithms. 
Finally, we open the discussion of the challenging problems in safe RL, hoping to inspire future research on this thread. 
To advance the study of safe RL algorithms, we release an open-sourced repository containing the implementations of major safe RL algorithms at the link\footnote{\scriptsize \url{https://github.com/chauncygu/Safe-Reinforcement-Learning-Baselines.git}}.

\end{abstract}


\begin{highlights}
\item A review of safe Reinforcement Learning (RL) methods is provided with theoretical and application analyses.
\item The key question that safe RL needs to answer is proposed, and five problems \textbf{``2H3W"} are analyzed to address the key question.
\item To examine the effectiveness of safe RL methods, several safe single-agent and multi-agent RL benchmarks are investigated.  
\item The challenging problems are pointed out to guide the research directions.
\end{highlights}

\begin{keyword}
  safe reinforcement learning; safety optimisation; constrained Markov decision processes; safety problems 



\end{keyword}

\end{frontmatter}




\section{Introduction}
Over the past decades, Reinforcement Learning (RL) has been widely adopted in many fields, e.g. transportation schedule~\citep{basso2022dynamic, chinchali2018cellular, li2019constrained, singh2020hierarchical}, traffic signal control~\citep{casas2017deep, chu2019multi}, energy management~\citep{mason2019review}, wireless security~\citep{lu2022safe}, satellite docking~\citep{dunlap2022run}, edge computing~\citep{xiao2020reinforcement}, chemical processes~\citep{savage2021model}, video games~\citep{berner2019dota, lee2018modular, mnih2015human, peng2017multiagent, sun2018tstarbots,  xu2019macro}, board games of Go, shogi, chess and arcade game PAC-MAN~\citep{jansen2020safe, silver2016mastering, silver2018general, silver2017mastering}, finance~\citep{abe2010optimizing, krokhmal2002portfolio, tamar2012policy,   tsitsiklis1999optimal}, autonomous driving~\citep{robotics11040081,  isele2018safe,  kendall2019learning, ma2021reinforcement, mirchevska2018high, sallab2017deep}, recommender systems~\citep{chow2017risk,   shani2005mdp,zhao2021dear},  resource allocation~\citep{junges2016safety, liu2021resource, mastronarde2011fast}, communication and networks~\citep{bovopoulos1992effect, boyan1993packet,hou2017optimization,  liu2020constrained}, smart grids~\citep{koutsopoulos2011control, schneider1999distributed}, video compression~\citep{xiao2021uav, mandhane2022muzero},  and robotics~\citep{AchiamHTA17,andrychowicz2020learning,bogert2018multi,brunke2021safe,   gu2021multi, gu2017deep,hu2020voronoi, kober2013reinforcement,leottau2018decentralized, miljkovic2013neural,ono2015chance,peng2018deepmimic, pham2018optlayer,     richter2019open}, etc. However, a challenging problem in this domain is: \textbf{how do we guarantee safety when we deploy RL in real-world applications?} After all, unacceptable catastrophes may arise if we fail to take safety into account during RL applications in real-world scenarios. For example, it must not hurt humans when robots interact with humans in human-machine interaction environments; false or racially discriminating information should not be recommended for people in recommender systems; safety has to be ensured when self-driving cars are carrying out tasks in real-world environments. More specifically, we introduce several safety definitions from different perspectives, which might be helpful for safe RL research.

\textbf{Safety definition.} The first type of safety definition: according to the definition of Oxford dictionary~\citep{stevenson2010oxford},  the phrase ``safety" is commonly interpreted to mean ``the condition of being protected from or unlikely to cause danger, risk, or injury."
The second type of safety definition: the definition of general ``safety" according to wiki~\footnote{\scriptsize \url{https://en.wikipedia.org/wiki/Safety}}, the state of being ``safe" is defined as ``being protected from harm or other dangers"; ``controlling recognized dangers to attain an acceptable level of risk" is also referred to as ``safety".  The third type of safety definition: according to Hans \textit{et al.}~\citep{hans2008safe}, humans need to label environmental states as ``safe" or ``unsafe," and agents are considered ``safe" if ``they never reach unsafe states". The fourth type of safety definition: agents are considered to be ``safe" by some research~\citep{hadfield2016cooperative, irving2018ai, leike2018scalable} if ``they act, reason, and generalize obeying human desires". The fifth type of safety definitions: Moldovan and  Abbeel~\citep{moldovan2012safe} consider an agent ``safe" if ``it meets an ergodicity requirement: it can reach each state it visits from any other state it visits, allowing for reversible errors". In this review, based on the above various definitions, we investigate safe RL methods, which are about optimizing cost objectives, avoiding adversary attacks, improving undesirable situations, reducing risk, and controlling agents to be safe, etc.

RL safety is a significant practical problem that confronts us in RL applications, and is one of the critical problems in AI safety~\citep{amodei2016concrete} that remains unsolved, though it has attracted increasing attention in the field of RL. Moreover, it is deduced that the mean minus variance~\citep{mannor2011mean} and percentile optimisation~\citep{delage2007percentile} of safe RL are, in general, NP-hard problems~\citep{moldovan2012risk}. In some applications, the agent's safety is much more important than the agent's reward~\citep{dulac2019challenges}. An attempt to answer the question above raises  some fundamental problems that we call {\textbf{``2H3W''}} problems:

\begin{itemize}
\item [(1)] 
\textbf{Safety Policy.} 
{\textbf{H}ow can we perform policy optimisation to search for a safe policy? }

\item [(2)] 
\textbf{Safety Complexity.} {\textbf{H}ow much training data is required to find a safe policy? }


\item [(3)] 

\textbf{Safety Applications.} {\textbf{W}hat is the up to date progress of safe RL applications?}

\item [(4)] 

\textbf{Safety Benchmarks.} \textbf{W}hat benchmarks can we use to fairly and holistically  examine safe RL performance?


\item [(5)] 
\textbf{Safety Challenges.} {\textbf{W}hat are the challenges faced in future safe RL research?}

\end{itemize}

\begin{figure*}[!htb]
 \centering
 {
 \includegraphics[width=0.97\linewidth]{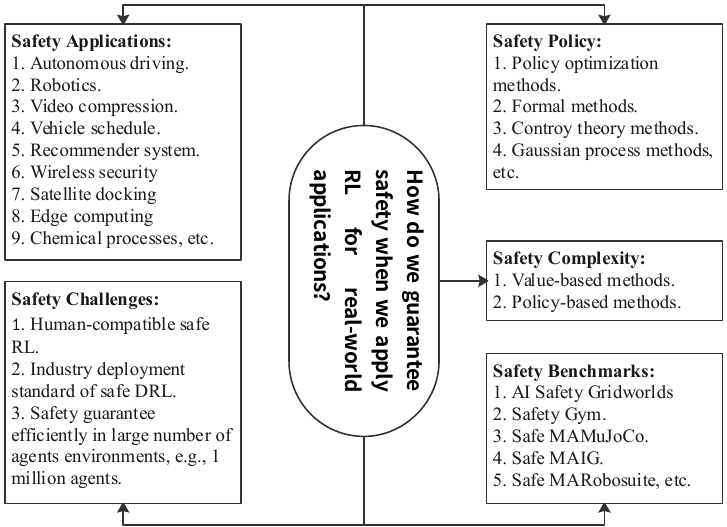}
}
    \vspace{-0pt}
 	\caption{\normalsize The framework of safe RL about \textbf{``2H3W"} problems.
 	} 
  \label{fig:over-review-introduction}
 \end{figure*}

Most of the research in this field is aimed at solving the above \textbf{``2H3W"} problems, and the framework of safe RL about \textbf{``2H3W"} problems is shown in Figure~\ref{fig:over-review-introduction}. 

As for the problem (1) (\textbf{safety policy}), in many practical applications, a robot must not visit some states, and
must not take some actions, which can be thought of as ``unsafe” either for itself or for elements of its environment. It is essential that a safe policy function or value function is provided so that agents can reach safe states or perform safe actions. To achieve safety policy, a growing number of approaches have been developed over the last few decades, such as primal-dual policy optimization methods \citep{chen2021primal,le2019batch,liang2018accelerated,paternain2019constrained,russel2020robust,satija2020constrained,tessler2019reward,xu2020primal}, trust region policy optimization with safety constraints methods \cite{AchiamHTA17,bharadhwaj2021conservative,ijcai2020-632,hanreinforcementl2020,vuong2019supervised,yang2020projection}, formal methods \citep{anderson2020neurosymbolic, bastani2018verifiable, beard2022safety, fulton2018safe, fulton2019verifiably, hunt2021verifiably, mirchevska2018high, pathak2018verification, riley2021reinforcement, tran2019safety, verma2019imitation, wang2021formal, zhu2019inductive}, control theory-based methods~\citep{choi2020reinforcement,chow2018lyapunov, chow2019lyapunov, dong2020principled, huh2020safe, jeddi2021lyapunov,kumar2017fuzzy, perkins2001lyapunov, perkins2002lyapunov}, Gaussian processes methods~\citep{akametalu2014reachability,cai2021safe,cowen2022samba,fan2019safety,   polymenakos2019safe,savage2021model, sui2015safe,  turchetta2016safe, wachi2018safe}.

 As for problem (2) (\textbf{safety complexity}), agents need to interact with environments and sample trajectories, such that the safe RL algorithms converge to below the constraint bound, and guarantee application safety. Since one of the natures of RL is exploration learning~\citep{sutton2018reinforcement},  it is usually hard to control the balance between exploitation and exploration, especially when we need to improve reward performance while satisfying cost constraints (cost is one way of encoding safety). Thus, we need to determine how many sample trajectories can make safe RL algorithms converge and satisfy safety bounds using the analysis of safety complexity during safe RL applications. Furthermore, the sample complexity of each safe RL algorithm is investigated from the viewpoints of value-based \cite{ding2021provably, he2021nearly,Kalagarla_Jain_Nuzzo_2021,  Sobhanreward-free2021}, and policy-based~\citep{ding2020natural, xu2021crpo, zeng2021finite} methods, etc.

As for problem (3) (\textbf{safety applications}), although there are many RL applications to date, most of the applications are merely simulations that do not take safety into account; some real-world experiments have been carried out, but there is still a long way to go before RL can be used in real-world applications. Generally, when we use safe RL in real-world applications, we need to consider the ego agent safety, environmental safety, and human safety. Most importantly, we need to consider the control safety, which prevents adversary attacks from destroying or controlling the agent~\citep{hendrycks2021unsolved}. Therefore, for the safe RL application research, we introduce some safe RL applications in this review, e.g., autonomous driving~\citep{robotics11040081,isele2018safe,  krasowski2020safe,  kalweit2020deep,kendall2019learning, mirchevska2017reinforcement, mirchevska2018high, wang2022ensuring}, robotics~\citep{  garcia2012safe, garcia2020teaching,perkins2002lyapunov, pham2018optlayer,slack2022safer, thomas2021safe}, video compression~\citep{mandhane2022muzero}, vehicle schedule~\citep{basso2022dynamic, li2019constrained}, etc.

As for problem (4) (\textbf{safety benchmarks}),  we need to determine how to design cost and reward functions considering the balance between RL reward performance and safety in each benchmark, since cost functions will typically disturb reward performance. If we have a loose cost function, we may not be able to guarantee agent safety during the learning process; if we take too conservative cost functions, for example, in a constrained policy optimization process, when we set the cost constraint bound as zero or a negative value, which may result in lousy reward performance. Thus, we should pay more attention to designing the cost and reward function in the benchmarks. In some safe RL benchmarks, such as AI Safety Gridworlds~\citep{leike2017ai}, Safety Gym~\citep{ray2019benchmarking}, Safe MAMuJoCo~\citep{gu2021multi}, Safe MAIG~\footnote{\scriptsize \url{https://github.com/chauncygu/Safe-Multi-Agent-Isaac-Gym.git}}, Safe MARobosuite~\citep{gu2021multi},  the cost and reward functions are tailored to specific tasks well in examining experiments.

As for problem (5) (\textbf{safety challenges}), firstly, when we consider RL safety, it is a significant challenge about how to consider human safety factors or environmental safety factors during deploying RL in real-world applications. 
Secondly, a further important aspect when considering RL safety is how to take robot safety factors into account during RL applications~\citep{amodei2016concrete, hendrycks2021unsolved}. Another critical challenge is the social dilemma problem with safety balance. For example, the game of the trolley problem~\cite{thomson1976killing}. In the game, an out-of-control trolley will eventually kill five people if no action is taken, but you can redirect the trolley to another track, where only one person will be killed. The open question is how to balance safety wights when using RL. Moreover, the application standard and safe Multi-Agent RL (MARL) should be considered for future research. 

The problem (5) (\textbf{safety challenges}) appear to be dilemma problems. They are not more straightforward problems compared to problem (1) (\textbf{safety policy}), problem (2) (\textbf{safety complexity}), problem (3) (\textbf{safety applications}),  and the problem (4)(\textbf{safety benchmarks}). We must guarantee agent, human, and environmental safety when we apply RL for practical applications by providing sophisticated algorithms. In addition, few studies have focused on problem (2) (\textbf{safety complexity}) and problem (4) (\textbf{safety benchmarks}), especially in industrial use. Answering problem (3) (\textbf{safety applications}) and problem (5) (\textbf{safety challenges}) may reveal RL application situations and provide a clue for future RL research. In this review, we will summarise the progress of safe RL, answer the five problems and analyze safe RL algorithms, theory, and applications. In general, we mainly sort out the safe RL research of the past two decades (Though we are unfortunately unable to include some impressive safe RL literature in this review for space reasons).


The main contributions of this paper: first, we investigate safe RL research and give an indication of the research progress. Second, the main practical question of RL applications is discussed, and five fundamental problems are analyzed in detail. Third, algorithms, theory, and applications of safe RL are reviewed in detail, e.g., safe model-based learning and safe model-free learning, in which we present a bird's eye view to summarising the progress of safe RL. Finally, the challenges we face when using RL for applications are explained.


The remainder of this paper is organized as follows: safe RL background, including safe RL problem formulation and related safe RL surveys, are introduced in Section \ref{section:review-background}; an overview of safe RL is provided in Section \ref{section:overview-safe-DRL}; safe RL theory and algorithm analysis are presented in Section \ref{section:theory-safe-DRL}; the application analysis of safe RL is introduced in Section \ref{section:Applications-safe-DRL}; several safe RL benchmarks are introduced in detail in Section \ref{section:Benchmarks-safe-DRL}; challenging problems and remaining questions are described in Section \ref{section:Challenges-safe-DRL}; a conclusion for the review is given in Section \ref{section:Conclusion-safe-DRL}.


\section{Background}
\label{section:review-background}

Safe reinforcement learning is often modeled as a Constrained Markov Decision Process (CMDP) \citep{altman1999constrained}, in which we need to maximize the agent reward while making agents satisfy safety constraints. A substantial body of literature has studied Constrained Markov Decision Process (CMDP) problems for both tabular and linear cases~\citep{altman1999constrained, beutler1985optimal, beutler1986time,book1983-Kallenberg, ross1989randomized, ross1989markov, ross1991multichain}. However, deep safe RL for high dimensional and continuous CMDP optimization problems is a relatively new area that has emerged in recent years, and proximal optimal values generally represent safe states or actions using neural networks. In this section, we illustrate the generally deep safe RL problem formulation concerning the objective functions of safe RL and offer an introduction to safe RL surveys.



\subsection{Problem Formulation of Safe Reinforcement Learning }

A CMDP problem \citep{altman1999constrained} is an extension of a standard Markov decision process (MDP) $\calM$ with a constraint set $\calC$. A tuple $\mathcal{M}=(\mathcal{S},\mathcal{A},\mathbb{P},{r},\rho_0,\gamma)$ is given to present a MDP  \citep{puterman2014markov}.
A state set is denoted as $\mathcal{S}$ ,  an action set is denoted as $\mathcal{A}$, 
$\mathbb{P}(s^{'}|s,a)$ denotes the probability of state transition from $s$ to $s^{\prime}$ after playing $a$.
A reward function is denoted as $r:\mathcal{S}\times\mathcal{A}\times\mathcal{S}\rightarrow \R$.
$\rho_{0}(\cdot):\mathcal{S}\rightarrow[0,1]$ is the starting state distribution, $\gamma$ denotes the discount factor.



In safe RL, the goal of an optimal policy $\pi$ is to maximize the reward and minimize the cost by selecting an action $a$, and $\pis$ is the policy set, $\tau$ is a trajectory, $\tau=\left(s_{0}, a_{0}, s_{1}, \ldots\right)$, in which an action depends on $\pi$, $s_{0}\sim\rho_{0}(\cdot)$, $a_{t}\sim\pi(\cdot|s_t)$, $s_{t+1}\sim \mathbb{P}(\cdot|s_{t}, a_{t})$.
$d_{\pi}^{s_0}(s)=(1-\gamma)\sum_{t=0}^{\infty}\gamma^{t}\mathbb{P}_{\pi}(s_t=s|s_0)$ denotes the state distribution (starting at $s_0$), the discounted state distribution based on the initial distribution $\rho_0 (\cdot)$ is present as $
d_{\pi}^{\rho_0}(s)=\mathbb{E}_{s_0\sim\rho_{0}(\cdot)}[d_{\pi}^{s_0}(s)]
$.


The state value function is shown in Function~(\ref{eq:state-value-function}), the state action value function is shown in Function~(\ref{eq:state-action-value-function}); the advantage function is shown in Function~(\ref{eq:advantage-function}); the reward objective is shown in Function~(\ref{J-objectiove}). 

\begin{flalign}
  \label{eq:state-value-function}
  V_{\pi}(s) = \mathbb{E}_{\pi}[\sum_{t=0}^{\infty}\gamma^{t}r_{t+1}|s_{0} = s].
\end{flalign}

\begin{flalign}
  \label{eq:state-action-value-function}
  Q_{\pi}(s,a) = \mathbb{E}_{\pi}[\sum_{t=0}^{\infty}\gamma^{t}r_{t+1}|s_{0} = s,a_{0}=a].
\end{flalign}

\begin{flalign}
  \label{eq:advantage-function}
  A_{\pi}(s,a)=Q_{\pi}(s,a) -V_{\pi}(s).
\end{flalign}

\begin{flalign}
  \label{J-objectiove}
  J(\pi)=\E_{s\sim \rho_0(\cdot)}[V_{\pi}(s)].
\end{flalign}


CMDP is based on MDP, in which the additional constraint set $\calC=\{(c_i,b_i)\}_{i=1}^{m}$ is considered,
where $c_i$ is the cost value functions, and $b_i$ is the safety constraint bound, $i \in[m]$, $m$ is the type number of cost constraints. 
Similarly, we have the cost value functions $V^{c_i}_{\pi}$, cost action-value functions $Q^{c_i}_{\pi}$, and cost advantage functions $A^{c_i}_{\pi}$, e.g., $V^{c_i}_{\pi}(s) = \mathbb{E}_{\pi}\left[\sum_{t=0}^{\infty}\gamma^{t}c_i(s_{t},a_{t})|s_{0} = s\right]$.
Based on the above definition, we have the expected cost function $C_{i}(\pi)=\E_{s\sim {\rho_0}(\cdot)}[V^{c_i}_{\pi}(s)]$.
The feasible policy set $\Pi_{\calC}$ that can satisfy the safety constraint bound  is as follows,
\[
\Pi_{\calC}=
\cap_{i=1}^{m}
\left\{
\pi\in\pis~~\text{and}~~C_{i}(\pi) \leq b_i
\right\}.
\]
The goal of safe RL  is to maximise the reward performance and minimise the cost values to guarantee safety,
\begin{flalign}
\label{def:problem-setting}
\max_{\pi\in\pis} J(\pi),~\text{such}~\text{that}~\bc(\pi)\preceq \bb,
\end{flalign}
where the vector $\bc(\pi)=\left(C_1(\pi),C_2(\pi),\cdots,C_m(\pi)\right)^{\top},$
and $\bb=\left(b_1,b_2,\cdots,b_m\right)^{\top}.$
Therefore, the solution of a CMDP problem (\ref{def:problem-setting}) can be provided as the optimization function~(\ref{basic-problem}):
\begin{flalign}
\label{basic-problem}
\pi_{\star}=\arg\max_{\pi\in\Pi_{\calC}} J(\pi).
\end{flalign}

\subsection{A Survey of Safe Reinforcement Learning related surveys}

Several surveys have already investigated safe RL problems and methods, e.g.,~\citep{brunke2021safe,garcia2015comprehensive,kim2020safe,liu2021policy}. However, the surveys do not provide comprehensive theoretical analysis for safe RL, such as sample complexity, nor do they focus on the critical problems of safe RL that we point out, \textbf{``2H3W"} problems. For example, \citep{brunke2021safe} investigates safe RL from the perspective of control theory and robotics; the soft constraints, probabilistic constraints, and hard constraints are defined in the the survey. Furthermore, they reviewed a large number of papers that are related to control theory and analyzed how to use control theory to guarantee RL safety and stability, such as using model predictive control~\citep{aswani2013provably,kamthe2018data,koller2018learning,  mayne2005robust, ostafew2016robust,rosolia2017learning,  soloperto2018learning}, adaptive control~\citep{gahlawat2020l1,hovakimyan2011l1, nguyen2011model,  sastry1990adaptive}, robust control~\citep{berkenkamp2015safe,  doyle1996robust, holicki2020controller, von2021probabilistic}, Lyapunov functions~\citep{chow2018lyapunov, chow2019lyapunov} for RL stability and safety. In the survey~\citep{garcia2015comprehensive}, they focus more on reviewing safe RL methods up to 2015. They categorize safe RL methods into two types: one is based on the safety of optimization criterion, where the worst case, risk-sensitive criterion, constrained criterion, etc., are taken into account to ensure safety. Another one is based on external knowledge or risk metric. In general, the external knowledge or risk metric is leveraged to guide the optimisation of RL safety. In contrast to the survey~\citep{garcia2015comprehensive}, the survey~{\citep{kim2020safe}} focuses more on the techniques of safe learning, including MDP and non-MDP methods, such as RL, active learning, evolutionary learning. The survey~\citep{liu2021policy} summarises safe model-free RL methods based on two kinds of constraints, namely cumulative constraints and instantaneous constraints. 

As for safe RL methods of cumulative constraints, three types of cumulative constraints are introduced:\\
 a discounted cumulative constraint:
\begin{equation}
J_{C_{i}}^{\pi_{\theta}}=\mathbb{E}_{\tau \sim \pi_{\theta}}\left[\sum_{t=0}^{\infty} \gamma^{t} C_{i}\left(s_{t}, a_{t}, s_{t+1}\right)\right] \leq b_{i},
\end{equation}
a mean valued constraint:
\begin{equation}
J_{C_{i}}^{\pi_{\theta}}=\mathbb{E}_{\tau \sim \pi_{\theta}}\left[\frac{1}{T} \sum_{t=0}^{T-1} C_{i}\left(s_{t}, a_{t}, s_{t+1}\right)\right] \leq b_{i},
\end{equation}
a probabilistic constraint:
\begin{equation}
J_{C_{i}}^{\pi_{\theta}}=P\left(\sum_{t} C_{i}\left(s_{t}, a_{t}, s_{t+1}\right) \leq b_{i} \right) \geq \eta.
\end{equation}

When it comes to instantaneous constraints, the instantaneous explicit constraints and instantaneous implicit constraints are given. The explicit ones have an accurate, closed-form expression that can be numerically checked, e.g., the cost that an agent generates during each step; the implicit does not have an accurate closed-form expression, e.g., the probability that an agent will crash into unsafe areas during each step. Although based on our investigation, most CMDP methods are based on cumulative cost optimization, a few CMDP methods focus on the immediate costs to optimize performance~\citep{regan2009regret}, and it is natural to take the cost of a whole trajectory rather than a state or action in some real-world applications, such as robot-motion planning and resource allocation~\citep{chow2018lyapunov}.
Compared to the related surveys~\citep{brunke2021safe,garcia2015comprehensive,kim2020safe,liu2021policy}, our survey pays more attention to answering the \textbf{``2H3W"} problems and provides safe RL algorithm analysis, sampling complexity analysis and convergence investigation from the perspectives of model-based and model-free RL.

\section{Methods of Safe Reinforcement Learning}
\label{section:overview-safe-DRL}

There are several types of safe RL methods based on different criteria. For example, \hl{from the perspective of objective optimization, several methods consider cost as one of the optimization objectives to achieve safety, e.g.,~\cite{basu2008learning,borkar2002q,heger1994consideration,howard1972risk,kadota2006discounted,  lutjens2019safe, nilim2005robust,    sato2001td, tamar2012policy}. From the perspective of knowledge utilization, some methods consider safety in the RL exploration process by leveraging external knowledge, e.g., }\cite{abbeel2010autonomous,chow2018lyapunov,clouse1992teaching,geramifard2013intelligent,moldovan2012safe,  tang2010parameterized,   thomaz2006reinforcement}. From the perspectives of policy and value-based methods for safe RL, summarise as follows, \hl{policy-based safe RL:}
\cite{AchiamHTA17, khattar2022cmdp, ding2020natural, gu2021multi,xu2021crpo,    yang2020projection, zeng2021finite}, 
{value-based safe RL:}
\cite{bai2021achieving,ding2021provably, efroni2020exploration,he2021nearly, Kalagarla_Jain_Nuzzo_2021,   liu2021learning, Sobhanreward-free2021, wei2021provably}. From the perspective of the agent number, we have safe RL methods in a safe single-agent RL setting and a safe multi-agent RL setting. More specifically,  numerous safe RL methods are about the single-agent setting. An agent needs to explore the environments to improve its reward while keeping costs below the  constraint bounds. In contrast to safe single-agent RL methods, safe multi-agent RL methods not only need to consider the ego agent’s reward and other agent's reward, but also have to take into account the ego agent’s safety and other agents’ safety in an unstable multi-agent system.

In this section, we provide a concise but holistic overview of safe RL methods from a bird's eye view and attempt to answer the \textbf{``Safety Policy"} problem. Especially, we will introduce safe RL methods from the perspectives of model-based methods and model-free methods, in which safe single-agent RL and multi-agent RL will be analyzed in detail. \hl{In particular, we will introduce policy optimization-based approaches, formal methods-based approaches, control theory-based approaches, and Gaussian processes-based approaches in model-based and model-free settings, respectively.}

In addition, the model-based and model-free safe RL analysis are summarised in Table~\ref{table:Model-based-safe-DRL-analysis} and Table~\ref{table:Model-free-safe-DRL-analysis} respectively.

\subsection{Methods of Model-Based Safe Reinforcement Learning}

Although accurate models are challenging to build and many applications lack models, model-based Deep RL (DRL) methods usually have a better learning efficiency than model-free DRL methods. There are still many scenarios for which we can apply model-based DRL methods, such as robotics, transportation planning, logistics, etc. Several works have shown that safe problems,  such as a safe robot control problem~\cite{tessler2018reward}, can be overcome by using model-based safe RL methods. \hl{In this section, we will investigate model-based safe RL methods from different perspectives: policy optimization-based approaches, control theory-based approaches, formal methods-based approaches, and Gaussian process-based approaches.}

\subsubsection{\textcolor{black}{Policy optimization-based approaches}} \hl{Policy optimization-based approaches usually search for a safe policy using cumulative cost values on trajectories.} For example, Moldovan and Abbeel use the Chernoff function~(\ref{eq:model-based-chernoff-risk-function}) in \cite{moldovan2012risk} to achieve near-optimal bounds with desirable theoretical properties. \hl{The cumulative expectation cost is represented in the function as $E_{s, \pi}\left[e^{J / \theta}\right]$,} especially for $\delta$, which can be used to adjust the balance of reward performance and safety. For instance, if $\delta$ is set as 1, this method will ignore the safety,  and if $\delta$ is set as 0, this method will fully consider the risk and optimize the cost. Moreover, they examine the method in grid world environments and air travel planning applications. However, the method needs a significant amount of time to recover policy from risk areas, which is ten times the value iteration. 
\begin{equation}
C_{s, \pi}^{\delta}[J]=\inf _{\theta>0}\left(\theta \log E_{s, \pi}\left[e^{J / \theta}\right]-\theta \log (\delta)\right)
\label{eq:model-based-chernoff-risk-function}
\end{equation}

Different from the Chernoff function based methods~\cite{moldovan2012risk}, Borkar \cite{borkar2005actor} proposes an actor-critic RL method to handle a CMDP problem based on the envelope theorem in mathematical economics~\cite{mas1995microeconomic, milgrom2002envelope}, in which the primal-dual optimization is analyzed in detail using a three-time scale process. The critic scheme, actor scheme, and dual ascent are on the fast, middle, and slow timescale. Bharadhwaj \textit{et al.} \cite{bharadhwaj2021conservative} also present an actor-critic method to address a safe RL problem, where they first develop a conservative safety critic to estimate the safety. The primal-dual gradient descent is leveraged to optimize the reward and cost value by constraining the failure probability. Although this method can bound the probability of failures during policy evaluation and improvement, this method still cannot guarantee total safety; a few unsafe corner actions may dramatically damage critical robot applications.


Akin to Borkar's method \cite{borkar2005actor},  Tessler \textit{et al.} \cite{tessler2019reward} also utilize a multi-timescale approach with regards to cost as a part of the reward in primal-dual methods. However, the method's learning rate is hard to tune for real-world applications because of imposing stringent requirements, and the method may not guarantee safety when agents are training.

\hl{Similar to the above methods, in actor-critic settings, with primal-dual optimization techniques,} Yu \textit{et al.}\cite{yu2019convergent} convert a non-convex constrained problem into a locally convex problem and guarantee the stationary point to the optimal point of non-convex optimization problem; they consider the state-action safety optimization, and the optimization process is motivated by~\cite{liu2019stochastic}, in which the Lipschitz condition is necessary to satisfy the results. Also, they need to estimate the policy gradient for optimization.

Analogously, using optimization theory,  the policy gradient and actor-critic methods are proposed by Chow \textit{et al.}~\cite{chow2017risk} to optimize risk RL performance, in which  CVaR\footnote{CVaR denotes the Conditional Value at Risk.}-constrained and chance-constrained optimization are used to guarantee safety. Specifically, the importance sampling~\cite{bardou2009computing, tamar2014policy} is used to improve policy estimation and provide convergence proof for the proposed algorithms. Nevertheless, this method may not guarantee safety during training~\cite{zhang2020first}. Paternain \textit{et al.} \cite{paternain2019constrained} provide a duality theory for CMDP optimization, and they prove the zero duality gap in primal-dual optimization even for non-convexity problems. Furthermore, they point out that the primal problem can be exactly solved by dual optimization. In their study, the suboptimal bound using neural network parametrization policy is also present \cite{hornik1989multilayer}.


\subsubsection{\textcolor{black}{Control theory-based approaches}} \hl{Control theory-based approaches mostly ensure more rigorous safety than policy optimization approaches. Nevertheless, control theory-based approaches may not be easy to transfer to other domains due to the strict requirements of dynamics models. For instance,} Berkenkamp \textit{et al.} \cite{berkenkamp2017safe} develop a safe model-based RL algorithm by leveraging Lyapunov functions to guarantee stability with the assumptions of Gaussian process prior; their method can ensure a high-probability safe policy for an agent in a continuous process. However, Lyapunov functions are usually hand-crafted, and it is not easy to find a principle to construct Lyapunov functions for an agent's safety and performance~\cite{chow2018lyapunov}. In addition, some safe RL methods are proposed from the perspective of Model Predictive Control (MPC)~\cite{zanon2020safe}, e.g., MPC is used to make robust decisions in CMDPs~\cite{aswani2013provably} by leveraging a constrained MPC method~\cite{mayne2000constrained}, which also introduces a general safety framework to make decisions~\cite{fisac2018general}. 

\hl{Apart from  Lyapunov functions and MPC-safe RL approaches, control barrier functions (CBFs) based approaches are also proposed to ensure learning safety~\cite{ma2021model, choi2020reinforcement, emam2022safe, cohen2023safe}. Similar to the above control methods, CBFs based safe RL approaches also require a dynamics model for the control system. Thus, it is not straightforward to deploy it in RL~\cite{marvi2021safe}. For example, \cite{ma2021model} propose a model-based safe RL by leveraging lagrangian optimization and CBFs; they further deploy their method in simulation and real-world autonomous driving experiments, the experiment results indicate that their method can improve sample efficiency and ensure safety. In contrast to \cite{ma2021model}, \cite{emam2022safe} not just use CBFs, but also robust CBFs~\cite{emam2019robust}, they develop a safe RL method based on SAC~\cite{sac} and robust CBFs. Specifically, they consider safety in robust settings by estimating the disturbed dynamics systems with GP models. \cite{cohen2023safe} propose a safe exploration framework based on model-based RL and CBFs, where Lyapunov-like CBFs~\cite{panagou2015distributed} are used to construct the safety CBFs, and they can ensure the learning safety and stability. The above methods show impressive performance in ensuring safety. However, designing CBFs necessitates knowledge of the model or safety certificates, which may be challenging in complex environments, even when the model is derived using alternative methods.}

\subsubsection{\textcolor{black}{Formal methods-based approaches}}
 \hl{Different from policy optimization-based approaches, control theory-based approaches, and Gaussian Process-based methods,} formal methods~\cite{anderson2020neurosymbolic} usually try to ensure safety without unsafe probabilities. However, most formal methods rely heavily on the model knowledge and may not show better reward performance than other methods. The verification computation might be expensive for each neural network~\cite {anderson2020neurosymbolic}. More generally, the curse of dimensionality problem is challenging to be solved, which appears when formal methods are deployed for RL safety~\cite{beard2022safety}, since formal methods may be intractable to verify RL safety in continuous and high-dimension space settings \cite{beard2022safety}.

 For instance, in \cite{anderson2020neurosymbolic}, Anderson \textit{et al.} provide a neurosymbolic RL method by leveraging formal verification to guarantee RL safety in a  mirror descent framework. In their method, the neurosymbolic and constrained classes using symbolic policies are leveraged to approximate gradient and conduct verification in a loop-iteration setting. The experiment results show promising performance in RL safety, though the fixed worst-case model knowledge is used in these environments, which may not be suitable for practical applications. Similarly, in \cite{beard2022safety},  Beard and  Baheri utilize formal methods to improve agent safety by incorporating external knowledge and penalizing behaviors for RL exploration. Nonetheless, the methods may need to be developed for scalability and continuous systems. Finally, in \cite{fulton2018safe},  Fulton and  Platzer also use external knowledge to ensure agent safety by leveraging the justified speculative control sandbox in offline formal verification settings.

 \subsubsection{\textcolor{black}{Gaussian processes-based approaches}}
In formal methods, determining how to measure unsafe areas is a challenging problem. Recently, many approaches have been proposed by leveraging a Gaussian Process (GP)~\cite{williams2006gaussian} to estimate the uncertainty and unsafe areas. Further, the information from GP methods is incorporated into the learning process for agent safety. For instance,  Akametalu \textit{et al.} \cite{akametalu2014reachability}~ develop a safe RL method based on reachability analysis, in which they use GP methods to measure the disturbances which may lead to unsafe states for agents, the maximal safe areas are computed iteratively in an unknown dynamics system. Like Akametalu \textit{et al.} \cite{akametalu2014reachability} method,  Berkenkamp and  Schoellig \cite{berkenkamp2015safe} utilize GP methods to measure the system uncertainty and further guarantee system stability.  Polymenakos \textit{et al.} \cite{polymenakos2019safe}develop a safe policy search approach based on PILCO (Probabilistic Inference
for Learning Control) method~\cite{deisenroth2011pilco} which is a policy gradient method derived from a GP, in which they improve agent safety using probability trajectory predictions by incorporating cost into reward functions. Similar to Polymenakos \textit{et al.} \cite{polymenakos2019safe}, Cowen \textit{et al.} \cite{cowen2022samba} also use PILCO to actively explore environments while considering risk, a GP is used to quantify the uncertainty during exploration, and agent safety probability is improved by leveraging a policy multi-gradient solver. 


 The above safe RL methods present excellent performance in terms of the balance between reward and safety performance in most challenging tasks. Nonetheless, training safety or stability may need to be further investigated rigorously, and a unified framework may need to be proposed to better examine safe RL performance. 

\begin{table}[htbp!]
\renewcommand\arraystretch{0.5}
	\centering
	\scriptsize{
	\begin{adjustbox}{width=0.99\textwidth,left}
	\renewcommand{\arraystretch}{3}
	{\begin{tabular}{p{2cm} p{5cm} p{4cm} p{2cm}}
			\hline
			Model-Based Safe RL &  Features                      &           Methods &   Convergence Analysis\\
			\hline
			{\cite[ Borkar (2005)]{borkar2005actor}} & {Based on the Lagrange
multiplier formulation  and the saddle point property.}         & An actor-critic algorithm for CMDP. & YES.\\
			
			 \cite[Chow et al., (2018)]{chow2017risk}
			 &  {Two risk-constrained MDP problems. The first one involves a CVaR constraint
and the second one involves a chance constraint.}        & {Policy gradient and actor-critic algorithms via CVaR.} & YES.  \\
			 \cite[Tessler et al., (2019)]{tessler2019reward}&  {A novel constrained actor-critic approach RCPO\footnote{ RCPO denotes the Reward Constrained Policy Optimization.} uses a multi-timescale approach.}    & Reward constrained policy optimization.  & YES. \\
			 \cite[Yu et al., (2019)]{yu2019convergent}&  {Successive convex relaxation algorithm for CMDP. }          &  Actor-Critic update for constrained MDP. & YES. \\
			 
			 \cite[Koppel et al., (2019)]{koppel2019projected}&  {Saddle points with  a reproducing
Kernel Hilbert Space.}        & {Projected Stochastic Primal-Dual Method for CMDP.  }& YES.  \\
			 
			 \cite[ Polymenakos et al., (2019)]{polymenakos2019safe}&  {Ensure safety during training with high probability based on a PILCO framework.}        & {Policy gradient with a GP model.  }& NO.  \\
			 \cite[Paternain et al., (2019).]{paternain2019constrained}&  {Solving non-convex CMDP.}        & {Primal-dual algorithms
for constrained reinforcement learning.} & YES.  \\
			 \cite[Bharadhwaj et al., (2021)]{bharadhwaj2021conservative} &  {Without strong assumptions, e.g., unsafe state knowledge is required a prior, the method can avoid unsafe actions during training with high probability.}     & Conservative critics for safety exploration. & YES. \\
			 \cite[Miryoosefi and Jin (2021)]{Sobhanreward-free2021}& {CMDP with general convex constraints.}   &  {Reward-free approach to constrained reinforcement learning.} & NO.  \\
			 \cite[Paternain et al., (2022)]{paternain2019safe}&  {Safety into the problem as a probabilistic version
of positive invariance.}          &  Stochastic Primal-Dual for Safe Policies. & NO. \\
			 \hline
			
	\end{tabular}}
	\end{adjustbox}	  
	}
	\caption{Model-based safe RL analysis. 
	}
	\label{table:Model-based-safe-DRL-analysis}
\end{table}

\subsection{Methods of Model-Free Safe Reinforcement Learning}

\hl{Most studies pay attention to model-free safe RL since it can be deployed in many domains without requiring model dynamics. In this section, we also investigate safe RL methods from the perspectives of policy optimization, control theory, Gaussian processes, and formal methods.}

\subsubsection{\textcolor{black}{Policy optimization-based approaches}}

Constrained Policy Optimisation (CPO)\cite{AchiamHTA17} is the first policy gradient method to solve the CMDP problem based on model-free  deep RL. In particular, a policy has to be optimized to guarantee the reward of a  monotonic improvement while satisfying safety constraints. As a result, their methods can almost converge to safety bound and produce more comparable performance than the primal-dual method~\cite{chow2017risk} on some tasks. However, CPO's computation is more expensive than PPO\footnote{PPO denotes the Policy Proximal Optimization.}-Lagrangian, since it needs to compute the Fisher information matrix and uses the second Taylor expansion to optimize objectives. Moreover, the approximation and sampling errors may have detrimental effects on the overall performance, and the convergence analysis is challenging. Furthermore, the additional recovery policy may require more samples, which could result in wasted samples~\cite{zhang2020first}.

Derived from CPO~\cite{AchiamHTA17}, Projection-based Constrained Policy Optimisation (PCPO) \cite{yang2020projection} based on two-step methods constructs a cost projection to optimize cost and guarantee safety, which displays better performance than CPO on some tasks. PCPO leverages policy to maximize the reward via Trust Region Policy Optimization (TRPO)  method~\cite{schulman2015trust}, and then projects the policy to a feasible region to satisfy safety constraints. However, the second-order proximal optimization is used in both steps, which may result in a more expensive computation than the First Order Constrained
Optimization in Policy Space (FOCOPS) method~\cite{zhang2020first}, which only uses the first-order optimization. \cite{zhang2020first} is motivated by the optimization-based idea~\cite{peters2010relative}, where they use the primal-dual method, address policy search in the nonparametric space and project the policy into the parameter space, to carry out proximal maximization optimization in CMDPs. Although this method is easy to implement and shows better sample efficiency, it still needs to solve the problems of unstable saddle points and unsafe actions during training.

Similar to CPO~\cite{AchiamHTA17}, in Safe Advantage-based  Intervention for Learning policies with
Reinforcement (SAILR) \cite{wagener2021safe}, Wagener \textit{et al.} leverage the advantage function as a surrogate to minimize cost, and further achieve safe policy both during training and deployment. Furthermore,
In \cite{sohn2021shortest}, a Shortest-Path Reinforcement Learning (SPRL) method is proposed using off-policy strategies to construct safe policy and reward policy, and its applications are used for the shortest path problems in Travel Sale Path (TSP). Besides, based on a Gaussian process of a safe RL method~\cite{wachi2018safe}, the SNO-MDP\footnote{ The SNO-MDP represents the Safe Near-Optimal MDP.} \cite{wachi2020safe} is developed to optimize cost in the safe region and optimize the reward in the unknown safety-constrained region~\cite{sui2015safe, turchetta2016safe}. It is suggested that the maximization reward is more important than safety. The policy is often substantial in some cases. For example, staying in the current position for safety is extremely conservative. This method~\cite{wachi2020safe} shows the near-optimal cumulative reward under some assumptions, whereas it cannot achieve the near-optimal values while guaranteeing safety constraints.

Different from CPO, a first-order policy optimization method~\cite{liu2020ipo} is provided based on interior point optimization (IPO)~\cite{boyd2004convex}, in which the logarithmic barrier functions are leveraged to satisfy safety constraints. The method is easy to implement by tuning the penalty function. Although the empirical results on MuJoCo~\cite{AchiamHTA17} and grid-world environments~\cite{chow2015risk} have demonstrated their method's effectiveness, the theoretical analysis to guarantee the performance is still needed to be provided.

Unlike the above CMDP optimization, a meta algorithm~\cite{Sobhanreward-free2021} is proposed to solve safe RL problems with general convex constraints~\cite{brantley2020constrained}. They can, in particular, solve CMDP problems with a 
small number of samples in reward-free settings. The algorithms can also be used to solve approachability problems~\cite{blackwell1956analog, miryoosefi2019reinforcement}. Although the theoretical results have shown their method's effectiveness, practical algorithms and experiments ought to be proposed and carried out to further evaluate their method. In~\cite{paternain2019safe}, an attempt is made to solve safe RL  using the trajectory probability in a safe set. \hl{The probability invariance is positive, and constraint gradients are obtained using the policy parameters.} More importantly, the related problem can have an arbitrarily small duality gap. However, this method may also encounter the stability problems of primal-dual methods, and the saddle points using Lagrangian methods may be unstable~\cite{panageas2019first}.

\hl{Comparable to the above primal-dual settings, Ghosh \textit{et al.}~\cite{ghosh2022provably} have developed a model of safe RL based on least-squares value iteration-upper confidence bound \cite{pmlrv125jin20a} in primal-dual settings, which is suitable for large-scale optimization, and the regret bound and safety violation are analyzed. While similar methodologies have been proposed, they are predominantly utilized within tabular settings, as exemplified by the work of Wei et al.~\cite{wei2021provably}. Furthermore, Xiong \textit{et al.}~\cite{xiong2024provably} introduce a step-wise safe RL method based on upper confidence bound value iteration \cite{azar2017minimax}. This method employs optimistic estimations of transition kernels and cost functions, thereby enhancing its capacity to ensure safety in reward-free scenarios. Notably, Xiong \textit{et al.} also provide analyses of safety violations and regret bounds, facilitating the extension of this method to additional contexts such as multi-robot systems. Although these methods demonstrate improved performance over related baselines, further exploration into their applications across diverse domains, such as safe multi-robot control and planning, could yield even more beneficial insights.} \textcolor{black}{In recent years, primal-based methods \citep{xu2021crpo} have emerged as powerful alternatives to primal-dual-based approaches \citep{AchiamHTA17,yang2022cup}, demonstrating remarkable performance in robotics applications. Notable examples include CRPO \citep{xu2021crpo} and PCRPO \citep{gu2024balance}. A significant advantage of primal-based methods is that they do not require tuning dual parameters and are not influenced by initialization \citep{xu2021crpo}. Specifically, CRPO introduced a primal-based framework for safe RL, but it may have experienced oscillations between reward and safety optimization, potentially undermining its performance. To address this issue, PCRPO introduced a soft policy optimization technique with gradient manipulation, exhibiting superior performance compared to CRPO and state-of-the-art primal-dual-based methods such as PCPO \citep{yang2020projection}, CUP \citep{yang2022cup}, and PPOLag \citep{Ray2019}.}

\subsubsection{\textcolor{black}{Control theory-based approaches}}
\hl{In contrast to safe policy optimization such as meta algorithms, CPO and IPO based on model-free RL, control theory is are also leveraged to ensure RL safety. For example, the first Lyapunov functions used for a safe RL may be~\cite{perkins2002lyapunov}, in which the agent's actions are constrained by applying the control law of Lyapunov functions to learning systems and removing the unsafe actions in the action set.} The experiments have demonstrated that their method can achieve safe actions for the control problems in their study. However, the method requires the knowledge of a Lyapunov function in advance. If the environment dynamics model is unknown, it may be difficult to address safe RL problems with this method. Unlike \cite{perkins2002lyapunov}, in \cite{chow2018lyapunov} and \cite{chow2019lyapunov}, Chow \textit{et al.} propose several safe RL methods based on Lyapunov functions in discrete and continuous CMDP settings, respectively, where Lyapunov functions are used for safe RL to guarantee safe policy and learning stability. The methods can guarantee safety during training, and the Lyapunov functions can be designed by a proposed Linear Programming algorithm. However, the training stability and safety using Lyapunov functions still need to be improved, and more efficient algorithms in the setting may need to be proposed.
 
\hl{Apart from Lyapunov functions based safe RL, CBFs are also developed in model-free settings in recent years \cite{yang2023model, vu2021barrier},  \cite{yang2023model} develop a model-free safe RL method based CBFs, in which they learn the policy and CBFs with data-driven methods, which can help to reduce reliance on the environment models. Similarly, \cite{vu2021barrier} proposes a safe RL method for power system control with CBFs, in which they integrated CBFs into reward functions to search for safe policy. model free control barrier function, \cite{marvi2020safe}, \cite{zhang2024constrained}. Moreover, \cite{cheng2019end} leverages a model-free RL  and CBFs to guarantee learning safety while improving learning efficiency, in which CBFs are used to constrain the search space to enhance learning safety and efficiency based on a GP learning model. Moreover, \cite{marvi2021safe} also leverages CBFs to ensure learning safety with learned dynamics knowledge,  where they provide convergence and stability analysis, and their method outperforms the baselines in their experiments. 
}

Safety layer methods have been proposed in recent years, which are more directly to ensure RL safety than safe policy optimization methods that based on cost value optimization such as CPO and IPO. \hl{We categorize safety-layer-based approaches into control theory methods since they can be considered as a control filter in control theory.} 

For example, Pham \textit{et al.} propose an OptLayer~\cite{pham2018optlayer} method, in which they leverage stochastic control policies to attain the reward performance, and a layer of the neural network is integrated to pursue safety during applications. The real-world applications also demonstrate the effectiveness of the safety layer. Qin \textit{et al.} \cite{qin2021density} propose the DCRL (Density Constrained Reinforcement Learning) method that is to optimize the reward and cost from the perspective of an optimization criterion, in which they consider the safety constraints via the duality property~\cite{dai2018boosting, nachum2019dualdice, nachum2020reinforcement, tang2019doubly} with regard to the state density functions, rather than the cost functions, reward functions and value functions~\cite{AchiamHTA17,  dalal2018safe,  ding2020natural}. This method lies in model-free settings whereas Chen \textit{et al.} \cite{chen2020density} provide similar methods in model-based settings. A-CRL (state Augmented Constrained Reinforcement Learning) method \cite{calvo2021state} is proposed to address a CMDP problem whereby the optimal policy may not be achieved via regular rewards. Their method focuses on solving the monitor problem in CMDP while the dual gradient descent is used to find the feasible trajectories and guarantee safety. Nonetheless, OptLayer~\cite{pham2018optlayer}, A-CRL \cite{calvo2021state} and
DCRL \cite{qin2021density} all lack convergence rate analysis.

 \subsubsection{\textcolor{black}{Gaussian processes-based approaches}}
Relative to safe policy optimization methods that attend more to cost values, Lyapunov function-based techniques, which emphasize safe actions, and Safety layer approaches, GP methodologies primarily concentrate on the safe exploration via modeling the safe states. 
In a GP of model-free settings, Sui \textit{et al.} \cite{sui2015safe} use a GP method to present the unknown reward function from noise samples, the exploration by leveraging a GP method is improved to reduce uncertainty and ensure agent safety. More particularly, the GP method is used to predict unknown function, and guide the exploration in bandit settings which do not need to state transitions. Their real-world experiments in movie recommendation systems and medical areas indicate that their method can achieve near-optimal values safely. Like Sui \textit{et al.} \cite{sui2015safe},  Turchetta \textit{et al.} \cite{turchetta2016safe} also leverage a GP method to approximate unknown functions prior for safe exploration. Nevertheless, they focus more on finite MDP settings considering explicitly reachability. Nonetheless, the method may not optimize reward objectives while considering safety.  Wachi \textit{et al.}~\cite{wachi2018safe} represent unknown reward and cost functions with GP methods to ensure safety with probability and optimize reward.  Furthermore, the safe, unsafe, and uncertain states are denoted for agent optimistic and pessimistic exploration, and their method can adapt the trade-off between exploration and exploitation. Nevertheless, the convergence guarantee with finite-time rates, optimization for multiple and heterogeneous objectives may need to be provided.

Although GP methods have shown impressive performance with regard to RL safety, most of them ensure safety with probability. How to rigorously guarantee RL safety during exploration still remains open.

\subsubsection{\textcolor{black}{Formal methods-based approaches}}

\hl{Distinct from employing GP methods to model the safe state, Hasanbeig et al. \cite{hasanbeig2020cautious} propose a safety-enhanced RL approach that leverages LTL. In this novel methodology, logical formulas serve as constraints during the exploration phase of policy synthesis. Although this method demonstrates notable safety performance, it is imperative to carefully define the logical constraints to ensure safe exploration and effectively balance the trade-off between safety performance and reward acquisition. Murugesan \textit{et al.}~\cite{murugesan2019formal} employ a formal tool, satisfiability modulo theories \cite{clarke2018handbook}, as a safety verification layer to enhance learning safety. Furthermore, Hasanbeig \textit{et al.}~\cite{hasanbeig2020towards} encode properties of LTL into the reward function through reward engineering using a limit deterministic Büchi automaton \cite{sickert2016limit} to ensure safety. While the aforementioned methods exhibit impressive performance in ensuring safety in most tasks, their deployment in real-world applications still needs to be further investigated.}

\begin{table}[tb!]
	\centering
	\scriptsize{
	\begin{adjustbox}{width=0.99\textwidth,left}
	\renewcommand{\arraystretch}{3}
	{\begin{tabular}{p{3cm} p{6cm} p{3cm} p{2cm}}
			\hline
			Model-Free Safe RL &  Features                      &           Methods &   Convergence Analysis \\
			\hline
			{\cite[ Miryoosefi and Jin (2021)]{Sobhanreward-free2021}} & {Provide a  meta algorithm to  solve CMDP  problems using  model free methods.}         & Reward-free methods for CMDP. & NO.\\
			 \cite[Liang et al., (2022)]{paternain2019safe}&  {Provide analysis to primal-dual gap can be arbitrarily small, and can converge for any small step-size $\eta_{\lambda}$.}    & {Primal-dual methods.} & YES, by~\cite{durrett2019probability}.   \\
			 \cite[Achiam et al., (2017)]{paternain2019safe}&  {Constrained policy optimisation.}    & {Trust region with safety constraints.} & NO.  \\
			 
			 \cite[Liu et al., (2020)]{paternain2019safe}&  {PPO with logarithmic barrier functions.}    & {Primal-dual methods.} & NO.  \\
			 \cite[Chow et al., (2018,2019)]{chow2018lyapunov,chow2019lyapunov}&  { An effective Linear Programming method to generate Lyapunov
functions, and the algorithms can ensure feasibility, and optimality for discrete and continuous system under certain
conditions.}    & {Lyapunov
function methods.} & NO.  \\
 \cite[Yang et al., (2020)]{yang2020projection}&  { Two-step optimisation, first reward, second safety.}    & {Trust region with safety constraints.} & YES.  \\
  \cite[Zhang et al.,(2020)]{zhang2020first}&  { First order optimisation with two-step optimisation, first  nonparameterized policy, second  parameterized policy.}    & {Primal-dual and Trust region methods.} & YES.  \\

			 \cite[Wagener et al., (2021).]{wagener2021safe}&  {Two-step optimisation with advantage. First, MDP optimisation Second, CMDP optimisation by chance optimisation.}        & {An intervention-based method.} & NO.  \\
			 \cite[Wachi et al.,(2020)]{wachi2020safe} &  {Optimse CMDP under unknow safety constraints. First, search safe policy by expanding safe region. Second, optimse reward in safe regions.}     & Gaussian process optimisation. & YES. \\
			 \cite[Pham et al., (2018)]{pham2018optlayer}& {Select safe action by a constrained neural network, and carry out experiments in simulation and real-world environments.}   &  {Trust region method with a neural network constraints.} & NO.  \\
			 \hline
			
	\end{tabular}}
	\end{adjustbox}	  
	}
	\caption{Model-free safe RL analysis. 
	}
	\label{table:Model-free-safe-DRL-analysis}
\end{table}




\subsection{Safe Multi-Agent Reinforcement Learning}

Safe RL has received increasing attention both from academia and industry. However, most current RL methods are based on the single-agent setting. Safe MARL is still a  relatively new area that has emerged in recent years. Little research has yet been carried out that considers the safe multi-agent RL, which can be seen as a multi-agent CMDP problem. Safe multi-agent RL not only needs to consider the ego agent's safety and other agents' safety, but also needs to take into account the ego agent's reward and other agents' reward. In this section, we briefly introduce the safe multi-agent RL problem formulation (since merely a few safe MARL methods are developed up to date, MARL problem formulation is not given in Section~\ref{section:review-background}), and some safe multi-agent RL methods are analyzed in detail.

\subsubsection{Problem Formulation of Safe Multi-Agent Reinforcement Learning }

In this section, We mainly investigate MARL in a fully cooperative setting, and it's can be seen as a multi-agent constrained Markov game considered as the tuple $\langle \mathcal{N}, \mathcal{S}, \boldsymbol{\mathcal{A}}, \mathrm{P}, \rho^0, \gamma, R, \boldsymbol{C}, \boldsymbol{c}\rangle$.  A set of agents is present as $\mathcal{N}= \{1, \dots, n\}$, $\mathcal{S}$ is the state space, $\boldsymbol{\mathcal{A}} = \prod_{i=1}^{n}\Pi^i$ is the product of  agents' action spaces, known as the joint action space, $P: \mathcal{S} \times \boldsymbol{\mathcal{A}} \times \mathcal{S} \rightarrow[0,1]$ is a function of probabilistic state transition, the initial state distribution is $\rho^0$, $\gamma \in(0,1)$ is a discount factor, $R:\mathcal{S}\times\boldsymbol{\mathcal{A}}\rightarrow\mathbb{R}$ is the joint reward function, $\boldsymbol{C} = \{C^i_j\}^{i\in\mathcal{N}}_{1 \leq j\leq m^i}$ is the set of cost functions {\small $C^i_j:\mathcal{S}\times\mathcal{A}^i\rightarrow \mathbb{R}$} of individual agents, and finally the set of cost-constraining values is given by $\boldsymbol{c} = \{c^i_j\}^{i\in\mathcal{N}}_{1 \leq j\leq m^i}$. At time step $t$, the agents are in state $\rs_t$, and every one of them performs an action according to its policy $\pi^i(\ra^i|\rs_t)$.  Together with other agents' actions, this results in a joint action $\rva_t = (\ra^1_t, \dots, \ra^n_t)$ drawn from the joint policy $\boldsymbol{\pi}(\rva|\rs) = \prod_{i=1}^{n}\pi^i(\ra^i|\rs)$. The agents receive the reward $R(\rs_t, \rva_t)$, and each agent $i$ pays the costs $C^i_j(\rs_t, \ra^i_t)$, $\forall j=1, \dots, m^i$. The whole system moves to the state $\rs_{t+1}$ with probability $\mathrm{P}(\rs_{t+1}|\rs_t, \rva_t)$. The agents' goal is to maximise the joint return, given by
\vspace{-10pt}
\begin{smalleralign}
J(\boldsymbol{\pi}) = \E_{\rs_0\sim\rho^0, \rva_{0:\infty}\sim\boldsymbol{\pi}, \rs_{1:\infty} \sim \boldsymbol{P}}\left[ \sum\limits_{t=0}^{\infty}\gamma^t R(\rs_t, \rva_t) \right], 
\end{smalleralign}
while obeying the safety constraints
\vspace{-10pt}
\begin{smalleralign}
J^i_j(\boldsymbol{\pi}) = 
\E_{\rs_0\sim\rho^0, \rva_{0:\infty}\sim\boldsymbol{\pi}, \rs_{1:\infty} \sim {\boldsymbol{P}}}\left[ \sum\limits_{t=0}^{\infty}\gamma^t C^i_j(\rs_t, \ra^i_t) \right].
\end{smalleralign}

The state-action value function, as well as the state-value function are defined as
\begin{smalleralign}
Q_{\boldsymbol{\pi}}(s, \va) = 
\E_{\rs_{1:\infty} \sim \mathrm{P}, \rva_{0:\infty}\sim\boldsymbol{\pi}}\left[ \sum\limits_{t=0}^{\infty}\gamma^t R(\rs_t, \rva_t) \Big| \rs_0 = s, \rva_0 = \va \right], 
\nonumber
\end{smalleralign}

\begin{smalleralign}
V_{\boldsymbol{\pi}}(s) = \E_{\rva\sim\boldsymbol{\pi}}\left[ Q_{\boldsymbol{\pi}}(s, \rva) \right],\nonumber
\end{smalleralign}
respectively. 

In multi-agent settings, the contribution of actions of subsets of the agent set via the multi-agent state-action value function and state value function is introduced as  $Q$-function and  $V$-function:
\begin{equation}
\begin{cases}
Q^{\pi^{i_{1:h}}, \boldsymbol{\pi^{-i_{h}}}}\left(s_{t}, a^{i_{1:h}}_{t}, \boldsymbol{a^{-i_{h}}_{t}}\right) = \mathbb{E}_{s_{t+1}, a^{i_{1:h}}_{t+1}, \boldsymbol{a^{-i_{h}}_{t+1}} }\left[\sum_{l=0}^{\infty} \gamma^{l} R\left(s_{t+l}, a^{i_{1:h}}_{t+l}, \boldsymbol{a^{-i_{h}}_{t+l}}\right)\right] \\
V^{\pi_{i_{1:h}}, \boldsymbol{\pi_{-i_{h}}}}\left(s_{t}\right)=\mathbb{E}_{a^{i_{1:h}}_{t}, \boldsymbol{a^{-i_{h}}_{t}}, s_{t+1} }\left[\sum_{l=0}^{\infty} \gamma^{l} R\left(s_{t+l}, a^{i_{1:h}}_{t+l}, \boldsymbol{a^{-i_{h}}_{t+l}}\right)\right] 
\end{cases}
\end{equation}
{The advantage for policy optimisation is given as}
\begin{equation}
A^{\pi^{i_{1:h}}, \boldsymbol{\pi^{-i_{h}}}}\left(s_{t}, a^{i_{1:h}}_{t}, \boldsymbol{a^{-i_{h}}_{t}}\right) = Q^{\pi^{i_{1:h}}, \boldsymbol{\pi^{-i_{h}}}}\left(s_{t}, a^{i_{1:h}}_{t}, \boldsymbol{a^{-i_{h}}_{t}}\right)-V^{\pi^{i_{1:h}}, \boldsymbol{\pi^{-i_{h}}}}\left(s_{t}\right)
\end{equation}


Similar to reward, cost for $j$ constraint of any agent $i$ in a multi-agent system:

\begin{equation}
\begin{cases}
Q_j^{\pi_{i_{1:h}}, \boldsymbol{\pi_{-i_{h}}}}\left(s_{t}, a^{i_{1:h}}_{t}, \boldsymbol{a^{-i_{h}}_{t}}\right) = \mathbb{E}_{s^{t+1}, a_{i_{1:h}}^{t+1}, a_{-i_{h}}^{t+1} }\left[\sum_{l=0}^{\infty} \gamma^{l} C^{i}_{j}\left(s_{t+l}, a^{i_{1:h}}_{t+l}, \boldsymbol{a^{-i_{h}}_{t+l}}\right)\right] \\
V_{j}^{\pi_{i_{1:h}}, \boldsymbol{\pi_{-i_{h}}}}\left(s_{t}\right)=\mathbb{E}_{a^{i_{1:h}}_{t}, a^{-i_{h}}_{t}, s_{t+1} }\left[\sum_{l=0}^{\infty} \gamma^{l} C^{i}_{j}\left(s_{t+l}, a^{i_{1:h}}_{t+l}, \boldsymbol{a^{-i_{h}}_{t+l}}\right)\right]
\end{cases}
\end{equation}

\begin{equation}
A_{j}^{\pi^{i_{1:h}}, \boldsymbol{\pi^{-i_{h}}}}\left(s_{t}, a^{i_{1:h}}_{t}, \boldsymbol{a^{-i_{h}}_{t}}\right) = Q_j^{\pi_{i_{1:h}}, \boldsymbol{\pi_{-i_{h}}}}\left(s_{t}, a^{i_{1:h}}_{t}, \boldsymbol{a^{-i_{h}}_{t}}\right) -V_{j}^{\pi^{i_{1:h}}, \boldsymbol{\pi^{-i_{h}}}}\left(s_{t}\right)
\end{equation}


The goal of a safe MARL algorithm under the feasible set constraint~(\ref{eq:feasible-set-of-goal-safe-marl}) is expressed as equation~(\ref{eq:safe-goal-marl}), the optimal policy~(\ref{eq:optimal-policy-of-safe-marl}) is that the policy can maximize the agents' reward and satisfy the constrained conditions,
\begin{equation}
\begin{aligned}
\label{eq:safe-goal-marl}
&\max ~  J\left(\pi^{i_{1:h}}, \boldsymbol{\pi^{-i_{h}}}\right), \\ \text { s.t. }  J_{j}^{i}\left(\pi^{i_{1:h}}, \boldsymbol{\pi^{-i_{h}}}\right) \leq c^{i}_{j} &  \text { for all }  1 \leq j \leq m,~ 1 \leq i \leq n, 1 \leq h \leq n.
\end{aligned}
\end{equation}

\begin{equation}
\label{eq:feasible-set-of-goal-safe-marl}
\Omega_{C}=\left\{\pi^{i_{1:h}}, \boldsymbol{\pi^{-i_{h}}}: \forall ~1 \leq j \leq m, 1 \leq i \leq n, 1 \leq h \leq n, J^{i}_{j}\left(\pi^{i_{1:h}}, \boldsymbol{\pi^{-i_{h}}}\right) \leq c^{i}_{j}\right\},
\end{equation}

\begin{equation}
\label{eq:optimal-policy-of-safe-marl}
(\pi^{-i_{1:h}}_{*},\boldsymbol{\pi^{-i_{h}}_{*}})=\underset{{\pi^{i_{1:h}}, \boldsymbol{\pi^{-i_{h}}}}\in\Omega_{C}} {\operatorname{argmax}} \{J\left(\pi^{i_{1:h}}, \boldsymbol{\pi^{-i_{h}}}\right)\} \quad \forall i.
\end{equation}

\subsubsection{Methods of Safe Multi-Agent Reinforcement Learning }

The Multi-Agent Constrained Policy Optimisation (MACPO) algorithm~\citep{gu2021multi} is the first safe model-free MARL algorithm which is developed based on CPO~\citep{AchiamHTA17} and Heterogeneous-Agent
Trust Region Policy Optimisation (HATRPO)~\citep{kuba2021trust}; it can guarantee monotonic improvement in reward, while theoretically satisfying safety constraints during each iteration. The algorithm was tested on several challenging tasks, e.g. safe multi-agent Ant tasks, safe multi-agent Lift tasks. However, MACPO is an expensive computation algorithm since  the Fisher Information Matrix is computed to approximate conjugate gradients. In contrast to MACPO, MAPPO-Lagrangian~{\citep{gu2021multi}} is the model-free and first-order safe MARL algorithm that requires less time for computation on most of challenging tasks, and is easily implemented. In addition, by adaptive updating of  Lagrangian multipliers, MAPPO-Lagrangian can avoid the problem of the sensitivity to the initialization of
the Lagrange multipliers, that other Lagrangian relaxation methods have.  Nonetheless, MAPPO-Lagrangian does not guarantee hard safety constraints.  \citep{gu2024safe} presents a soft-constrained policy optimization approach for real-world robot control, which addresses sim-to-real challenges. Despite these advancements, further investigation is required to establish robust theoretical guarantees.

Safe  Decentralized Policy Gradient (Dec-PG)~\citep{lu2021decentralized} using a decentralized policy descent-ascent method is the most closely-related algorithm to MACPO. The saddle point considering reward and cost is searched by a primal-dual framework in a multi-agent system. 
However, a consensus network used in this method imposes an extra constraint of parameter sharing among neighboring agents, which could yield suboptimal solutions \citep{kuba2021trust}. 
In contrast, MACPO does not require the assumption of parameter sharing.
Furthermore, multi-agent policy gradient methods can suffer from high variance by parameter sharing  \citep{ kuba2021settling}. 

Robust MARL~\citep{zhang2020robust} is a model-based MARL algorithm, which takes into account the reward uncertainty and transition uncertainty. In their experiments, they take into account the reward uncertainty by randomizing a nature policy with Gaussian noise, as a constraint to optimize the agent performance. Nevertheless, the payoff assumption for each agent to get a higher payoff when other agents do not satisfy the equilibrium policies, and the equilibrium assumption for global optimization, may be too strong in real-world applications.


{Another recently safe MARL work is CMIX~\citep{liu2021cmix}, which considers peak and average constraints using QMIX~\citep{rashid2018qmix}. Even though CMIX can satisfy the multi-agent cost constraint in these experiments, it does not provide theoretical analysis to guarantee safety for a multi-agent system. CMIX is also the parameter-sharing algorithm that suffers from problems similar to those of Safe Dec-PG \citep{lu2021decentralized}.}

Based on formal methods~\citep{woodcock2009formal}, Safe MARL via shielding is proposed\citep{elsayed2021safe}, in which they study the safe MARL problem by one of the formal methods, Linear Temporal Logic (LTL)~\citep{pnueli1977temporal} that is already adopted in safe single-agent settings~\citep{alshiekh2018safe}. Specifically, an attempt is made to ensure the safety of multi-agent systems by leveraging central shielding or factored shielding methods, which are used to specify the safe actions that agents only can take, and also correct the unsafe actions that agents explored. Furthermore, they evaluate the methods based on Multi-Agent Deep Deterministic Policy Gradient (MADDPG)~\citep{lowe2017multi} and Coordinating Q-learning (CQ-learning) methods~\citep{de2010learning}. Although this method can guarantee safety for multi-agent systems to some extent, it requires a prior LTL safety specification and problem-specific knowledge in advance. In a similar way that \citep{elsayed2021safe} ensures safety by optimizing the exploration, \citep{sheebaelhamd2021safe} developed a safe MARL algorithm by adding a safety layer based on safe DDPG~\citep{dalal2018safe} and MADDPG~\citep{lowe2017multi} for continuous action space settings. As a result, their method greatly enhanced safety on some MARL tasks. However, the method does not guarantee zero violation in all tasks that they carried out.

\subsection{Summary of Safe Reinforcement Learning Methods}
In this section, we introduce safe RL algorithms from different perspectives, such as value-based safe RL, policy-based safe RL; safe single-agent RL, safe multi-agent RL; model-based safe RL, model-free safe RL. Even though a number of algorithms display impressive performance in terms of reward scores, there is a long way to go toward real-world applications. In addition, based on our investigation, in MARL settings, \citep{lauer2000algorithm} is one of the earliest methods that present convergence guarantee for multi-agent systems in 2000. However, only a few algorithms consider safety constraints,  and safe MARL is still a relatively new area that requires a lot more attention.

\section{Theory of Safe Reinforcement Learning}
\label{section:theory-safe-DRL}


\hl{This section investigates theoretical techniques for analyzing safe RL. Firstly, we introduce the essential theoretical differences between standard RL and safe RL. Secondly, we investigate theories for primal-dual approaches and constrained policy optimization, two prominent methods in theoretical safe RL. Finally, we delve into safe RL sampling complexity and safety violation analysis, and discuss other theoretical frameworks for safe RL. This section aims to comprehensively investigate safe RL theories by clearly delineating the theoretical foundations and analytical techniques. This exploration demonstrates the critical differences from standard RL and highlights the theoretical underpinnings of various safe RL settings and their associated analyses.}

\subsection{Difference between RL and Safe RL}
  
\hl{
The standard RL exists a deterministic stationary policy \cite{puterman2014markov}, while CMDP may not exist uniformly optimal stationary policies  \cite{altman1999constrained}, which is one of the main differences between safe RL and standard RL. We present it as follows, and the blog \cite{szepesvari2020} presents a two-state example to illustrate this issue.}
{{
  \begin{theorem}
{If $\Pi_{\calC}\ne\varnothing$, then there exists a $\pi_{\star}\in\Pi$ is the optimal policy of CMDP, where the optimal policy $\pi_{\star}$ belongs to the following two cases:}
\begin{itemize}
\item \hl{$\pi_{\star}$ is a deterministic stationary policy;}
\item \hl{or $\pi_{\star}$ is a randomized stationary policy. Its randomness only occurs on the unique state $\tilde{s}\in\calS$, where the policy $\pi_{\star}$ randomly plays two actions on $\tilde{s}\in\calS$. For the state  $s\ne\tilde{s}$, the policy $\pi_{\star}$ is a deterministic stationary policy.}
\end{itemize}
\end{theorem}

{Additionally, the next theorem shows that checking a safe policy is time-expensive.}
  \begin{theorem}{\cite{feinberg2000constrained} Checking feasibility in CMDPs among deterministic policies is NP-hard.}
\end{theorem}
}}

\subsection{{\color{black}{Theory for Primal-Dual Approaches}}}

A standard way to solve a CMDP problem is the Lagrangian approach \cite{chow2017risk} that is also known as primal-dual optimization,
\begin{flalign}
\label{min-max-search}
\resizebox{.7\hsize}{!}{$
(\pi_{\star},\bm{\lambda}_{\star})=\arg\min_{\bm{\lambda} \succeq\bm{0}}\max_{\pi\in\pis}
\left\{
J(\policy)-\bm{\lambda}^{\top}(\bc(\pi)-\bb)
\right\}.
$}
\end{flalign}
Extensive canonical algorithms are proposed to solve problem (\ref{min-max-search}), e.g., \cite{chen2021primal,le2019batch, liang2018accelerated, paternain2019constrained, russel2020robust,satija2020constrained,tessler2019reward,xu2020primal,ding2024last}.


The work \cite{abad2003self} presents a policy-based algorithm to solve the CMDP problem with average cost finite states, which is 
a stochastic approximation algorithm. Furthermore, the work \cite{abad2003self} shows the locally optimal policy of the proposed algorithm. 
Chen et al. \cite{tessler2019reward} propose the Reward Constrained Policy Optimization (RCPO) that is a multi-timescale algorithm for safe RL, and \cite{tessler2019reward} show the asymptotical convergence of RCPO. The main idea of achieving such an asymptotical convergence is stochastic approximation \cite{borkar2009stochastic,robbins1951stochastic} that has been widely used in RL, e.g., \cite{pollard2000asymptopia,2005Stochastic,yang-ijcai2018-414}. 
{
\hl{ For the global non-asymptotic convergence guarantees, see \cite{borkar2005actor,bhatnagar2012online,vaswani2022near,abad2002self,chow2018risk,li2024faster,hasanzadezonuzy2021model}.
}}

Based on Dankin’s Theorem and Convex Analysis \cite{bertsekas2015convex}, \cite{paternain2019constrained} provides theoretical support to the primal-dual (i.e., the Lagrange multiplier) method with a zero duality gap, which implies that the primal-dual method can be solved precisely in the dual domain.
In \cite{le2019batch}, they consider a framework for off-line RL under some constraint conditions, 
 present a specific algorithmic instantiation, and show the performance guarantees that preserve safe learning. Moreover, due to the off-line RL, according to the off-policy policy evaluation, the work \cite{le2019batch} shows the error bound with respect to Probably Approximately Correct (PAC) style bounds \cite{1984Valiant}.
The work \cite{russel2020robust} is proposed to merge the theory of constrained CMDP with the theory of robust Markov decision process (RMDP).
The RMDP leads to a formulation of a robust constrained Markov decision process (RCMDP), which is the essential formulation to build a robust soft-constrained Lagrange-based algorithm.
The work \cite{russel2020robust} claims the convergence of the proposed algorithm can follow the method \cite{borkar2009stochastic} because the proposed algorithm can present a time-scale and stochastic approximation.
The work \cite{satija2020constrained} firstly turn the cost-based constraints into state-based constraints, then propose a policy improvement algorithm with a safety guarantee.
In \cite{satija2020constrained}, the authors propose backward value functions that play a role in estimating the expected cost according to the data by the agent,
their main idea is policy iteration \cite{santos2004convergence}.


\subsection {{\color{black}{Theory for Constrained Policy Optimization}}}

{\hl{
The work CPO \cite{AchiamHTA17} suggests to use a surrogate cost function which evaluates the constraint $J^{c}(\pi_{{\bm{\theta}}})$ according to the samples collected from the current policy $\pi_{{\bm{\theta}}_k}$.
}}
Although using a surrogate function to replace the cumulative constraint cost has appeared in \cite{dalal2018safe,el2016convex,gabor1998multi}, CPO \cite{AchiamHTA17} firstly show their algorithm guarantees for near-constraint satisfaction.
\begin{flalign}
\label{cpo-objective}
&\resizebox{.7\hsize}{!}{$\pi_{{\bm{\theta}}_{k+1}}=\arg\max_{\pi_{{\bm{\theta}}}\in\Pi_{{\bm{\theta}}}}~~~~\E_{s\sim d^{\rho_0}_{\pi_{{\bm{\theta}}_k}}(\cdot),a\sim\pi_{{\bm{\theta}}}(\cdot|s)}\left[A_{\pi_{{\bm{\theta}}_k}}(s,a)\right]$}\\
\label{cost-constraint}
&\resizebox{.7\hsize}{!}{$\text{s.t.}~~J^{c}(\pi_{{\bm{\theta}}_k})+\dfrac{1}{1-\gamma}\E_{s\sim d^{\rho_0}_{\pi_{{\bm{\theta}}_k}}(\cdot),a\sim\pi_{{\bm{\theta}}}(\cdot|s)}\left[A^{c}_{\pi_{{\bm{\theta}}_k}}(s,a)\right]\leq b,$}\\
\label{trust-region}
&\resizebox{.7\hsize}{!}{$\bar{D}_{\text{KL}}(\pi_{{\bm{\theta}}},\pi_{{\bm{\theta}}_k})=\E_{s\sim d^{\rho_0}_{\pi_{{\bm{\theta}}_k}}(\cdot)}[\text{KL}(\pi_{{\bm{\theta}}},\pi_{{\bm{\theta}}_k})[s]]\leq\delta.$}
\end{flalign}

Existing recent works (e.g., \cite{AchiamHTA17,bharadhwaj2021conservative,ijcai2020-632,hanreinforcementl2020,vuong2019supervised,yang2020projection}) try to find some convex approximations to replace the terms $A_{\pi_{{\bm{\theta}}_k}}(s,a)$ and $\bar{D}_{\text{KL}}(\pi_{{\bm{\theta}}},\pi_{{\bm{\theta}}_k})$.
Concretely, \cite{AchiamHTA17} suggest using the first-order Taylor expansion to replace (\ref{cpo-objective})-(\ref{cost-constraint}), the second-oder approximation to replace (\ref{trust-region}).
\hl{Such first-order and second-order approximations turn a non-convex problem (\ref{cpo-objective})-(\ref{trust-region}) into a convex problem. This would appear to be a simple solution, but this approach results in many error sources and troubles in practice.}
Firstly, it still lacks a theory analysis to show the difference between the non-convex problem (\ref{cpo-objective})-(\ref{trust-region}) and its convex approximation.
Policy optimization is a typical non-convex problem \cite{yang2021sample}; its convex approximation may introduce some errors for its original issue.
Secondly, CPO updates parameters according to conjugate gradient \cite{suli2003introduction}, and its solution involves the inverse Fisher information matrix,
which requires expensive computation for each update.
Later, the work \cite{yang2020projection} proposes PCPO that also uses second-order approximation, resulting in an expensive computation.

Conservative Safety Critics (CSC) \cite{bharadhwaj2021conservative} provides a \hl{new estimator for the safe exploration of RL}. CSC uses a conservative safety critic to estimate the environmental state functions. The likelihood of catastrophic failures has been bounded for each round of the training.
The work \cite{bharadhwaj2021conservative} also shows the trade-off between policy improvement and safety constraints for its method. During the training process, CSC keeps the safety constraints with a high probability, which is no worse asymptotically than standard RL.

First Order Constrained Optimization in Policy Space (FOCOPS) \cite{zhang2020first} and onservative update policy (CUP) methods \cite{yang2022cup} suggest that the \hl{non-convex implementation solving constrained policy optimization contains a two-step approach: policy improvement and projection.} The work \cite{zhang2020first} has shown the upper boundedness of the worst case for safety learning, where it trains the model via first-order optimization. CUP \cite{yang2022cup} provides a theoretical analysis extending the bound concerning the generalized advantage estimator (GAE) \cite{schulman2016high}.
\hl{GAE significantly reduces variance while achieving a tolerable level of bias, which is one of the critical steps when designing efficient algorithms} \cite{yang2019policy}. In addition, the asymptotic results that safe RL methods can achieve are summarised in Table~\ref{table:safe-rl-theory-Asymptotical Results}.

\begin{table}[H]
	\centering
	\scriptsize{
	\begin{adjustbox}{width=0.82\textwidth,center}
	\renewcommand{\arraystretch}{3}
	{\begin{tabular}{p{2cm} l l p{2cm}}
			\hline
			Reference &        Main  Technique              &           Method       &    Implementation \\
			 \hline
			  \cite{abad2003self}&  SA \cite{borkar2009stochastic} and Iterate Averaging  \cite{1995Analysis}    & Primal-Dual & Policy Gradient\\
			RCPO \cite{tessler2019reward}&  SA and ODE \cite{borkar2009stochastic}      & Primal-Dual & Actor-Critic \\
			  \cite{borkar2005actor}&  SA \cite{borkar2009stochastic} and Envelope Theorem \cite{milgrom2002envelope}    & Primal-Dual & Actor-Critic \\
        \cite{paternain2019constrained} &  Dankin’s Theorem and Convex Analysis \cite{bertsekas2015convex}   &Primal-Dual & Actor-Critic \\
			 \cite{le2019batch} &  FQI \cite{munos2008finite} and PAC \cite{1984Valiant} &Primal-Dual & Value-based \\
			 \cite{russel2020robust} &   SA \cite{borkar2009stochastic} &Primal-Dual & Policy Gradient \\
			 \cite{satija2020constrained} &   Policy Iteration \cite{santos2004convergence} &Primal-Dual & Policy Gradient \\
			CPO \cite{AchiamHTA17} &  Convex Analysis \cite{bertsekas2015convex} &CPO-based & Policy Gradient \\
			PCPO \cite{yang2020projection} &  Convex Analysis \cite{bertsekas2015convex} &CPO-based & Policy Gradient \\
			FOCOPS \cite{zhang2020first}&  Non-Convex Analysis &CPO-based & Actor-Critic \\
			CUP \cite{yang2022cup}&  Non-Convex Analysis &CPO-based & Actor-Critic \\
			CSC \cite{bharadhwaj2021conservative}&  On-line Learning \cite{agarwal2021theory}&CPO-based & Actor-Critic \\
			SAILR \cite{wagener2021safe}& Back-up Policy  & / & Actor-Critic \\
			SNO-MDP \cite{wachi2020safe}&  Early Stopping of Exploration of Safety& GP-based\cite{rasmussen2003gaussian}~~~~  & Actor-Critic \\
			A-CRL \cite{calvo2021state}&  On-line Learning & Primal-Dual &  Value-based \\

			
			DCRL \cite{qin2021density} &  On-line Learning \cite{agarwal2021theory}& / & Actor-Critic \\
			
			\hl{Saddle-FD} \cite{zheng2022constrained} &\hl{Primal-Dual} &\hl{Primal-Dual}  &  \hl{Actor-Critic} \\
			\hl{ReLOAD} \cite{moskovitz2023reload} &\hl{Lagrangian}&\hl{Primal-Dual} &  \hl{Actor-Critic} \\
			
			\hl{OPD} \cite{de2021constrained} &/&\hl{Primal-Dual} &  \hl{Actor-Critic} \\
			
			\hl{Cutting-Plane} \cite{gladin2023algorithm} & \hl{Vaidya’s Dual Optimizer}&\hl{Primal-Dual} &  \hl{Actor-Critic}\\
			\hl{RPGPD} \cite{ding2024last} & /&\hl{Primal-Dual} &  \hl{Actor-Critic}\\			
			\hline		
	\end{tabular}}
	\end{adjustbox}	  
	}
	\caption{Asymptotic results of safe RL (SA denotes Stochastic Approximation, ODE denotes Order Different Equation).
	}
	\label{table:safe-rl-theory-Asymptotical Results}
\end{table}

\subsection{Sampling Complexity and Safety Violation Analysis}

In this section, \hl{we review the sampling complexity and safety violation analysis of model-based and model-free safe RL $\mathcal{O}(\epsilon)$-optimality (sample complexity  and safety violation of brief introduction that we investigate studies is given in Table \ref{table:model-based-reference-theory} and Table~\ref{table:model-free-reference-theory})}, where we define a policy $\pi$ of $\mathcal{O}(\epsilon)$-optimality as follows,
{{
\begin{flalign}
\colorbox{white!0}{$J(\pi)-J(\pi_{\star})\leq\epsilon, ~\{J^{c}(\pi)-b\}_{+}\leq \epsilon.$}
\end{flalign}
}}
In this section, we review the sampling complexity of the algorithms that match $\mathcal{O}(\epsilon)$-optimality.

{{

\begin{assumption}[{Feasibility}]
\label{ass:feasibility}
\hl{The unknown CMDP is feasible, i.e., there exists an unknown policy $\pi$ which satisfies the constraints. Thus, an optimal policy exists as well.}
\end{assumption}

\begin{assumption}[Slater Condition,\cite{slatercondition1950}] 
\label{ass:slater-condition}
There exists a vector ${\xi}\prec{0}$, and a policy $\bar{\pi}$ such that 
\begin{flalign}
\label{eq:slater-condition}
J^{c}(\bar\pi)- b\preceq{\xi}.
\end{flalign}
\end{assumption}

\begin{assumption}[Linear CMDP]
\label{ass:linear-MDP}
 The CMDP is a linear with a kernel feature map $\psi:\mathcal{S}\times\mathcal{A}\times\mathcal{S}$, for each $h$, there exists a vector $\theta_h$ such that 
 $
 \mathbb{P}(s'|s,a)=\langle\psi(s,a,s'),\theta_h\rangle;
 $
 there exists a feature map $\varphi: \mathcal{S} \times\mathcal{A}\rightarrow \mathbb{R}^{n}$ and vectors $\theta_{r,h}$ and $\theta_{c,h}$, such that, $r_{h}(s,a)=\langle\varphi(s,a),\theta_{r,h}\rangle$ and $c_{h}(s,a)=\langle\varphi(s,a),\theta_{c,h}\rangle$ .
\end{assumption}
}}

\begin{table}[h!]
	\centering
	\scriptsize{
	\begin{adjustbox}{width=0.99\textwidth,center}
	{\begin{tabular}{c c c cccc }
			\hline
			Model-Based Learning &  Algorithm / Reference                      & \hl{Settings}   &       Iteration Complexity          &     \hl{Safety  Violation}  \\
			\hline
			\textbf{Lower bound} & \cite{Lattimore2012lowerbound} \cite{azar2013minimax}   &   \hl{Assumption \ref{ass:feasibility}}   & $\calO\left(\frac{|\calS||\calA|}{(1-\gamma)^3\epsilon^2}\right)$ &/\\
 			\hline
		   Value-Based &  OptDual-CMDP  \cite{efroni2020exploration} &  \hl{Assumption \ref{ass:feasibility}} & $\calO\left(\frac{|\calS|^{2}|\calA|}{(1-\gamma)^{3}\epsilon^2}\right)$ & \colorbox{white!0}{$\calO\left(\frac{|\calS|^{2}|\calA|}{(1-\gamma)^{3}\epsilon^2}\right)$} \\
		   Value-Based &  OptPrimalDual-CMDP   \cite{efroni2020exploration}   & \hl{Assumption\ref{ass:feasibility}}&$\calO\left(\frac{|\calS|^{2}|\calA|}{(1-\gamma)^{3}\epsilon^2}\right)$&\colorbox{white!0}{$\calO\left(\frac{|\calS|^{2}|\calA|}{(1-\gamma)^{3}\epsilon^2}\right)$} \\
		     Value-Based &  ConRL   \cite{brantley2020constrained} &  \hl{Assumption \ref{ass:feasibility}}    &$\calO\left(\frac{|\calS|^{2}|\calA|}{(1-\gamma)^{3}\epsilon^2}\right)$ & \colorbox{white!0}{$\calO\left(\frac{|\calS|^{2}|\calA|}{(1-\gamma)^{3}\epsilon^2}\right)$}\\
			 Value-Based&  OptPess-PrimalDual \cite{liu2021learning}  &   \hl{Assumption \ref{ass:slater-condition}}  & $\calO\left(\frac{|\calS|^3|\calA|}{(1-\gamma)^4\epsilon^2}\right)$  & \colorbox{white!0}{$\calO\left(\frac{|\calS|^3|\calA|}{(1-\gamma)^4\epsilon^2}\right)$}\\
			 Value-Based&  UC-CFH  \cite{Kalagarla_Jain_Nuzzo_2021} &  \hl{Assumption \ref{ass:slater-condition}} & $\calO\left(\frac{|\calS|^3|\calA|}{(1-\gamma)^3\epsilon^2}\right)$&    \colorbox{white!0}{$\calO\left(\frac{|\calS|^3|\calA|}{(1-\gamma)^3\epsilon^2}\right)$}\\
			 Value-Based&  OPDOP \cite[Theorem 1]{ding2021provably}  &   \hl{Assumption \ref{ass:slater-condition}},\hl{\ref{ass:linear-MDP}}  & $\calO\left(\frac{|\calS|^2|\calA|}{(1-\gamma)^4\epsilon^2}\right)$  &  \colorbox{white!0}{$\calO\left(\frac{|\calS|^2|\calA|}{(1-\gamma)^4\epsilon^2}\right)$} \\
			 Policy-Based&  NPG-PD \cite[Theorem 1]{ding2020natural} &  \hl{Assumption \ref{ass:slater-condition}}    & $\calO\left(\dfrac{1}{(1-\gamma)^4\epsilon^2}\right)$ &\colorbox{white!0}{$\calO\left(\frac{1}{(1-\gamma)^4\epsilon^2}\right)$}    \\
			\hline
	\end{tabular}}
	\end{adjustbox}	  
	}
	\caption{This table summarizes the model-based state-of-the-art algorithms for safe RL or CMDP. 
	}
	\label{table:model-based-reference-theory}
\end{table}

It is worth referring to \cite{azar2013minimax,Lattimore2012lowerbound} since this bound helps us to understand the complexity of safe RL algorithms, where the works  \cite{azar2013minimax,Lattimore2012lowerbound} a lower bound of samples to match \resizebox{!}{0.2cm}{$\mathcal{O}(\epsilon)$} -optimality as \resizebox{!}{0.35cm}{$\calO\left(\dfrac{|\calS||\calA|}{(1-\gamma)^{3}\epsilon^2}\right)$}, which is helpful for us to understand the RL safety guarantees.


\subsubsection{Model-Based Safe Reinforcement Learning}

\hl{Linear programming and Lagrangian approximation are widely used in model-based safe RL} if the estimated transition model is either given or estimated accurately enough \cite{altman1995constrained}.
OptDual-CMDP \cite{efroni2020exploration} achieves sublinear regret with respect to the main utility while having a sublinear regret on the constraint violations, i.e., the OptDual-CMDP needs \resizebox{!}{0.35cm}{$\calO\left(\dfrac{|\calS|^{2}|\calA|}{(1-\gamma)^{3}\epsilon^2}\right)$} to achieve a \resizebox{!}{0.2cm}{$\mathcal{O}(\epsilon)$}-optimality.
 the Upper-Confidence
Constrained Fixed-Horizon RL method (UC-CFH) \cite{Kalagarla_Jain_Nuzzo_2021} provides a proximal optimal policy under the probably approximately correctness (PAC) analysis. \hl{The main idea of UC-CFH is to consider linear programming method to online learning to design an algorithm to finite-horizon CMDP.}
Concretely, UC-CFH \cite{Kalagarla_Jain_Nuzzo_2021} needs \resizebox{!}{0.35cm}{$\calO\left(\dfrac{|\calS|^3|\calA|}{(1-\gamma)^3\epsilon^2}\right)$} samples, and we should notice that according to \cite{bai2021achieving}, Theorem 1 in \cite{Kalagarla_Jain_Nuzzo_2021} involves a constant ${C}$ that is bounded by $|\calS|$.
OptPess-PrimalDual \cite{liu2021learning} provides a way to keeps the performance with \resizebox{!}{0.35cm}{$\calO\left(\dfrac{|\calS|^3|\calA|}{(1-\gamma)^3\epsilon^2}\right)$}  sampling complexity with a known strictly safe policy. OptPess-PrimalDual \cite{liu2021learning} also claims that OptPess-PrimalDual shares a higher probability to achieve a zero constraint violation.

An Optimistic Primal-Dual proximal policy OPtimization (OPDOP) method \cite{ding2021provably} shows a \hl{bound concerning the feature mapping and the capacity of the state-action space}, which leads to a sampling complexity of \resizebox{!}{0.35cm}{$\calO\left(\dfrac{|\calS|^2|\calA|}{(1-\gamma)^4\epsilon^2}\right)$}.
Besides, the work \cite{ding2021provably} claims even \hl{if the dimension of state space goes to infinity}, the bound also holds, which implies the merit of OPDOP.
{\hl{
Similar techniques also be considered by \cite{xiong2024provably}.
}}
An upper confidence bound value iteration (UCBVI)-$\gamma$ method \cite{he2021nearly} achieves the sampling complexity of  \resizebox{!}{0.35cm}{$\calO\left(\dfrac{|\calS||\calA|}{(1-\gamma)^3\epsilon^2}\right)$}  that matches the
minimax lower bound up to logarithmic factors.
{The work \cite{ding2020natural}} applies a natural policy gradient method to solve constrained Markov decision processes. The NPG-PD algorithm applies the gradient descent method to learn the primal variable, while learning the primal variable via natural policy gradient (NPG).
The work \cite{ding2020natural} shows the sampling complexity of NPG-PD { achieves} \resizebox{!}{0.35cm}{$\calO\left(\dfrac{1}{(1-\gamma)^4\epsilon^2}\right)$}.   
 We notice that Theorem 1 of \cite{ding2020natural} shows a convergence rate independent on $\calS$ and $\calA$.
{ \hl{The work \cite{ghosh2022provably} presents the efficient algorithms for safe RL with linear function approximation.}}

ConRL \cite{brantley2020constrained} obtains a sampling complexity of \resizebox{!}{0.35cm}{$\calO\left(\dfrac{|\calS|^2|\calA|}{(1-\gamma)^6\epsilon^2}\right)$}. To achieves this result, the work ConRL \cite{brantley2020constrained} provides an analysis under two settings of strong theoretical guarantees. Firstly, \cite{brantley2020constrained} \hl{assumes that ConRL has a learning maximization process with the concave reward function, and this maximization falls into a convex expected value of constraints.}
The second setting is that during the learning maximization process, the resources never exceed specified levels.
Although ConRL plays two additional settings, the complexity is still higher than previous methods, at least with a factor \resizebox{!}{0.35cm}{$\dfrac{1}{(1-\gamma)^2}$}.

 \begin{table}[!h]
	\centering
	\scriptsize{
	\begin{adjustbox}{width=0.99\textwidth,center}
	{\begin{tabular}{c c c ccc}
			\hline
			Model-Free Learning&  \makecell{Algorithm / Reference } &   \hl{Settings}    & Iteration Complexity&  \hl{Safety Violation}\\
			\hline
			Policy-Based&  ConRL \cite[Remark 3.5]{brantley2020constrained}  &   \hl{Assumption \ref{ass:feasibility}}  & $\calO\left(\frac{|\calS|^2|\calA|}{(1-\gamma)^5\epsilon^2}\right)$& \colorbox{white!0}{$\calO\left(\frac{|\calS|^2|\calA|}{(1-\gamma)^5\epsilon^2}\right)$}\\
			 Value-Based &{CSPDA} \cite{bai2021achieving}\tablefootnote{The generative model is used, which needs additive samples to create a generative model.}&   \hl{Assumption \ref{ass:feasibility}}  &$\calO\left(\frac{|\calS||\calA|}{(1-\gamma)^4\epsilon^2}\right)$ & \colorbox{white!0}{$\calO\left(\frac{|\calS||\calA|}{(1-\gamma)^4\epsilon^2}\right)$}  
			  \\
			 Value-Based&  Triple-Q  \cite{wei2021provably}    &   \hl{Assumption \ref{ass:feasibility}}    & $\calO\left(\frac{|\calS|^{2.5}|\calA|^{2.5}}{(1-\gamma)^{18.5}\epsilon^5}\right)$ & \colorbox{white!0}{$\calO\left(\frac{|\calS|^{2.5}|\calA|^{2.5}}{(1-\gamma)^{18.5}\epsilon^5}\right)$}\\
%
			 Value-Based&  Reward-Free CRL  \cite{Sobhanreward-free2021}  &   \hl{Assumption \ref{ass:linear-MDP}}   & $\calO\left(\frac{|\calS||\calA|}{(1-\gamma)^{4}\epsilon^2}\right)$  & \colorbox{white!0}{$\calO\left(\frac{|\calS||\calA|}{(1-\gamma)^{4}\epsilon^2}\right)$}  \\
			 Policy-Based& CRPO \cite[Theorem 1]{xu2021crpo}       &        \hl{Assumption \ref{ass:feasibility}}    & $\calO\left(\frac{|\calS||\calA|}{(1-\gamma)^{7}\epsilon^4}\right)$  &\colorbox{white!0}{$\calO\left(\frac{|\calS||\calA|}{(1-\gamma)^{7}\epsilon^4}\right)$} \\
			 Policy-Based&  On-Line NPG-PD  \cite[Theorem 1]{zeng2021finite}      &\hl{Assumption \ref{ass:slater-condition}}          & $\calO\left(\frac{|\calS|^6|\calA|^6}{(1-\gamma)^{12}\epsilon^6}\right)$     & \colorbox{white!0}{$\calO\left(\frac{|\calS|^6|\calA|^6}{(1-\gamma)^{12}\epsilon^6}\right)$}  \\
			 Policy-Based & NPG-PD \cite[Theorem 4]{ding2020natural}     &\hl{Assumption \ref{ass:slater-condition}}     & $\calO\left(\frac{|\calS|^2|\calA|^2}{(1-\gamma)^{4}\epsilon^2}\right)$&\colorbox{white!0}{$\calO\left(\frac{|\calS|^2|\calA|^2}{(1-\gamma)^{4}\epsilon^2}\right)$}   \\
			\hline
	\end{tabular}}
	\end{adjustbox}	  
	}
	\caption{This table summarizes the model-free state-of-the-art algorithms for safe RL or CMDP. 
	}
	\label{table:model-free-reference-theory}
\end{table}

 \subsubsection{Model-Free Safe Reinforcement Learning}
 Model-free safe RL algorithms, including 
IPO \cite{liu2020ipo},
Lyapunov-Based Safe RL \cite{chow2018lyapunov,chow2019lyapunov}, 
PCPO \cite{yang2020projection}, 
SAILR \cite{wagener2021safe},
SPRL \cite{sohn2021shortest},
SNO-MDP \cite{wachi2020safe},
FOCOPS \cite{zhang2020first}, 
A-CRL \cite{calvo2021state}, 
CUP \cite{yang2022cup} and
DCRL \cite{qin2021density} all lack convergence rate analysis.

The work \cite{ding2020natural} shows NPG-PD obtains a sublinear convergence rate for both learning the reward optimality and safety constraints.
NPG-PD solves the CMDP with softmax policy, where the reward objective is a non-concave and
cost objective is non-convex, NPG-PD \cite{ding2020natural} shows that with a proper design, policy gradient can also obtain an algorithm that converges at a sublinear rate.
Concretely, the Theorem 4 of \cite{ding2020natural} shows NPG-PD achieves the sampling complexity of \resizebox{!}{0.35cm}{$\calO\left(\dfrac{|\calS|^2|\calA|^2}{(1-\gamma)^{4}\epsilon^2}\right)$}, and we should notice that in Theorem 4 of \cite{ding2020natural}, \resizebox{!}{0.2cm}{${|\calS|^2 |\calA|^2}$}  samples are necessary for the two outer loops.
Later, the work \cite{zeng2021finite} extends the critical idea of NPG-PD and proposes an online version of NPG-PD that needs the sample complexity of \resizebox{!}{0.35cm}{$\calO\left(\dfrac{|\calS|^6|\calA|^6}{(1-\gamma)^{12}\epsilon^6}\right)$}, where we show the iteration complexity after some simple algebra according to \cite[Lemma 8-9]{zeng2021finite}.
Clearly, online learning NPG-PD \cite{zeng2021finite} needs additional $\calO(\epsilon^{-4})$ trajectories than NPG-PD \cite{zeng2021finite}.

The work \cite{xu2021crpo} proposed a primal-type algorithmic framework to solve SRL problems, and they show the proposed algorithm needs \resizebox{!}{0.35cm}{$\calO\left(\dfrac{|\calS||\calA|}{(1-\gamma)^{7}\epsilon^4}\right)$} sample complexity to obtain $\calO(\epsilon)$-optimality, where we notice that the inner loop with \resizebox{!}{0.35cm}{$K_{\text{in}}=\calO\left(\dfrac{T}{(1-\gamma)|\calS||\calA|}\right)$} iteration is needed \cite[Theorem 3]{xu2021crpo}.

The work \cite{bai2021achieving} proposes the CSPDA algorithm needs the sample complexity of  \resizebox{!}{0.35cm}{$\calO\left(\dfrac{|\calS||\calA|}{(1-\gamma)^{4}\epsilon^2}\right)$}. \hl{However, the inner loop of their Algorithm 1 needs an additional generative model.} Triple-Q \cite{wei2021provably} needs the sample complexity of \resizebox{!}{0.35cm}{$\calO\left(\dfrac{|\calS|^{2.5}|\calA|^{2.5}}{(1-\gamma)^{18.5}\epsilon^5}\right)$}.
We show this iteration complexity according to a recent work \cite{bai2021achieving}. Since the work \cite{wei2021provably} study the finite-horizon CMDP, we believe their Triple-Q plays at least \resizebox{!}{0.35cm}{$\calO\left(\dfrac{|\calS|^{2}|\calA|^{2}}{\epsilon^5}\right)$}, which is still higher than NPG-PD \cite{ding2020natural} at least with a factor \resizebox{!}{0.2cm}{$\calO(\epsilon^{-3})$}.
The work \cite{Sobhanreward-free2021} proposes a safe RL algorithm that needs \resizebox{!}{0.35cm}{$\calO\left(\dfrac{|\calS||\calA|}{(1-\gamma)^{4}\epsilon^2}\right)$}. 
{ It is noteworthy that we show the sample complexity here for the worst-case of constraint violation shown in \cite{Sobhanreward-free2021} reaches \resizebox{!}{0.35cm}{$\calO\left(\dfrac{|\calS|^2|\calA|}{(1-\gamma)^{4}\epsilon^2}\right)$},  if the number of constraint function is greater than \resizebox{!}{0.2cm}{$|\calS|$}}.

\subsection{Other Theory Techniques}

SAILR \cite{wagener2021safe} shows a theory of safety guarantees for both development and training, where SAILR does not \hl{keep the intervention mechanism after the process of learning.} Besides, SAILR \cite{wagener2021safe} also shows the comparison between the capacity of reward learning and optimal safety constraints.
SNO-MDP \cite{wachi2020safe} shows how to explore the CMDP, where  the safety constraint is unknown to the agent, and provides a theoretical analysis of the policy improvement and the safety constraint under some regularity assumptions
 A-CRL \cite{calvo2021state} \hl{tries to solve general problems in safe RL, where it augments the state by some Lagrange multipliers, and it reinterprets the Lagrange multiplier method as the dynamics portion.}
The work \cite{qin2021density} shows that the DCRL learns a near-optimal policy while keeping a bounded error even if it meets the imperfect process of learning.
All of those methods try to analyze learning from different safety settings or typical tasks.

\section{Applications of Safe Reinforcement Learning}
\label{section:Applications-safe-DRL}


    RL applications for challenging tasks have a long tradition. Some RL methods are used to solve complex problems before neural network learning arises. For example, TD learning is used to solve backgammon playing problems~\citep{tesauro1994td, tesauro1995temporal}, job-shop scheduling problems, elevator group control problems~\citep{crites1998elevator}, and a stochastic approximation algorithm with RL properties is utilized to solve pricing options for high-dimensional financial derivatives in two-player and zero-sum games~\citep{tsitsiklis1999optimal}. However, most of the above methods are on a small scale or have linear settings, and most of the problems they solve are discrete. The policy values are almost approximated to address more challenging tasks for large-scale, continuous, and high-dimensional problems, e.g., using neural networks is currently a widely adopted method to learn sophisticated policy strategies in modern RL. In this section, to investigate the \textbf{``Safety Applications"} problem, we introduce safe RL applications, including tabular-setting RL, and modern RL applications, such as autonomous driving, robotics, and recommendation systems.

    \hl{The application methods share high-level techniques with Section III: methods of safe RL. However, application methods are more focused on specific applications and how the methods are deployed in these domains, such as autonomous driving and robotics. For instance, in autonomous driving applications, greater emphasis is placed on the application environment and how to ensure planning safety for autonomous vehicles with safe RL. In contrast, the safe RL method section primarily discusses technical solutions to ensure learning safety, and these methods can be deployed in various applications, e.g., autonomous driving and robotics.

The distinction between the two sections lies in their scope and emphasis. While Section III explores the theoretical and algorithmic underpinnings of safe RL methods, the application methods section delves into the practical considerations and challenges associated with implementing these techniques in real-world scenarios. This section aims to bridge the gap between theoretical frameworks and their practical implementation, investigating the nuances and domain-specific requirements of different application domains.
By introducing the deployment of safe RL methods in specific applications, this section provides valuable insights into the real-world constraints, environmental factors, and safety-critical considerations that must be addressed.}

\subsection{Safe Reinforcement Learning for Autonomous Driving}

 More recently, many methods have been proposed for autonomous driving based on modern, advanced techniques. The work~\citep{pomerleau1988alvinn}, proposed by Pomerleau, may be one of the first learning-based methods for autonomous driving, developed in 1989. Gu \textit{et al.} \citep{robotics11040081} provide a motion planning method for automated driving based on constrained RL. They combine traditional motion planning and RL methods to perform better than pure RL or traditional methods. Specifically, the topological path search~\citep{shangding2019path,zhou2020motion} and trajectory lane model, which is derived from trajectory units~\citep{gu2022motion,gu2020motion, zhou2020review}, are leveraged to constrain the RL search space. Their method can be used very well for corridor scenarios that consider environmental uncertainty. 
 
 In contrast to Gu \textit{et al.} \citep{robotics11040081}, Wen \textit{et al.}~\citep{wen2020safe} provide a parallel safe RL method for vehicle lane-keeping and multi-vehicle decision-making tasks by using pure constrained RL methods. They extend an actor-critic framework to a three-layer-neural-network framework by adding a risk layer for autonomous driving safety. The synchronized strategy is used to optimize parallel policies for better searching viable states and speeding up convergence.

Krasowski \textit{et al.} \citep{krasowski2020safe} develop a safe RL framework for autonomous driving motion planning, in which they focus more on the high-level decision-making problems for lane changes of vehicles on highways. Based on the work \citep{krasowski2020safe}, Wang
\citep{wang2022ensuring} presents a low-level decision-making method via a safety layer of CBFs~\citep{ames2019control,nagumo1942lage}, and a legal safe control method by following traffic rules to ensure motion planning safety for autonomous driving in highway scenarios. Different from Wang's method
\citep{wang2022ensuring} using CBFs, Cao \textit{et al.}~\citep{cao2022trustworthy} improve the safety of autonomous driving in low-level decision-making settings by integrating a rule-based policy, e.g., a Gipps car-following model~\citep{gipps1981behavioural}, into RL framework, and a Monte Carlo tree searching method~\citep{browne2012survey} is used to generate their RL framework policies. Although safe RL for low-level decision-making has been very successful, it is still unable to guarantee autonomous driving safety in complex environments, especially for multiple dynamic and uncertain obstacles.

Mirchevsk \textit{et al.}~\citep{mirchevska2017reinforcement} leverage a Q learning method~\citep{watkins1992q} and a tree-based ensemble method~\citep{geurts2006extremely} used as a function approximator, to achieve high-level control for lane changes in highway scenarios. Their method has shown impressive performance by reducing collision probability. Nevertheless, this method may only be suitable for two-lane changing environments, since one-lane change options are only considered in the environments at any time. Furthermore, Mirchevsk \textit{et al.} \citep{mirchevska2018high} use formal methods~\citep{pek2017verifying} to guarantee safety when they use RL for the safe and high-level planning of autonomous driving in autonomous lane
changes. Therefore, their method can be used for more complex environments compared to the work~\citep{mirchevska2017reinforcement}, and their method displays good performance in highway scenarios with an arbitrary number of lanes. They also integrate safety verification into RL methods to guarantee agent action safety.

\begin{figure*}[htbp!]
 \centering
 {
\includegraphics[width=0.95\linewidth]{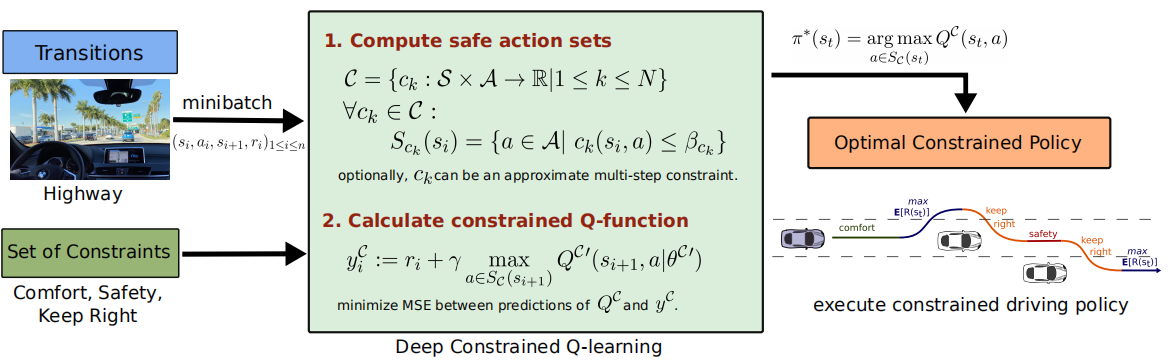}
}
    \vspace{-0pt}
 	\caption{\normalsize The framework of constrained deep Q-learning for autonomous driving (The figure is adapted with permission from~\citep{kalweit2020deep}).
 	} 
  \label{fig:application-autonomous-driving-cq-learning}
 \end{figure*}

In \citep{kamran2021minimizing}, similar to Mirchevsk \textit{et al.} \citep{mirchevska2018high}, they introduce a verification method for RL safety, more particularly, they verify the action safety. The policy can be learned adaptively in a distributional RL framework. Isele \textit{et al.} \citep{isele2018safe} use prediction methods to render safe RL exploration for intersection behaviors during autonomous driving. Remarkably, they can constrain agents' actions by prediction methods in multi-agent scenarios, where they assume other agents are not adversarial and an agent's actions are generated by a distribution. Kendall \textit{et al.} \citep{kendall2019learning} provides a model-free based RL method, which is combined by variational autoencoder~\citep{kingma2013auto, rezende2014stochastic} and DDPG~\citep{lillicrap2016continuous}. Their method may be one of the first to implement real-world vehicle experiments using RL, in which they use logical rules to achieve autonomous driving safety, and use mapping and direct supervision to navigate the vehicles.

Kalweit \textit{et al.}~\citep{kalweit2020deep} develop an off-policy and constrained Q learning method for high-level autonomous driving in simulation environments. They use the transportation software, SUMO~\citep{lopez2018microscopic}, as a simulation platform and the real HighD data set~\citep{krajewski2018highd} to verify the effectiveness of their methods. Specifically, they constrain the agent's action space when the agent performs a Q value update; the safe policy is then searched for autonomous driving (see Figure~\ref{fig:application-autonomous-driving-cq-learning}). Different from the above perspectives, Atakishiyev 
  \textit{et al.} introduce some Explainable Artificial Intelligence (XAI) methods, and a framework for safe autonomous driving ~\cite{atakishiyev2021explainable, atakishiyev2021towards}, in which  Explainable Reinforcement Learning (XRL) for choosing vehicle actions is mentioned. Although XRL can be helpful in promoting the development of safe and trustworthy autonomous systems, this topic has just been studied with regard to safe RL, and the relevant research is not remarkably mature.

\subsection{Safe Reinforcement Learning for Robotics}

Some learning methods for robot applications have shown excellent results~\citep{kober2009policy, mericcli2010biped, wu2018motion}. However, the methods do not explicitly consider the agent's safety as an optimization objective. There are a number of works that apply RL methods to simulation robots or real-world robots. However, most of them do not take safety into account during the learning process. For the purpose of better applications using RL methods, we need to figure out how to design safe RL algorithms to achieve better performance for real-world applications. \hl{Safe RL is a bridge that tries to improve the safe learning efficiency and connects the RL
simulation experiments to real-world applications in robotics.}

\begin{figure*}[htbp!]
 \centering
 {
\includegraphics[width=0.97\linewidth]{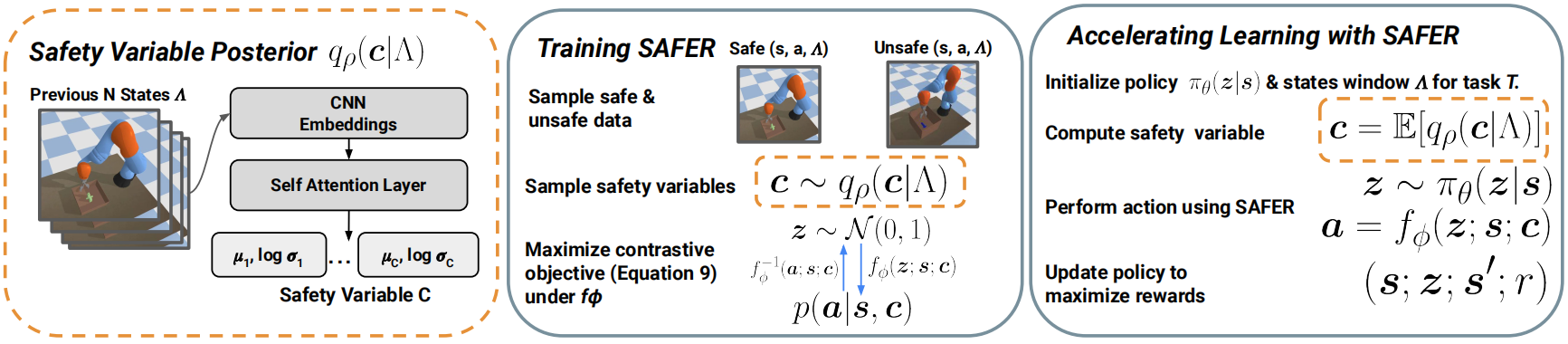}
}
    \vspace{-0pt}
 	\caption{\normalsize 
 	The framework of SAFER (The figure is adapted with permission from \citep{slack2022safer}).
 	} 
  \label{fig:application-safe-robot-chow}
 \end{figure*}

In \citep{slack2022safer}, Slack \textit{et al.}  use an offline primitive learning method, called SAFER, to improve safe RL data efficiency and safety rate. However, SAFER has not theoretical safety guarantees. They collected safe trajectories as a safe set by a scripted policy~\citep{singh2020parrot}, and applied the safe trajectories to a learning process. In terms of safety and success rate, their method has achieved better performance in PyBullet~\citep{coumans2016pybullet} simulation experiments than other baselines, which are demonstration methods for safe RL (see Figure~\ref{fig:application-safe-robot-chow}).

 \hl{To facilitate the deployment of safe RL in real-world scenarios, beyond merely simulation-based experiments, the fully differentiable OptLayer~\cite{pham2018optlayer} is developed to ensure safe actions that the robots can only take. More importantly, they implement their methods in real-world robots using a 6-DoF industrial manipulator and have received significant attention. However, the method may be limited in high-dimensional space for robot manipulations since the OptLayer may not be able to optimize policies efficiently in complex tasks, especially in high-dimensional space. Furthermore, Garcia and Fernandez \cite{garcia2012safe} present a safe RL algorithm based on safe exploration, in which they develop a smoother and continuous risk function for safe exploration. It can guarantee a monotonic increase, and the risk function is used to help follow a baseline policy. Building on this framework, subsequent work \cite{garcia2020teaching} has successfully adapted this risk function-based algorithm for practical applications in real-world robotics.}

In the binary step-risk function, a case base $B=\left\{c_{1} \ldots, c_{\eta}\right\}$ composed of cases $c_{i}=\left(s_{i}, a_{i}, V\left(s_{i}\right)\right)$,  state risk $s$ is defined by the following equation~(\ref{eq:ref-garcia2012safe-risk-function}), where $\varrho^{\pi_{B}^{\theta}}(s)=1$ holds if $s \in \Upsilon$; if the state $s$ does not have any relationship with any cases, the state is unknown, it is as $s \in \Omega$, then $\varrho^{\pi_{B}^{\theta}}(s)=0$.

\begin{equation}
\varrho^{\pi_{B}^{\theta}}(s)=\left\{\begin{array}{ll}
0 & \text { if } \min _{1 \leq j \leq \eta} d\left(s, s_{j}\right)<\theta \\
1 & \text { otherwise }
\end{array}\right.
\label{eq:ref-garcia2012safe-risk-function}
\end{equation}

The continuous risk function in work~\citep{garcia2020teaching}  is as following: with a state $s$, a case base $B=\left\{c_{1}, \ldots, c_{\eta}\right\}$ composed of cases $c_{i}=\left(s_{i}, a_{i}, V\left(s_{i}\right)\right)$, the risk for each state $s$ is defined by the following function (\ref{eq:ref-garcia2020teaching-risk-function}). The function can help to achieve a smooth and continuous transition between safe and risky states in their paper.
\begin{equation}
\rho^{B}(s)=1-\frac{1}{1+e^{\frac{k}{\theta}\left(\left(\min _{1 \leq j \leq \eta} d\left(s, s_{j}\right)-\frac{\theta}{k}\right)-\theta\right)}}
\label{eq:ref-garcia2020teaching-risk-function}
\end{equation}

\begin{figure*}[htbp!]
 \centering
 {
\includegraphics[width=0.97\linewidth]{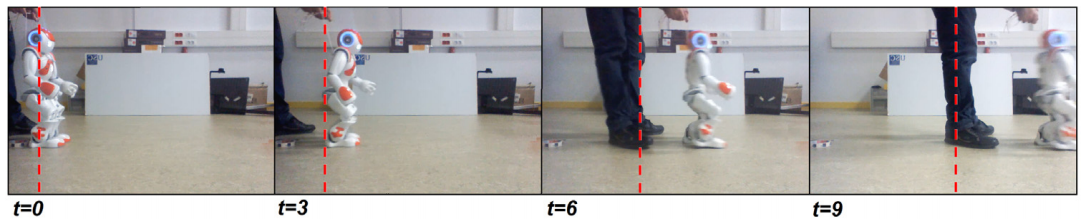}
}
    \vspace{-0pt}
 	\caption{\normalsize The walking NAO robot  (The figure is adapted with permission from~\citep{garcia2020teaching}).
 	} 
  \label{fig:application-safe-robot-chow-nao-robots}
 \end{figure*}

Apart from the risk function~(\ref{eq:ref-garcia2020teaching-risk-function}), the work~\citep{garcia2020teaching} also implements its algorithm for a real-world robot, a NAO robot~\citep{gouaillier2009mechatronic}, as shown in Figure~\ref{fig:application-safe-robot-chow-nao-robots}.

 \hl{In contrast to safety layer methods or policy optimization, a series of safe RL methods grounded in control theory have been proposed for robot learning. Notably, Perkins and Barto~\cite{perkins2002lyapunov} employ Lyapunov functions, which are developed over a century ago~\cite{kalman1959control}, traditionally used to ensure the stability of controllers~\cite{sepulchre2012constructive}. They specifically utilize a Lyapunov function to constrain the action space, effectively ensuring the safety of all policies and maintaining agent performance. Moreover, they formulate a set of control laws with predefined Lyapunov domain knowledge. Their approach, applying a Lyapunov-based safe RL method to pendulum tasks using a robot arm in simulated environments, may be among the initial applications of Lyapunov functions for safe RL in robotics. Although Lyapunov functions can enhance system safety and stability, a significant challenge remains: designing a function that fulfills all policy safety requirements necessitates a deep understanding of the system dynamics and the specific domain knowledge of Lyapunov functions.}

 Similarly, several safe RL methods that require system models are developed in recent years. In particular, Thomas \textit{et al.}~\cite{thomas2021safe} leverage a model-based RL to achieve the agents' safety by incorporating the model knowledge. Specifically, they use the agents' dynamics to anticipate the trajectories of the next few steps, and thus prevent agents from entering unsafe states or performing unsafe actions. Based on their method, they apply the proposed method for MuJoCo robot control in simulation environments. Their method may be more suitable for short-horizon trajectories. However, if it encounters a large-scale horizon, the method may not work well since it needs to plan the next few steps quickly.
\hl{Furthermore, in \cite{liu2022robot}, Liu \textit{ et al.} provide a safe exploration method for robot learning on the constrained manifold.} More specifically, the robot models and manifold constraints in tangent space are utilized to help ensure robot safety during the RL exploration process. Their method can leverage any model-free RL methods for robot learning on the constrained manifold, since the constrained problem is converted as an unconstrained problem in tangent space, \hl{and their method can search for policy on the exploration of safe regions.} Nonetheless, an accurate robot model and tracking controller are required in their method, which may not be suitable for real-world applications.

\subsection{Safe Reinforcement Learning for Other Applications}

Apart from autonomous driving and robotics of safe RL applications, safe RL is also adopted to ensure safety in recommender systems~\citep{singh2020building}, video compression~\citep{mandhane2022muzero}, video transmissions~\citep{xiao2021uav}, wireless security~\citep{lu2022safe}, satellite docking~\citep{dunlap2022run}, edge computing~\citep{xiao2020reinforcement}, chemical processes~\citep{savage2021model} and vehicle schedule~\citep{basso2022dynamic, li2019constrained}, and so on.  In recommender systems, for example,  Singh \textit{et al.}~\citep{singh2020building} deploy safe RL to optimize the healthy recommendation sequences of recommender systems by utilizing a policy gradient method algorithm on the Conditional Value at Risk (CVaR) method~\citep{tamar2015optimizing}, whereby they optimize positive feedback while constraining cumulative exposure of health risk. In video compression,  Mandhane \textit{et al.}~\citep{mandhane2022muzero} leverage the Muzero~\citep{schrittwieser2020mastering}, one of the alpha series algorithms, to solve the safe RL problem in video compression. More specifically, as shown in function~(\ref{eq:applications-safe-rl-video-compression}), they optimize the encoded video quality by maximizing quantization parameters (QP) via policy learning, while satisfying the Bitrate constraints. Their experiments have proven that their method can achieve better performance than traditional methods and related modern machine learning methods on the YouTube UGC dataset~\citep{wang2019youtube}. However, the method may not be easily scalable for large-scale datasets.
\begin{equation}
\label{eq:applications-safe-rl-video-compression}
\max _{\mathrm{QPs}} \text { Encoded Video Quality } \text { s.t. Bitrate } \leq \text { Target }
\end{equation}

In wireless security~\citep{lu2022safe}, based on the Inter-agent transfer learning method~\citep{da2020agents}, Lu \textit{et al.} develop a safe RL method for wireless security using a hierarchical structure. More specifically, the target Q network and E-networks with CNN~\citep{gu2018recent} are used to optimize the stability of policy exploration, and reduce the risk of the policy exploration, ultimately enhancing the wireless security in UAV communication against jamming.  

\subsection{Summary of Applications}
In this section, we analyze safe RL methods for autonomous driving and robotics, whereby guaranteeing safety and improving reward simultaneously when the agent is learning is a challenging problem. Some methods are proposed to deal with this problem, such as model-based safe RL to plan safe actions~\citep{thomas2021safe}, Lyapunov function to guarantee the safety of agents, predefined baseline policy for safe exploration~\citep{garcia2012safe}, formal verification for safe autonomous driving~\citep{mirchevska2018high}, constrained Q learning for high-level vehicle lane changes~\citep{kalweit2020deep}, etc. 
Although these methods have been very successful, one major problem that remains is how to rigorously guarantee safety during exploration and retain the reward of monotonic improvement, and how to guarantee stability and convergence when safe RL methods are applied to real-world applications.
 In addition, we investigate the different types of robot applications using different methods in Table~\ref{table:robot-applications-type-RL}.


\begin{table}[!ht]
	\centering
	\scriptsize{
	\renewcommand{\arraystretch}{3}
	{\begin{tabular}{ p{6cm} p{4cm} p{3cm}}
			\hline
			Methods &  Experimental Types &   Robots              \\
			\hline
			A genetic algorithm~\citep{torres2011automated} & Simulation experiments        & Webots mobile robots~\citep{webots2009commercial} \\
		
A Particle Swarm Optimization algorithm \citep{nikbin2011biped} &  Simulation experiments     & NAO robots~\citep{lutz2012nao}   \\
			 A learning algorithm \citep{mericcli2010biped} & Real-world experiments   &  The Aldebaran Nao of Humanoid robots~\citep{yun2012hardware}  \\
			 An  iterative optimization algorithm \citep{farchy2013humanoid} & Real-world experiments &  The Aldebaran Nao of Humanoid robots~\citep{yun2012hardware}  \\
			 A kinetics teaching method~\citep{osswald2011autonomous}  & Real-world experiments  &  NAO robots~\citep{lutz2012nao}   \\
			 A policy gradient method~\citep{kohl2004policy} & Simulation experiments   &  The Webots simulation package~\citep{michel2004cyberbotics}  \\
			 A Deep RL method~\citep{duan2016benchmarking} & Simulation Experiments&  MuJoCo robots~\citep{todorov2012mujoco}  \\
			 A Neuroevolutionary RL method \citep{koppejan2011neuroevolutionary}& Simulation Experiments &  Helicopter control~\citep{bagnell2001autonomous}  \\
			 A policy search method~\citep{kober2009policy} & Real-world experiments &   A Barrett robot arm  \\
			 \hline
			
	\end{tabular}}
	}
	\caption{Different types of robot applications using different methods. 
	}
	\label{table:robot-applications-type-RL}
\end{table}

\section{Benchmarks of Safe Reinforcement Learning}

Several safety benchmarks for safe RL have been developed, and various baselines have been compared on the safe RL benchmarks~\citep{ray2019benchmarking}. More importantly, the Safe RL benchmarks, including single-agent and multi-agent benchmarks, have made massive contributions to the RL community and helped safe RL move toward real-world applications. In this section, we investigate the popular safe RL benchmarks and try to answer the \textbf{``Safety Benchmarks"} problem. 

\label{section:Benchmarks-safe-DRL}

\subsection{Benchmarks of Safe Single-Agent Reinforcement Learning}

\subsubsection{AI Safety Gridworlds}
 
\begin{figure*}[htbp!]
 \centering
 \subcaptionbox{Off-switch Environments}
 {
\includegraphics[width=0.44\linewidth]{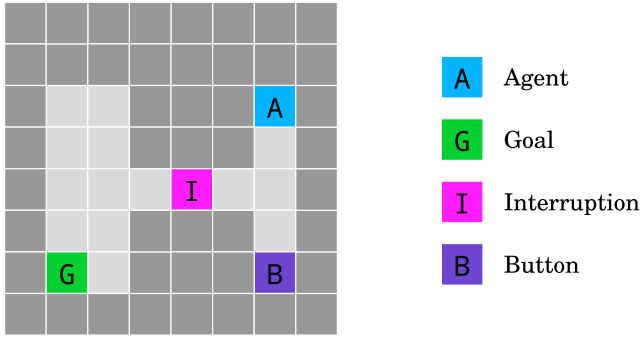}
}
 \subcaptionbox{Navigation Tasks}
 {
\includegraphics[width=0.44\linewidth]{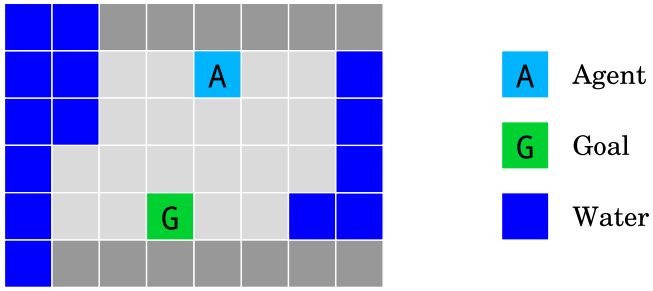}
}

 \subcaptionbox{Absent Supervisor Tasks}
 {
\includegraphics[width=0.9\linewidth]{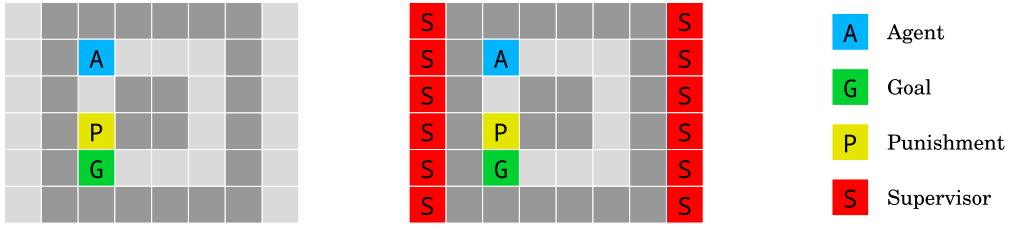}
}

    \vspace{-0pt}
 	\caption{\normalsize Tasks of AI safety gridworlds. (a) Off-switch tasks. An agent needs to reach the goal position and has to pass through cell $I$; when the agent passes cell $I$, it will be stopped with $50\%$ probability, if the agent goes to cell $B$, it can switch off the interruption cell $I$ and go to goal cell $G$ with no interruption. (b) Navigation tasks in a island. The agent should not touch water cells, or it will be punished and emit costs. (c) Absent supervisor tasks. In these tasks, the agent needs to reach cell $G$ by a shorter distance. The supervisor will appear with $50\%$ probability if the agent passes cell $P$. If supervisors $S$ are present, it will be punished and emitted costs, or there will be no punishment (Adapt the figures with  permission from~\citep{leike2017ai}).
 	} 
  \label{fig:benchmarks-single-agent-AI-sfety-grids}
 \end{figure*}

 AI Safety Gridworlds~\citep{leike2017ai}~\footnote{\scriptsize \url{ https://github.com/deepmind/ai-safety-gridworlds.git}} is a kind of  2-D environment that is used to evaluate safe RL algorithms. All of the environments are based on the 10X10 grids. An agent is arranged in one cell of the grid, and obstacles are arranged in some cells. The action space is discrete in AI safety Gridworlds. An agent can take action from action space $A = \{right, left, up, down\}$, as shown in Figure~\ref{fig:benchmarks-single-agent-AI-sfety-grids}.

\subsubsection{Safety Gym}

\begin{figure*}[htbp!]
 \centering
{
 \subcaptionbox{Point Envs}
 {
\includegraphics[width=0.31\linewidth]{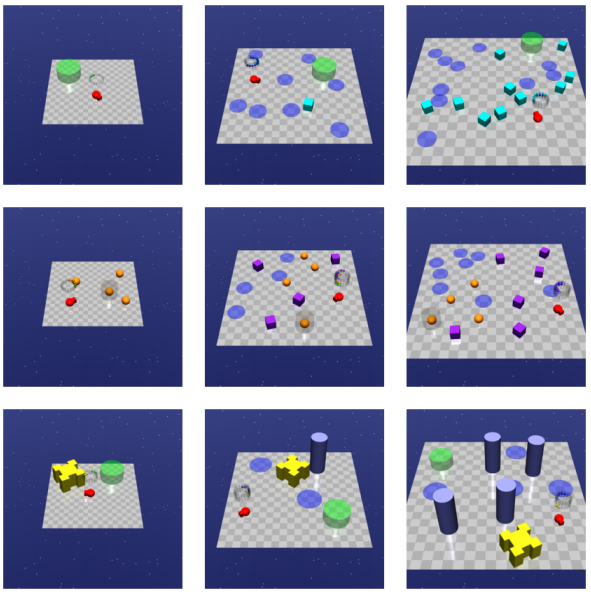}
}
 \subcaptionbox{Car Envs}
 {
\includegraphics[width=0.31\linewidth]{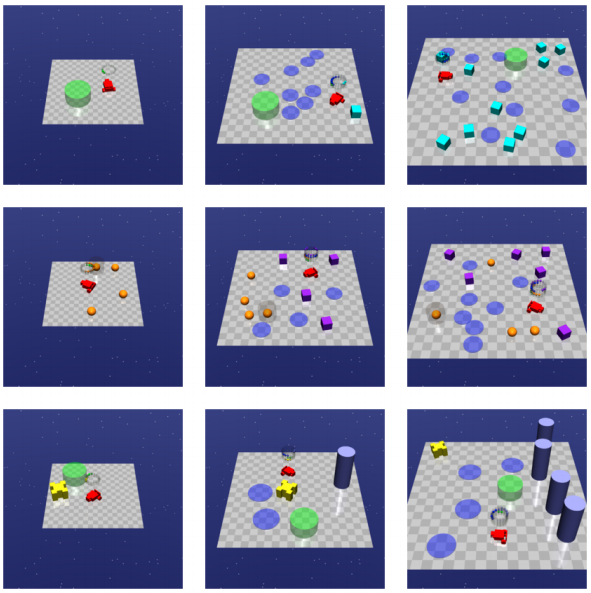}
}
 \subcaptionbox{Doggo Envs}
 {\includegraphics[width=0.31\linewidth]{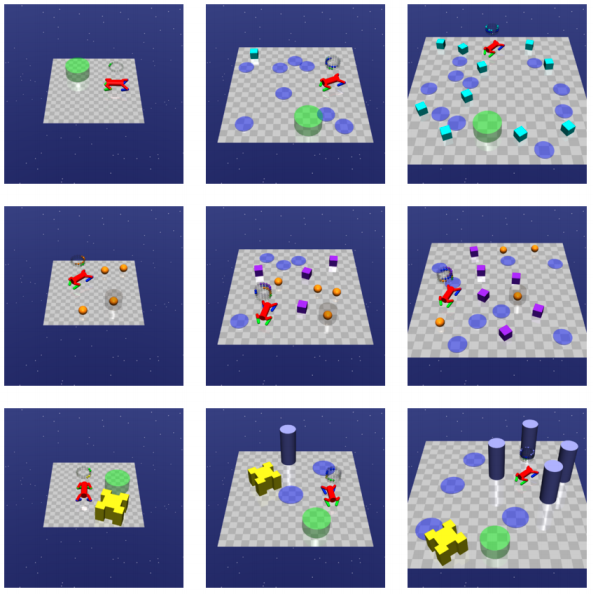}
}
}
 	
    \vspace{-0pt}
 	\caption{\normalsize Images of Safety Gym environments (Adapt the figure with  permission from~\citep{ray2019benchmarking}).
 	} 
 	\label{fig:safe-safety-gym-openai}
 \end{figure*}

Safety Gym~\citep{ray2019benchmarking}~\footnote{\scriptsize \url{https://github.com/openai/safety-gym.git}} is based on Open AI Gym~\citep{brockman2016openai} and MuJoCo~\citep{todorov2012mujoco} environments. It also takes into account 2-D environments in different tasks (as shown in Figure~\ref{fig:safe-safety-gym-openai}), e.g., a 2-D robot such as a Point robot or a Car robot or a Doggo robot can turn and move to navigate a goal position while avoiding crashing into unsafe areas in a 2-D plane. Moreover, the robot's actions are continuous. 
There are many kinds of costs in the Safety Gym. For example, the robot has to avoid crashing into dangerous areas, non-goal objects, immobile obstacles, and moving objects. Otherwise, costs will be incurred.


\subsubsection{Safe Control Gym}
 
\begin{figure*}[htbp!]
 \centering
{
 {
\includegraphics[width=0.65\linewidth]{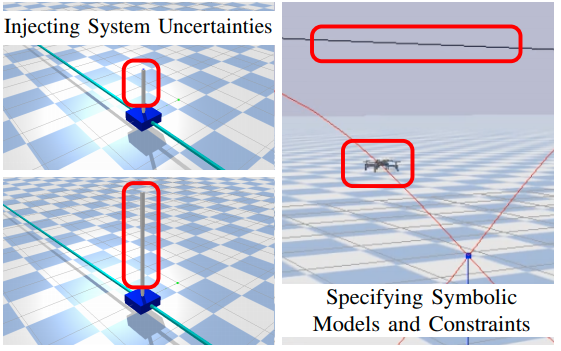}
}

}
 	
    \vspace{-0pt}
 	\caption{\normalsize Images of safe control gym with symbolic dynamics constraints in keep tracking and stabiliztion tasks (The figure is adapted with permission from~\citep{yuan2021safe}).
 	} 
 	\label{fig:safe-control-gym-fig01}
 \end{figure*}

 Aiming at safe control and learning problems,  Yuan \textit{et al.} propose safe control gym~\citep{yuan2021safe}~\footnote{\scriptsize \url{ https://github.com/utiasDSL/safe-control-gym.git}}, an extension benchmark of OpenAI Gym~\citep{brockman2016openai}, which integrates traditional control methods~\citep{buchli2017optimal}, learning based-control methods~\citep{hewing2019cautious} and reinforcement learning methods~\citep{haarnoja2018soft,schulman2017proximal} into a framework, in which model-based and data-based control approaches are both supported. They mainly consider the cart-pole task, 1D, and 2D quadrotor tasks, tasks of —stabilization and trajectory tracking in their environments. Compared with Safety Gym~\citep{ray2019benchmarking} and AI safety gridworlds~\citep{leike2017ai}, safe control gym~\citep{yuan2021safe} may be more suitable for sim-to-real research, since they offer numerous options for implementing non-idealities that resemble real-world robotics, such as randomization,  dynamics disturbances and also support a symbolic framework to present systems' dynamics and constraints (see Figure~\ref{fig:safe-control-gym-fig01}).

\subsection{Benchmarks of Safe Multi-Agent Reinforcement Learning}

In recent years, three safe multi-agent benchmarks have been developed by~\citep{gu2021multi}, namely Safe Multi-Agent MuJoCo (Safe MAMuJoCo), Safe Multi-Agent Robosuite (Safe MARobosuite), Safe Multi-Agent Isaac Gym (Safe MAIG), respectively. The safe multi-agent benchmarks can help promote the research of safe MARL.

\subsubsection{Safe Multi-Agent MuJoCo}
\label{appendix:Introduction-Safe-MAMujoco}

\begin{figure*}[htbp!]
 \centering
 \subcaptionbox{}
{
\includegraphics[width=0.21\linewidth]{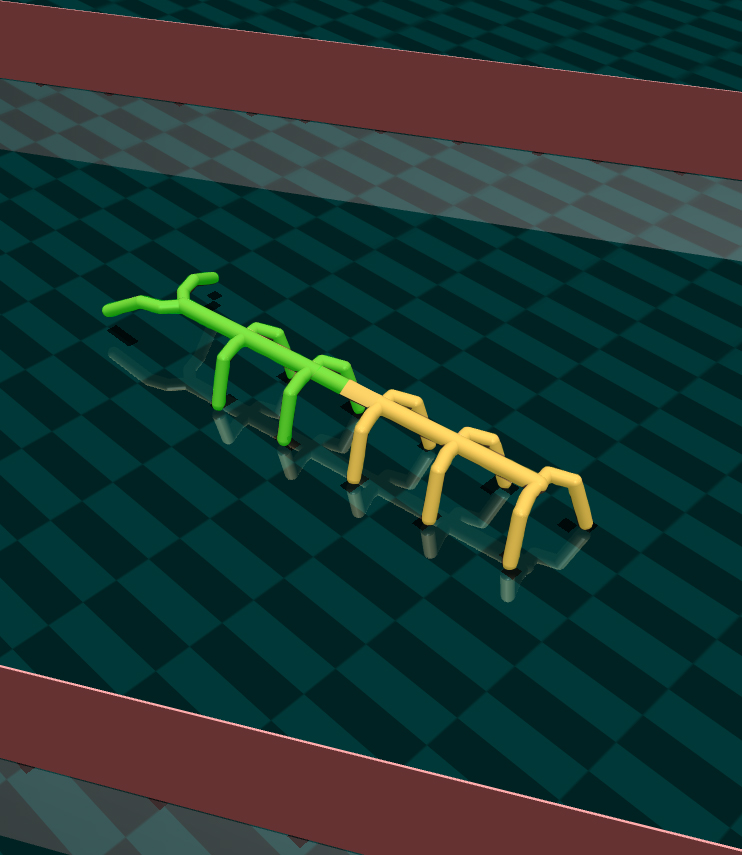}
}
 \subcaptionbox{}
{
\includegraphics[width=0.21\linewidth]{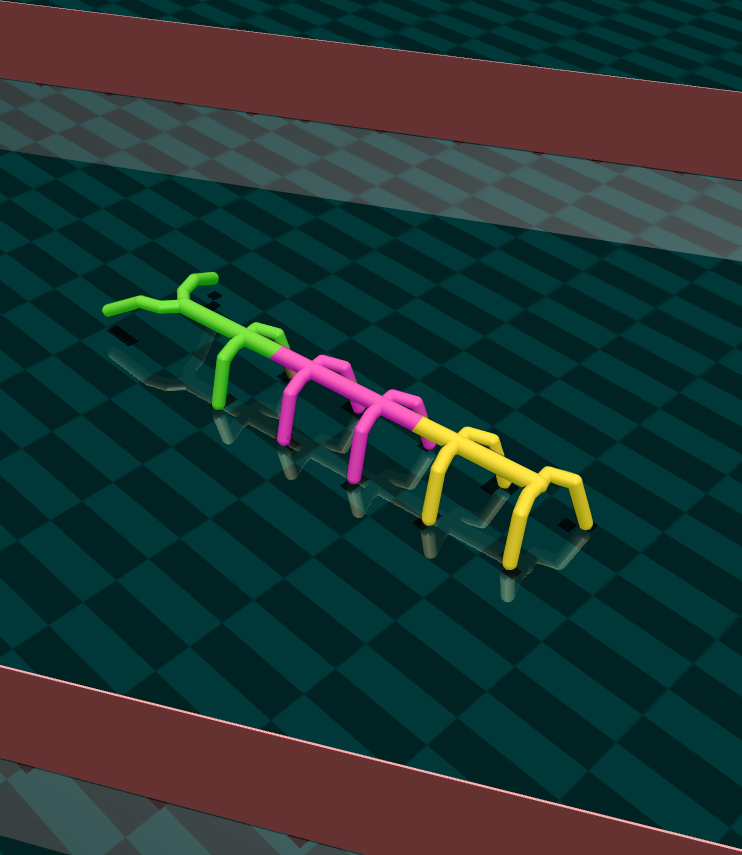}
}
\subcaptionbox{}
{
\includegraphics[width=0.21\linewidth]{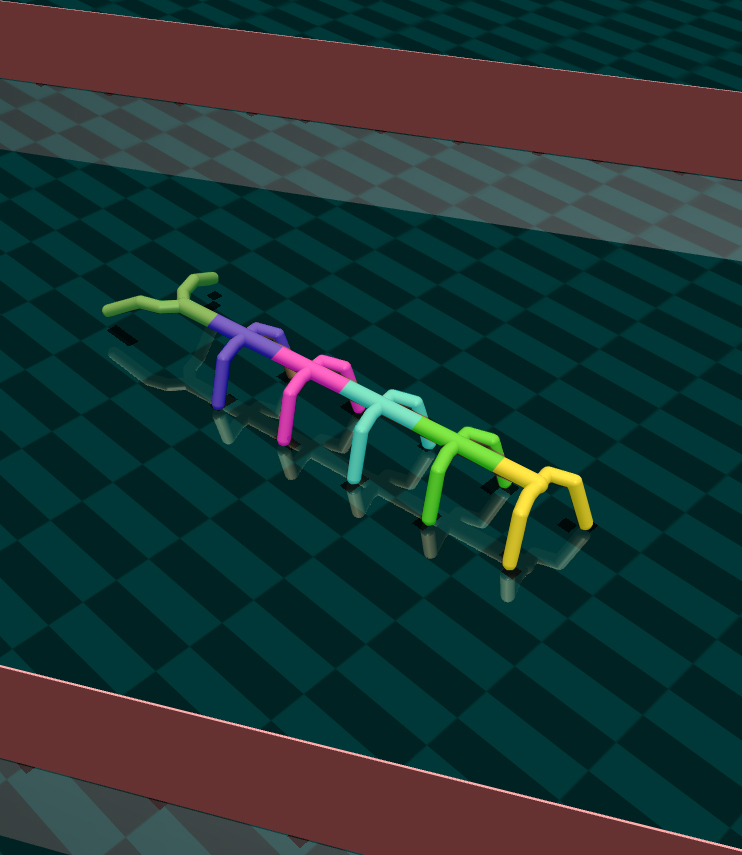}
}
\subcaptionbox{}
{
\includegraphics[width=0.222\linewidth]{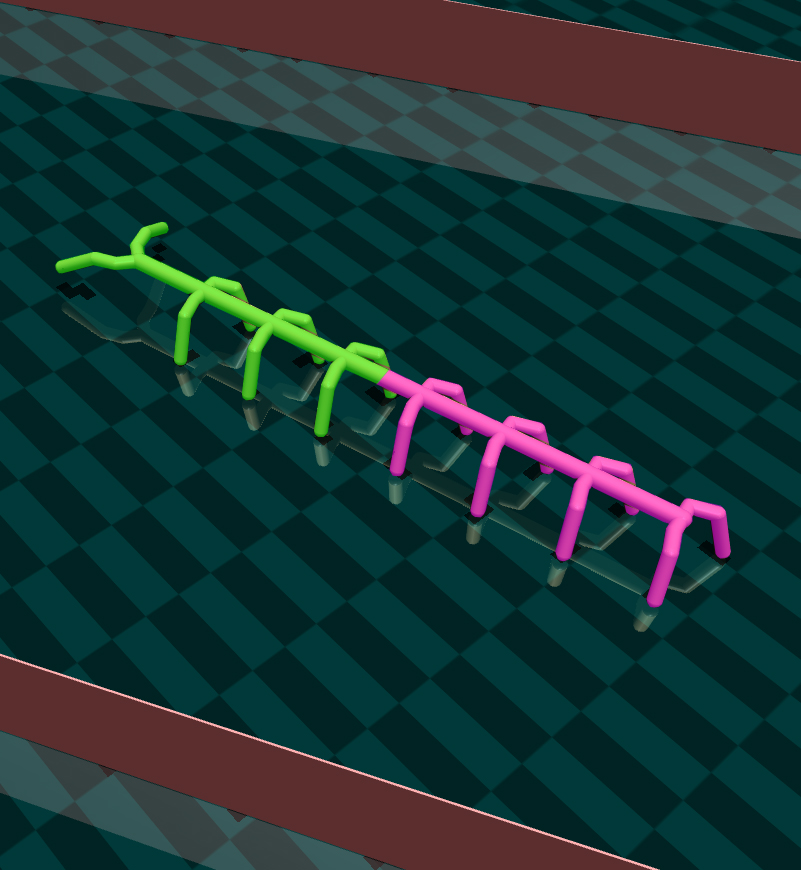}
}

\subcaptionbox{}
{
\includegraphics[width=0.194\linewidth]{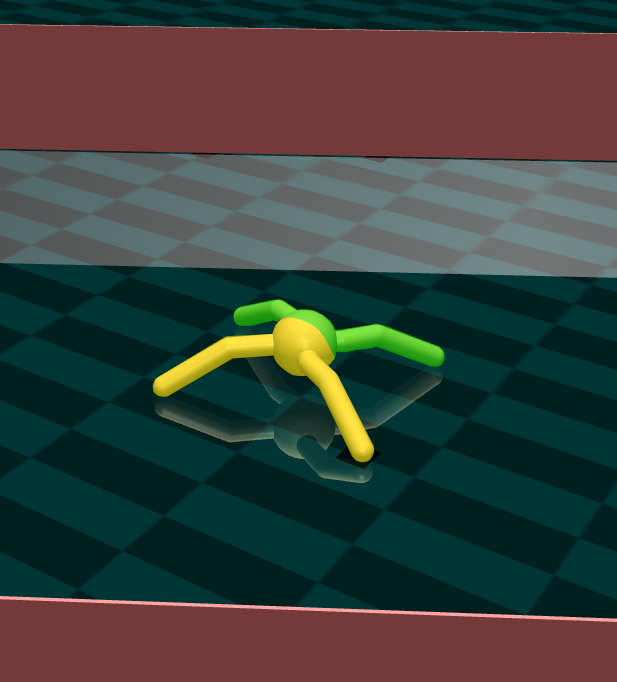}
}
\subcaptionbox{}
{
\includegraphics[width=0.239\linewidth]{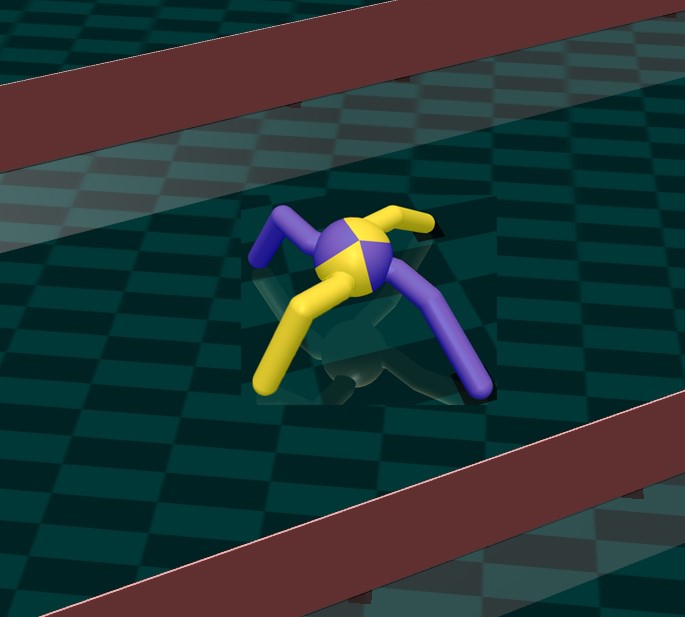}
}
\subcaptionbox{}
{
\includegraphics[width=0.205\linewidth]{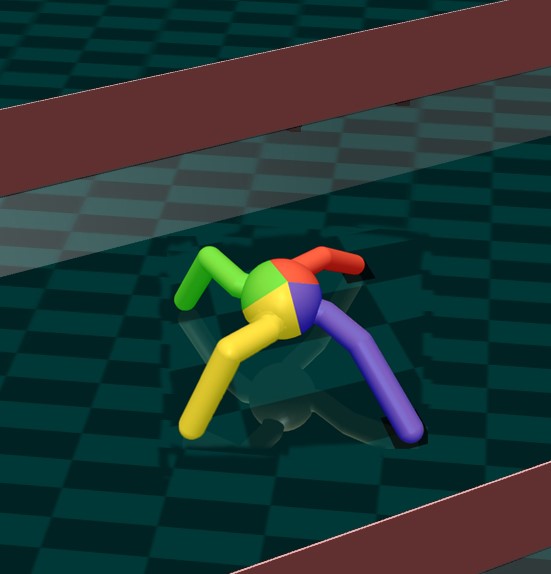}
}
\subcaptionbox{}
{
\includegraphics[width=0.225\linewidth]{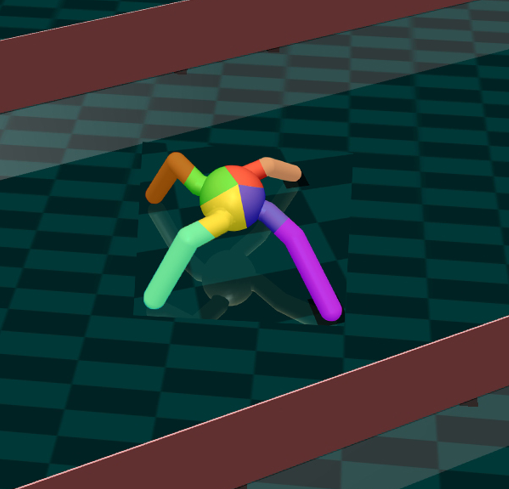}
}

\subcaptionbox{}
{
\includegraphics[width=0.233\linewidth]{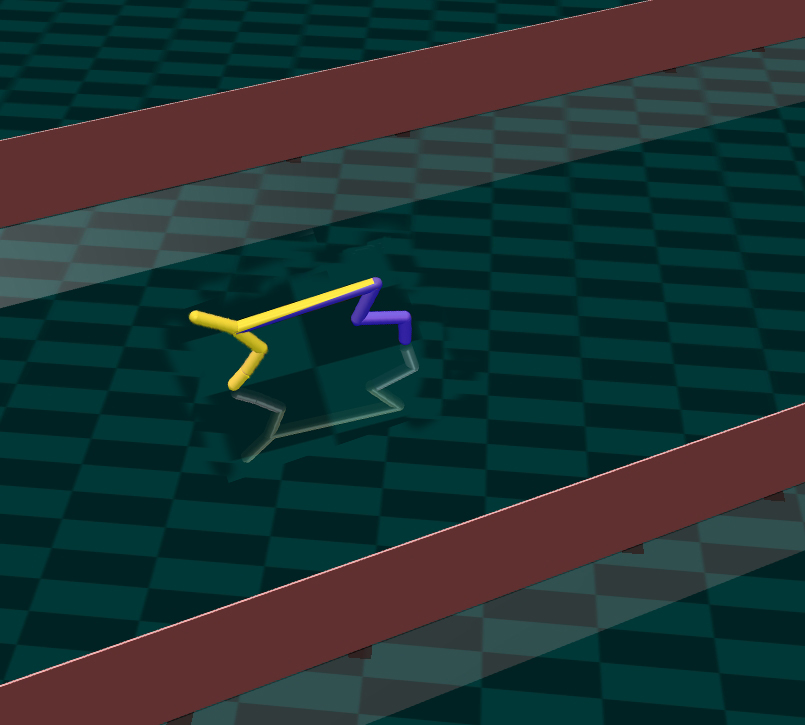}
}
\subcaptionbox{}
{
\includegraphics[width=0.205\linewidth]{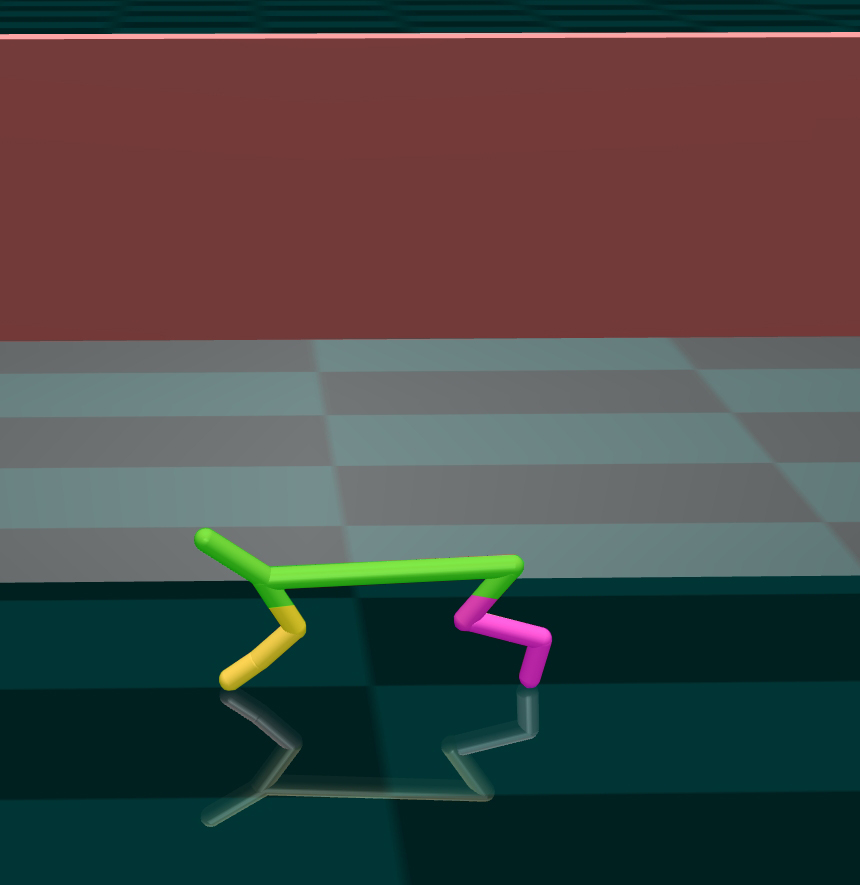}
}
\subcaptionbox{}
{
\includegraphics[width=0.205\linewidth]{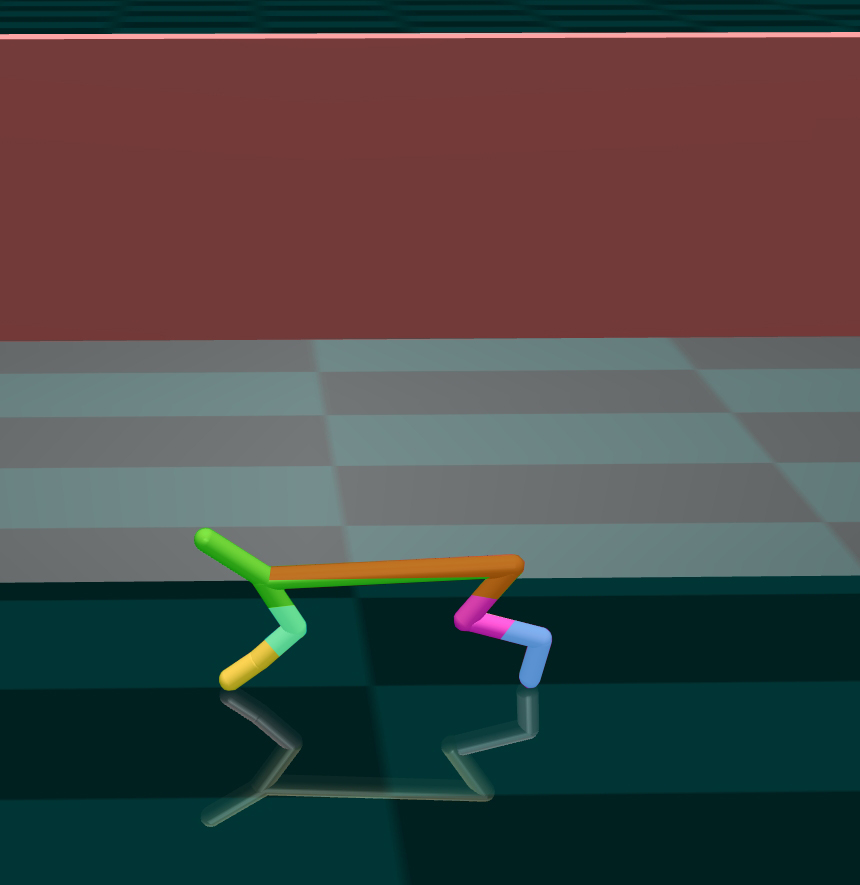}
}
\subcaptionbox{}
{
\includegraphics[width=0.218\linewidth]{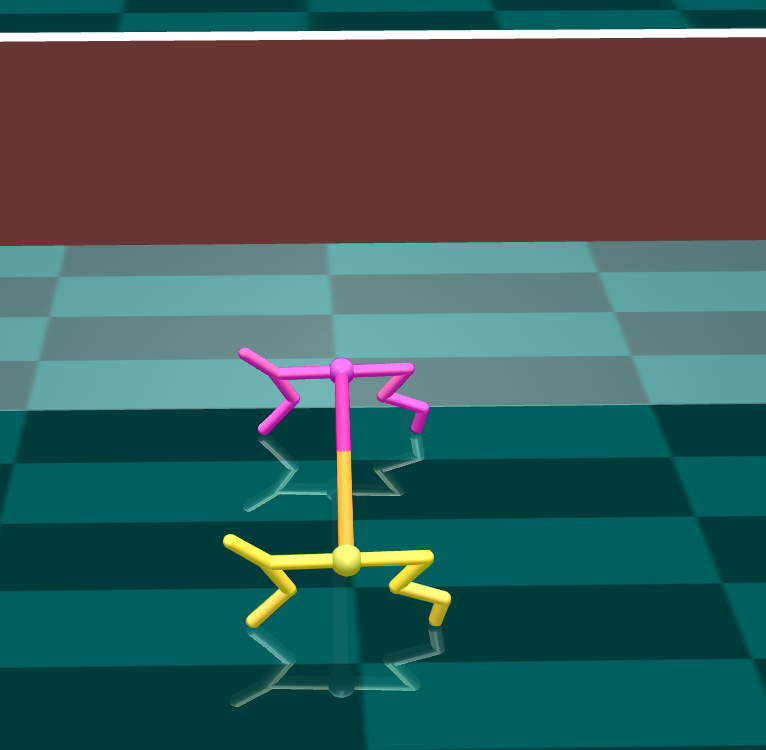}
}
    \vspace{-10pt}
 	\caption{\normalsize Example tasks in Safe Multi-Agent MuJoCo Environments, such as eight-agent Ant tasks and six-agent HalfCheetah tasks. Body parts of different colours are controlled by different agents. Agents jointly learn to manipulate the robot, while avoiding crashing into unsafe red areas, for details, see~\citep{gu2021multi} (Adapt the figures with permission from~\citep{gu2021multi}). 
 	} 
 	\label{fig:Safety-Environment-review-mamujoco}
 	\vspace{-10pt}
 \end{figure*}

\begin{figure*}[htbp!]
 \centering
 \subcaptionbox{ManyAgent Ant Task 1.0}
{
\includegraphics[width=0.4\linewidth]{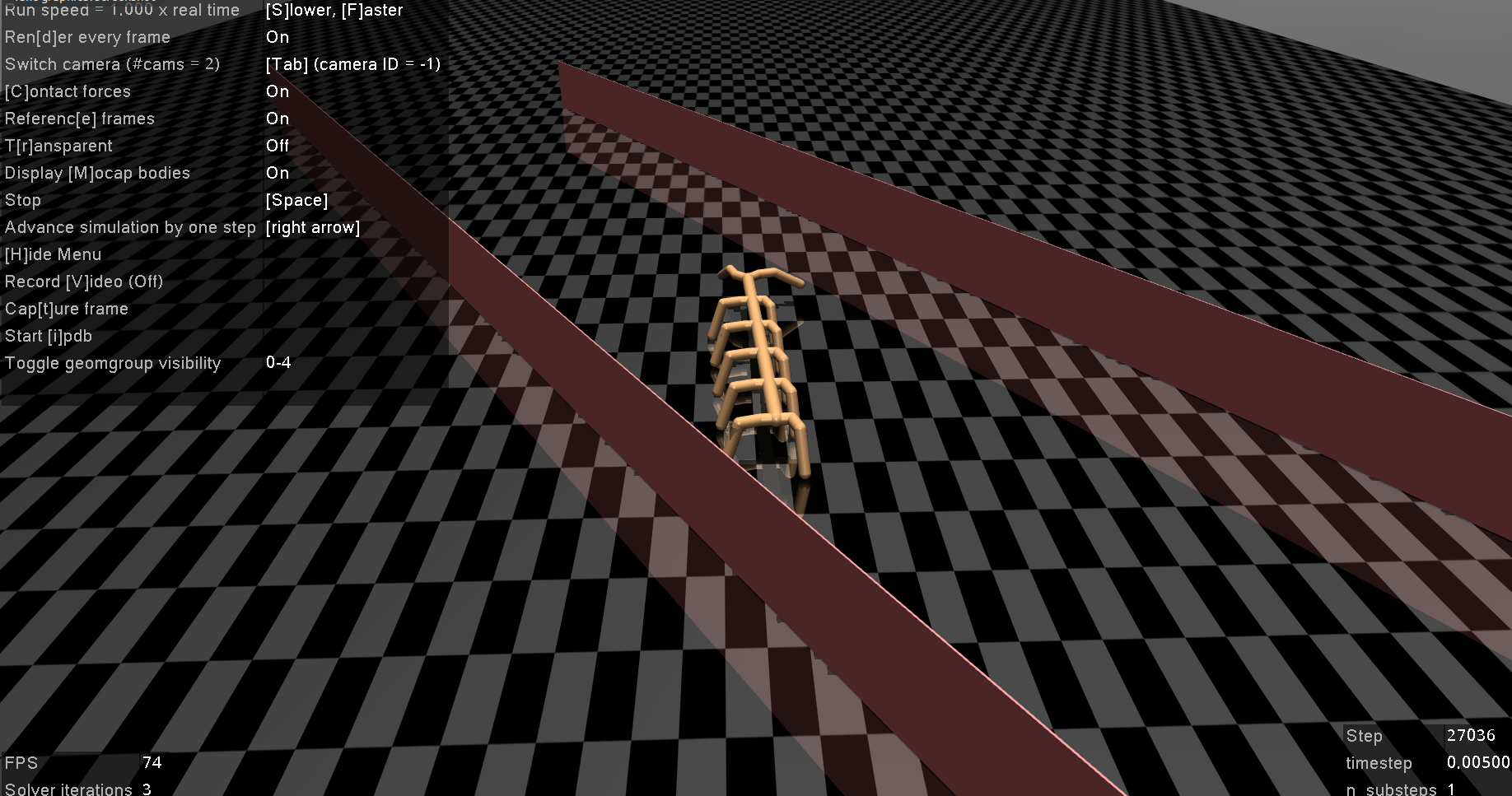}
}
 \subcaptionbox{Ant Task 1.0}
{
\includegraphics[width=0.4\linewidth]{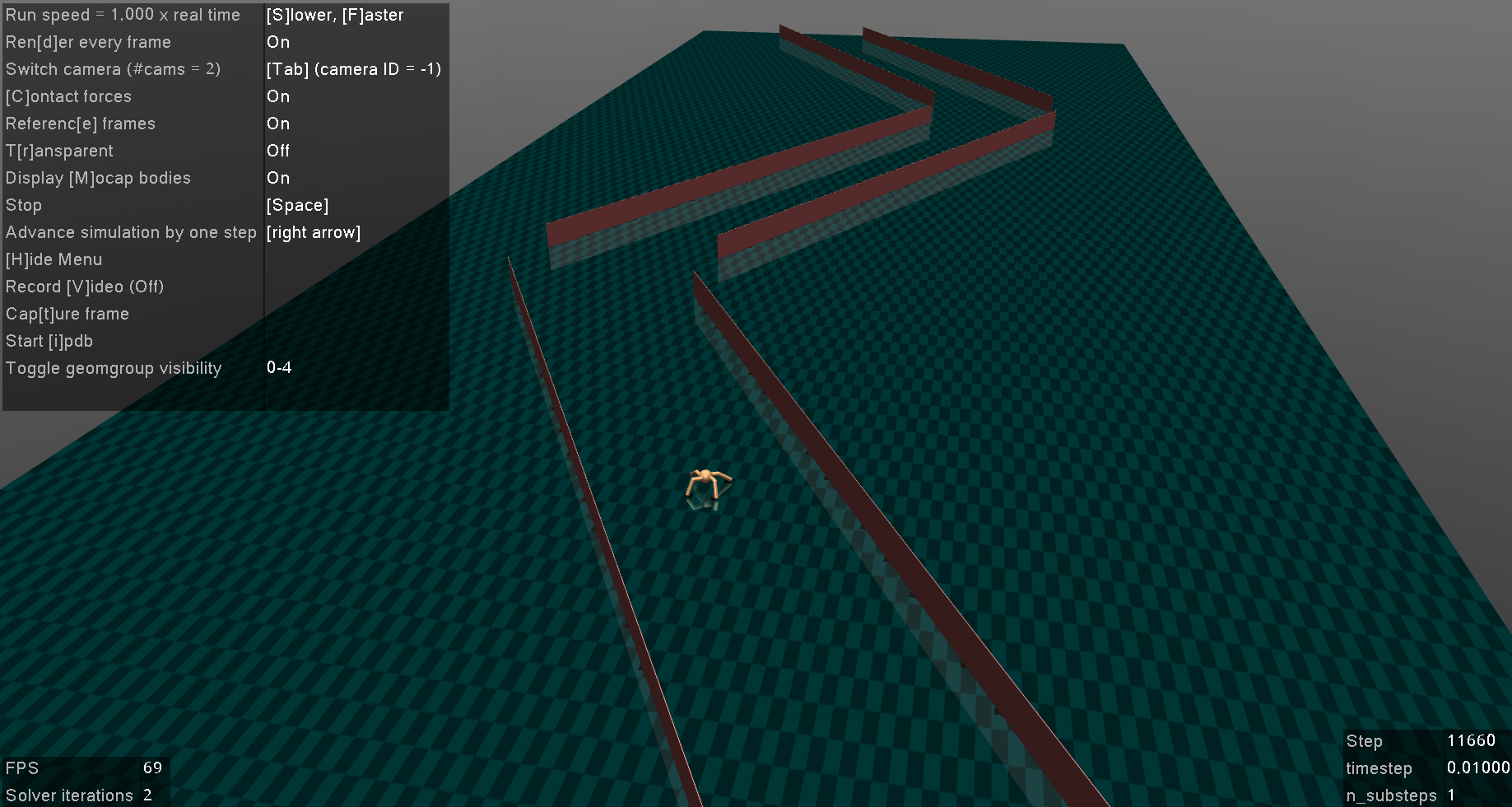}
}

 \subcaptionbox{HalfCheetah Task 1.0}
{
\includegraphics[width=0.3\linewidth]{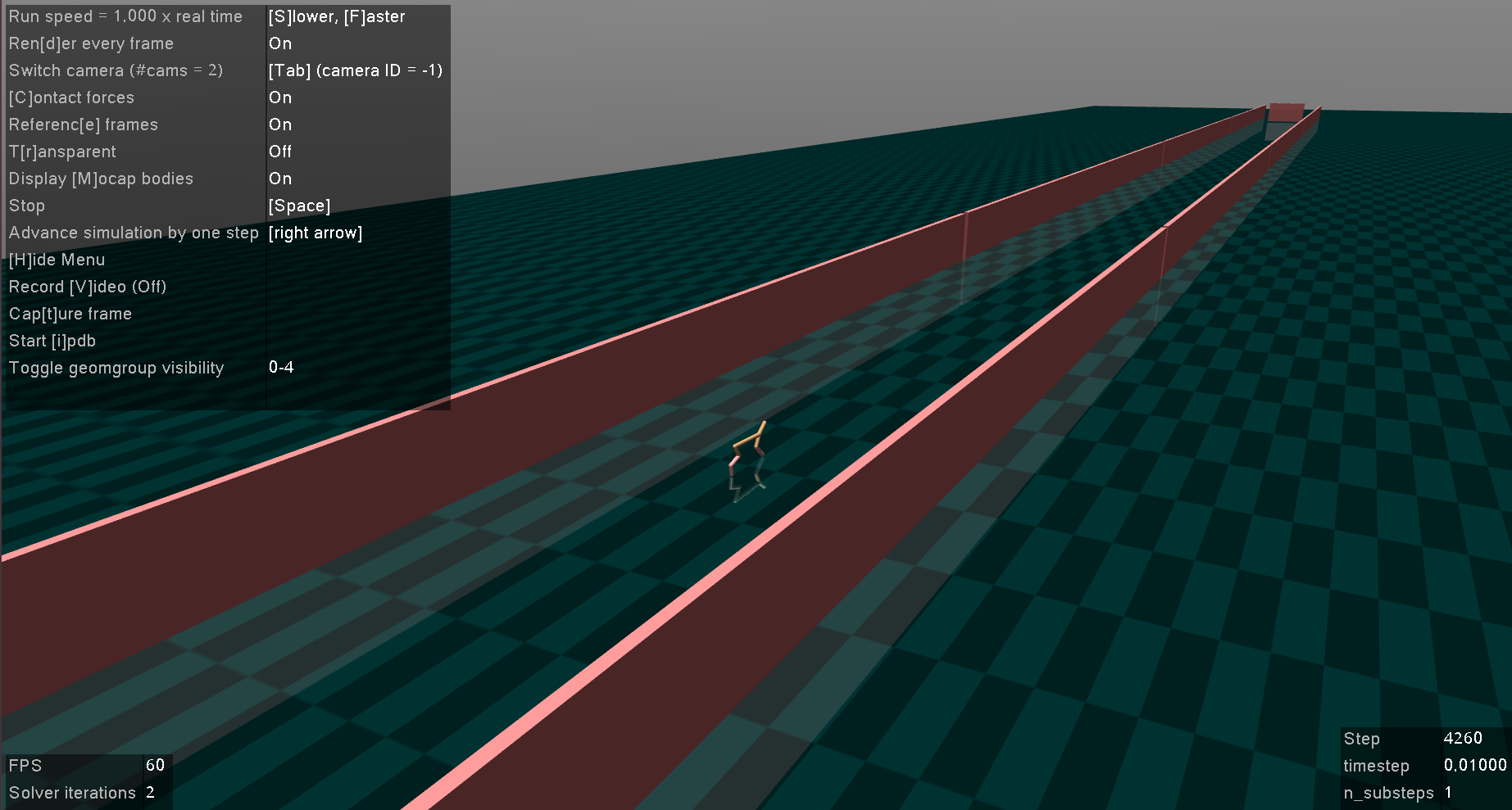}
}
 \subcaptionbox{Couple HalfCheetah Task 1.0}
{
\includegraphics[width=0.3\linewidth]{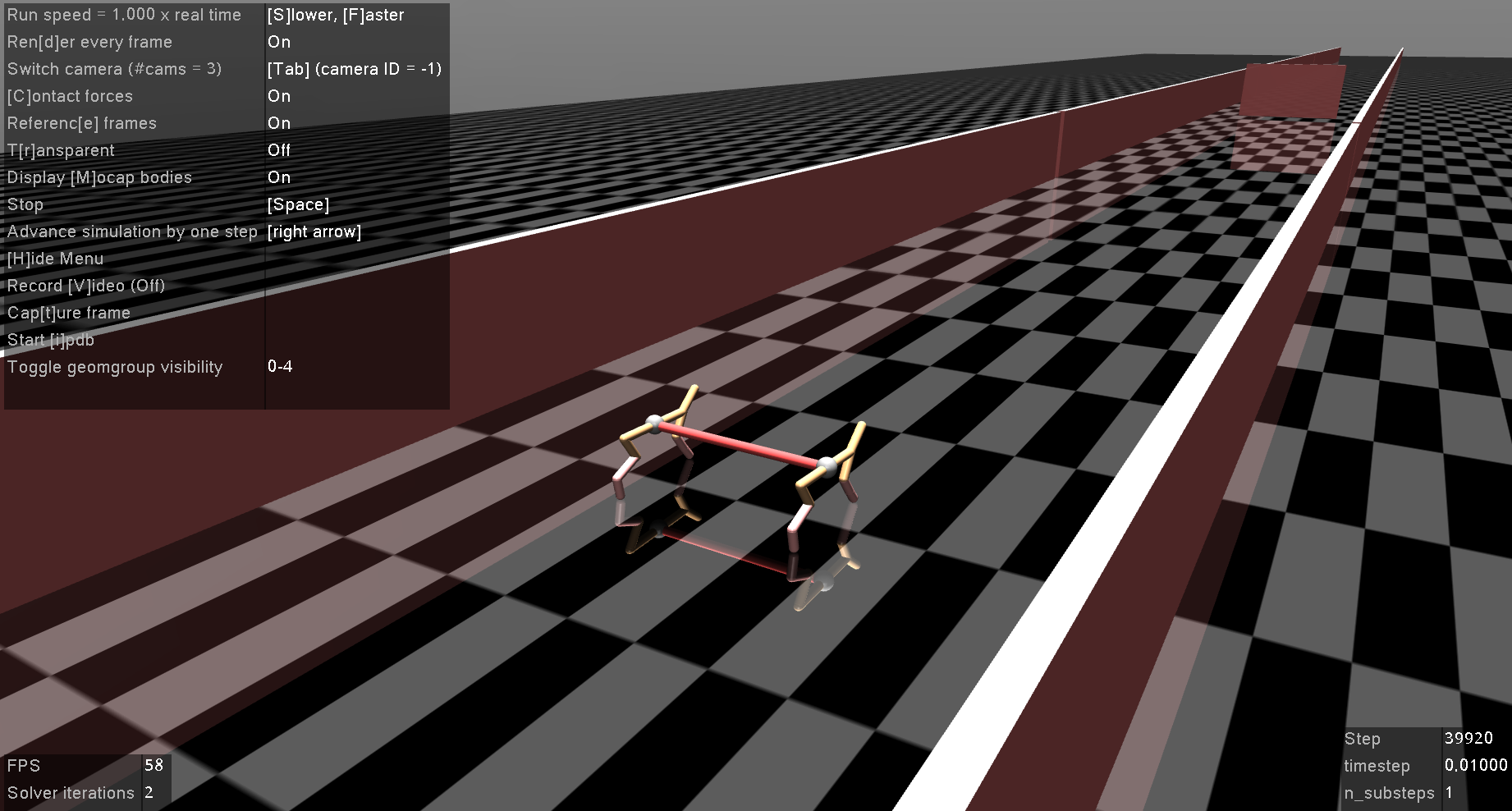}
}
 \subcaptionbox{ManyAgent Ant Task 2.1}
{
\includegraphics[width=0.3\linewidth]{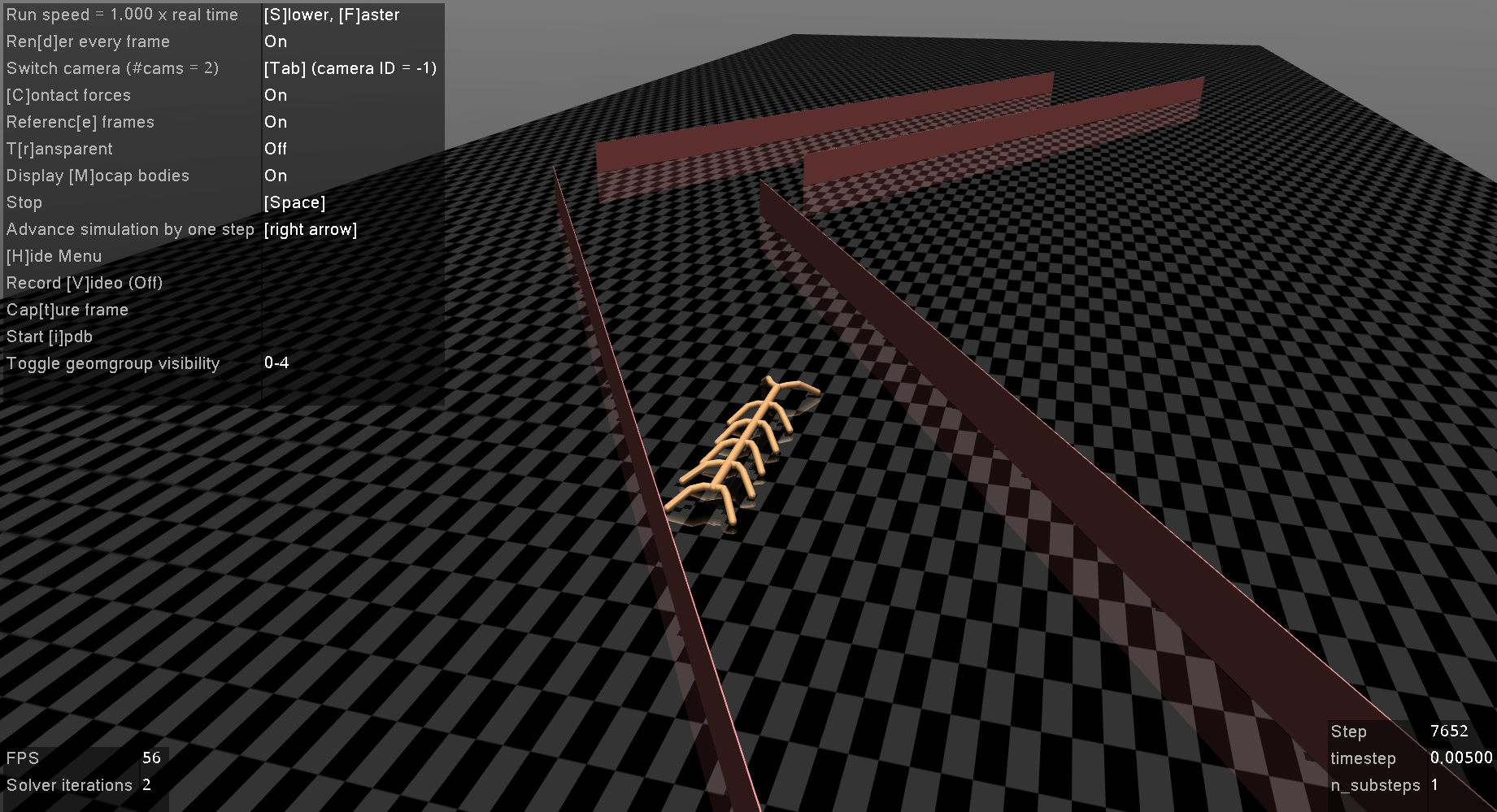}
}

 \subcaptionbox{ManyAgent Ant Task 2.2}
 {
\includegraphics[width=0.40\linewidth]{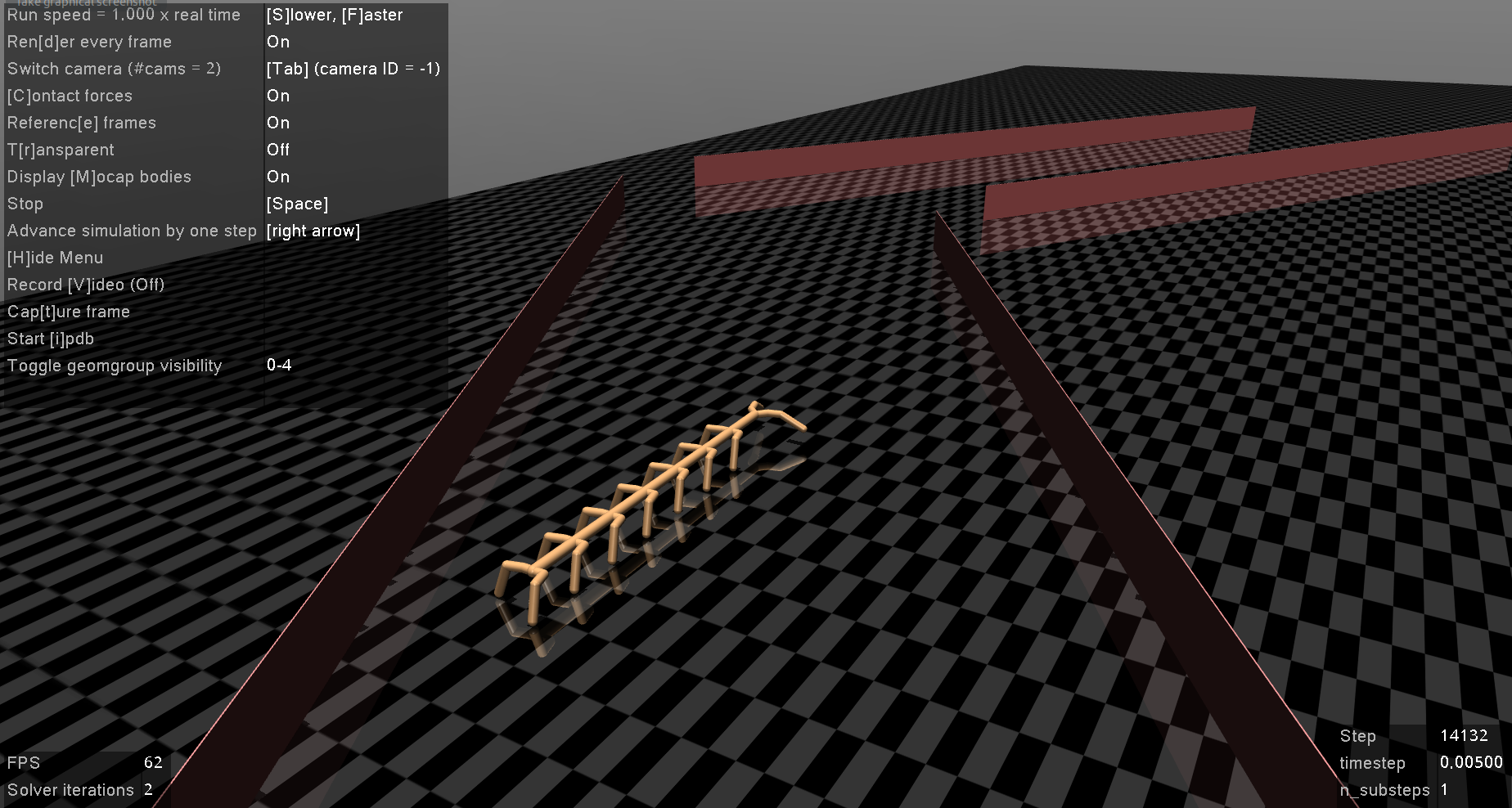}
}
\subcaptionbox{Ant Task 2.1}
 {
\includegraphics[width=0.30\linewidth, angle=0]{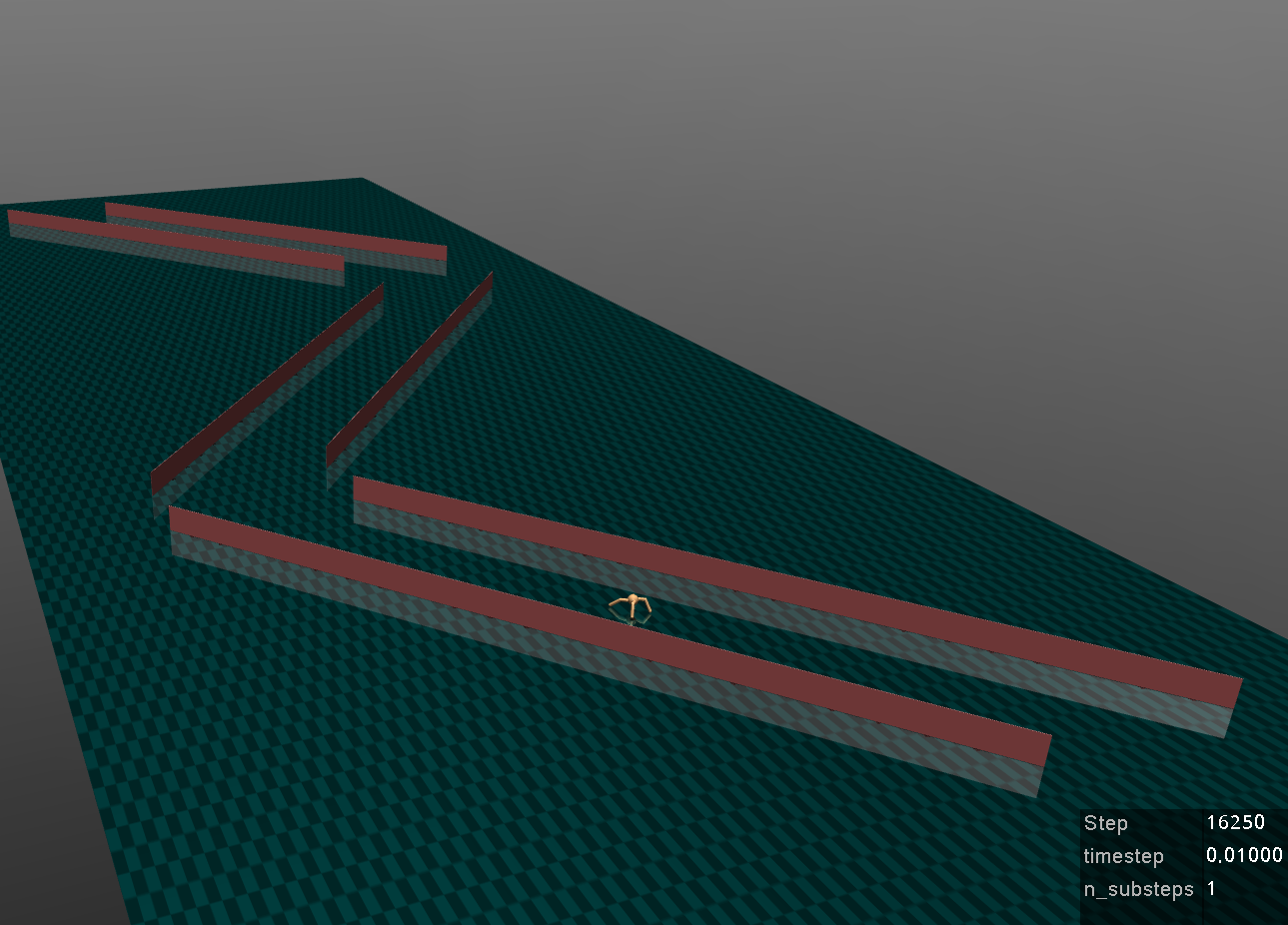}
}
\subcaptionbox{Ant Task 2.2}
 {
\includegraphics[width=0.20\linewidth, angle=0]{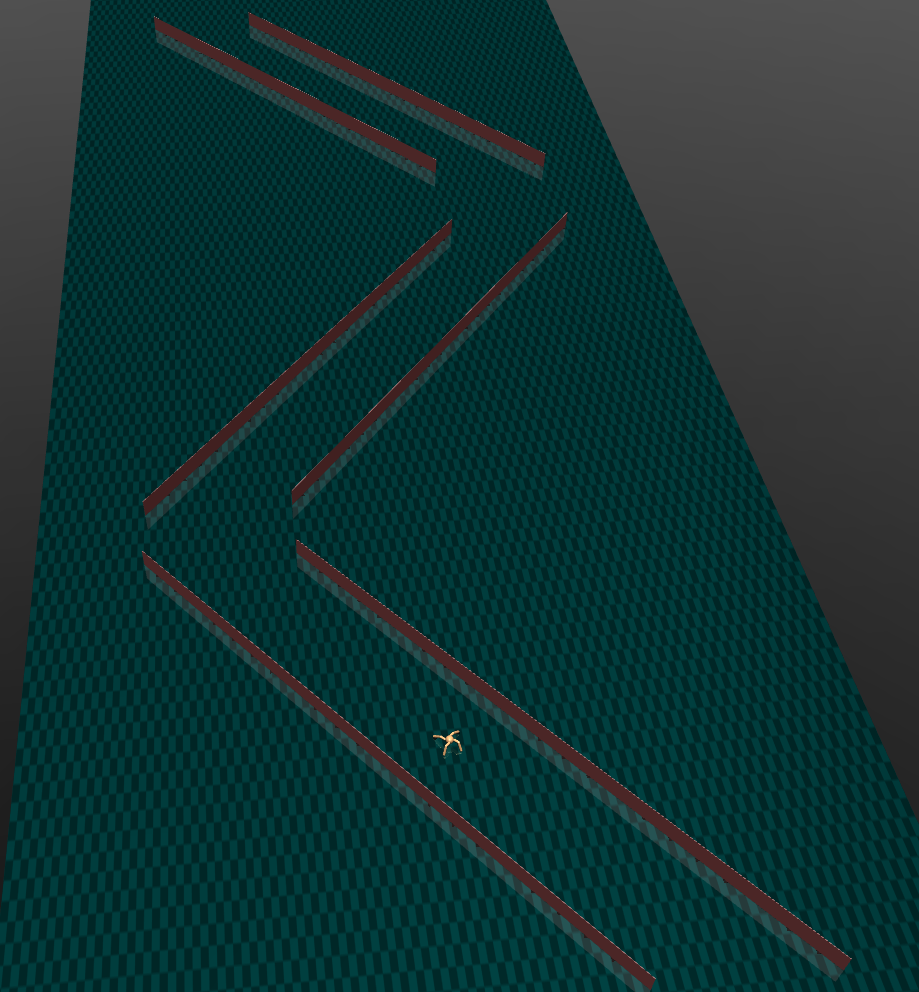}
}

    \vspace{-10pt}
 	\caption{\normalsize Specific tasks in Safe Multi-Agent MuJoCo Environments, e.g., 4x2-Ant Task 1.0, three folding jagged walls are Incorporated into the task; 3x2-ManyAgent Ant Task 2.1, two folding line walls is generated in the task. for details, see~\citep{gu2021multi}(Adapt the figures with permission from~\citep{gu2021multi}).}
 	\label{fig:Safety-mamujoco-Environment-specific}
 	\vspace{-10pt}
 \end{figure*}

 Safe MAMuJoCo~\citep{gu2021multi}~\footnote{\scriptsize \url{https://github.com/chauncygu/Safe-Multi-Agent-Mujoco.git}} is an extension of MAMuJoCo \citep{peng2020facmac}.  In Safe MAMuJoCo, safety-aware agents have to learn not only the skillful manipulations of a robot, but also how to avoid crashing into unsafe obstacles and positions. In particular, the background environment, agents, physics simulator, and reward function are preserved. However, unlike its predecessor, a Safe MAMuJoCo environment comes with obstacles like walls or bombs. Furthermore, with the increasing risk of an agent stumbling upon an obstacle, the environment emits cost \citep{brockman2016openai}. According to the scheme in \cite{zanger2021safe}, we characterize the cost functions for each task; the examples of Safe MAMuJoCo robots are shown in Figure \ref{fig:Safety-Environment-review-mamujoco} and the example tasks are shown in Figure \ref{fig:Safety-mamujoco-Environment-specific}.

\subsubsection{Safe Multi-Agent Robosuite}

\begin{figure*}[htbp!]
 \centering
\subcaptionbox{}
{
\includegraphics[width=0.302\linewidth]{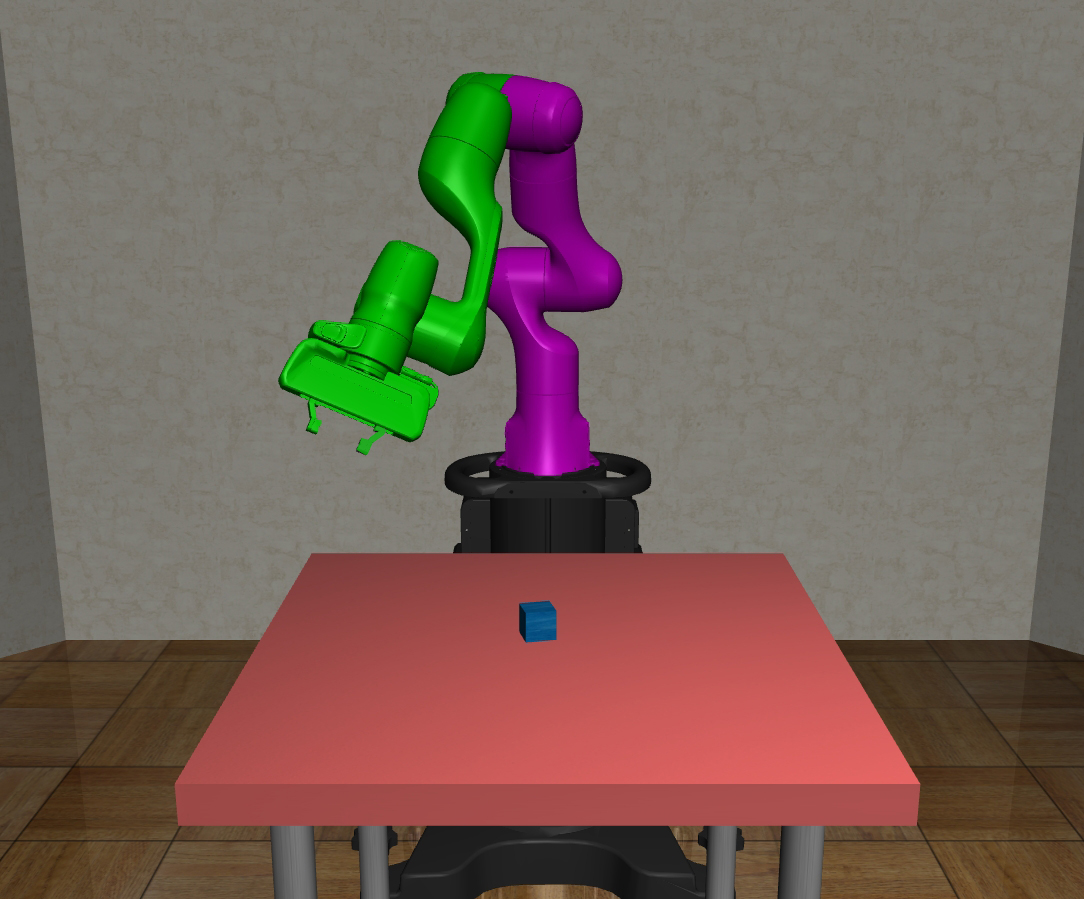}
}\quad \quad \quad
\subcaptionbox{}
{
\hspace{-27pt}
\includegraphics[width=0.273\linewidth]{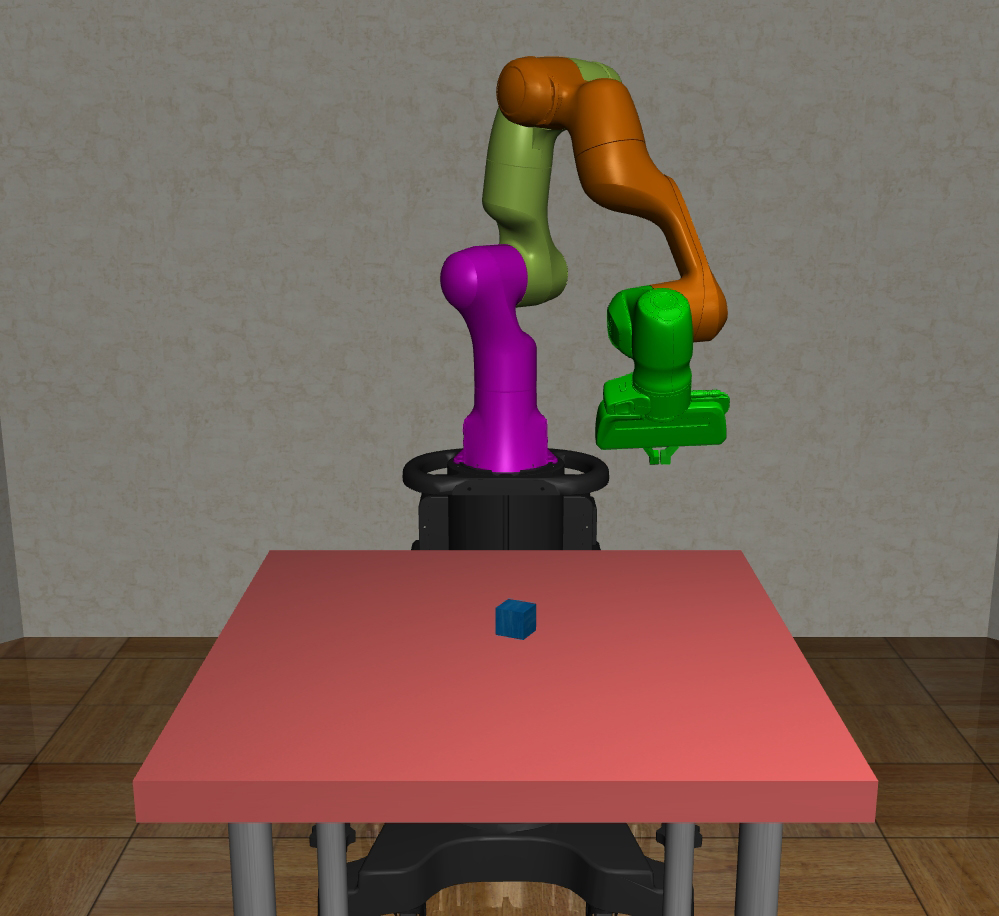}
}\quad \quad \quad
\subcaptionbox{}
{
\hspace{-27pt}
\includegraphics[width=0.302\linewidth]{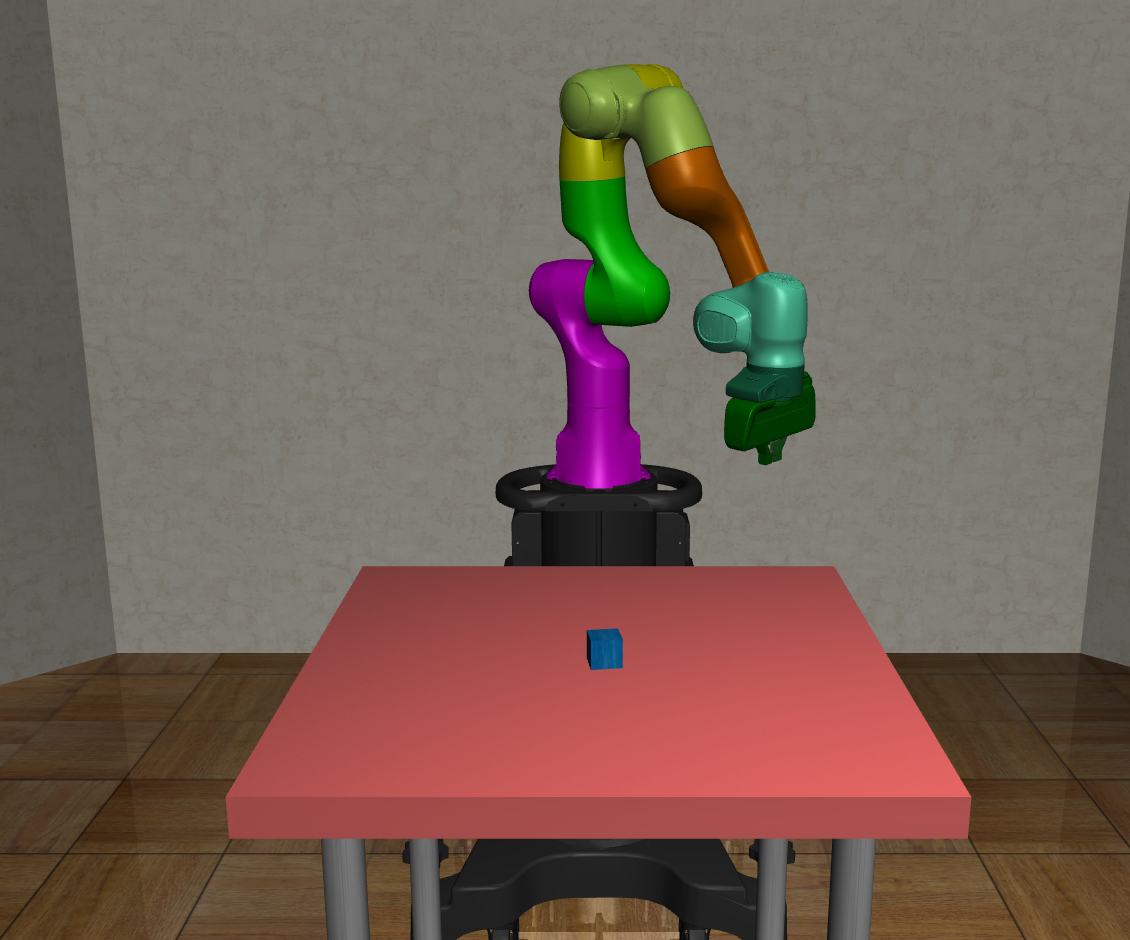}
}

\subcaptionbox{}
{
\includegraphics[width=0.28\linewidth]{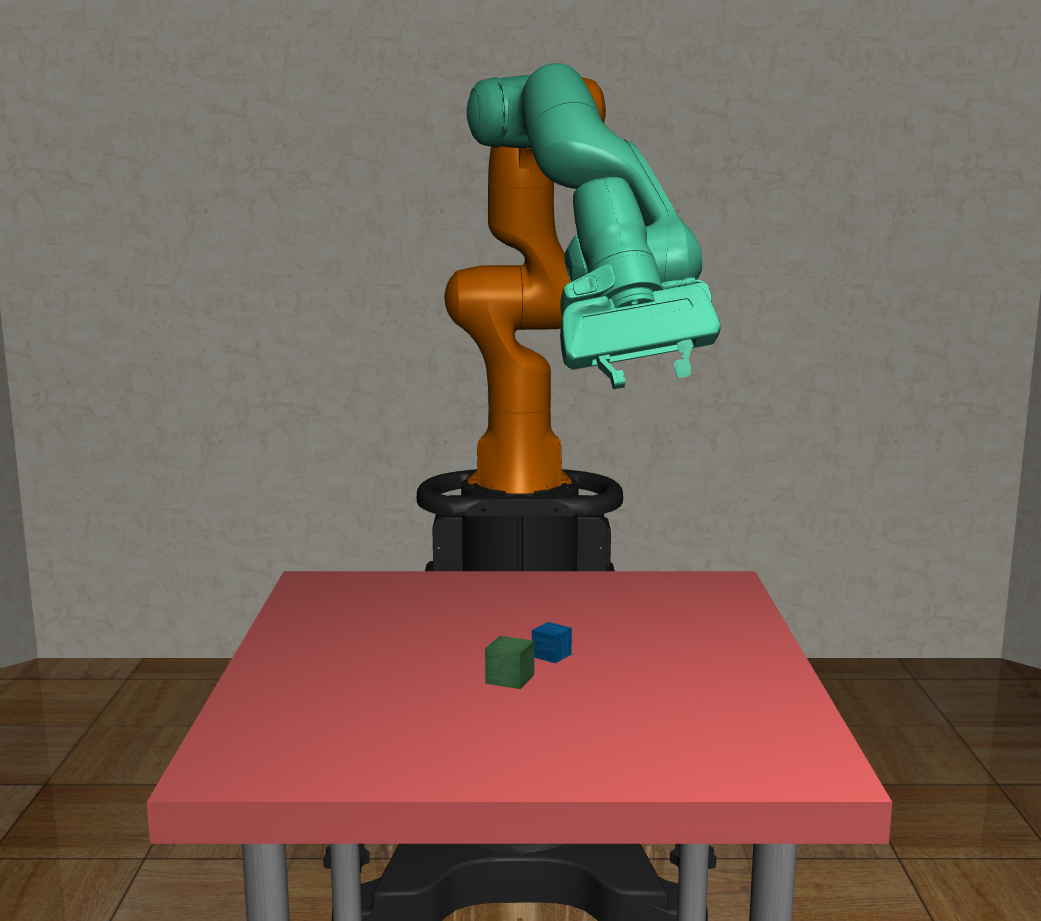}
}\quad \quad \quad
\subcaptionbox{}
{
\hspace{-30pt}
\includegraphics[width=0.31\linewidth]{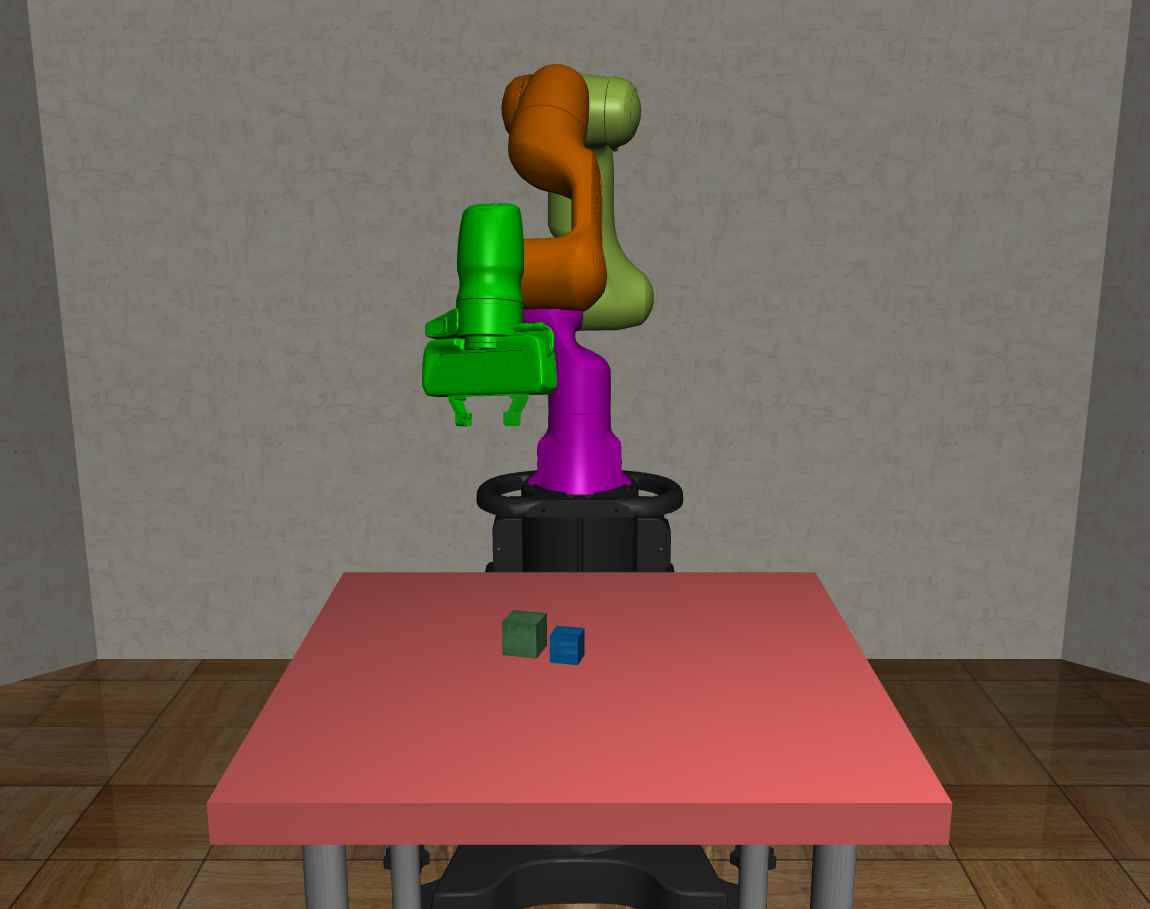}
}\quad \quad \quad
\subcaptionbox{}
{
\hspace{-30pt}
\includegraphics[width=0.295\linewidth]{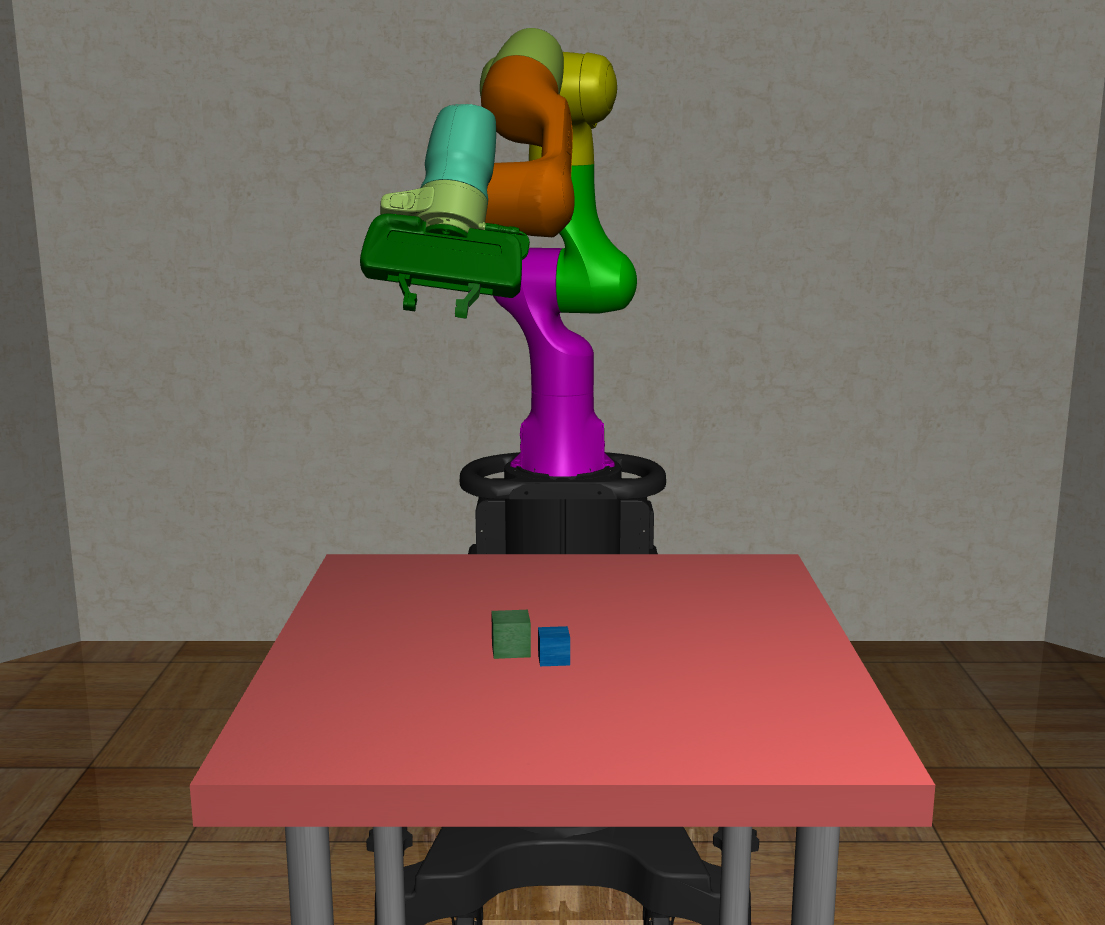}
}
\subcaptionbox{}
{
\includegraphics[width=0.387\linewidth]{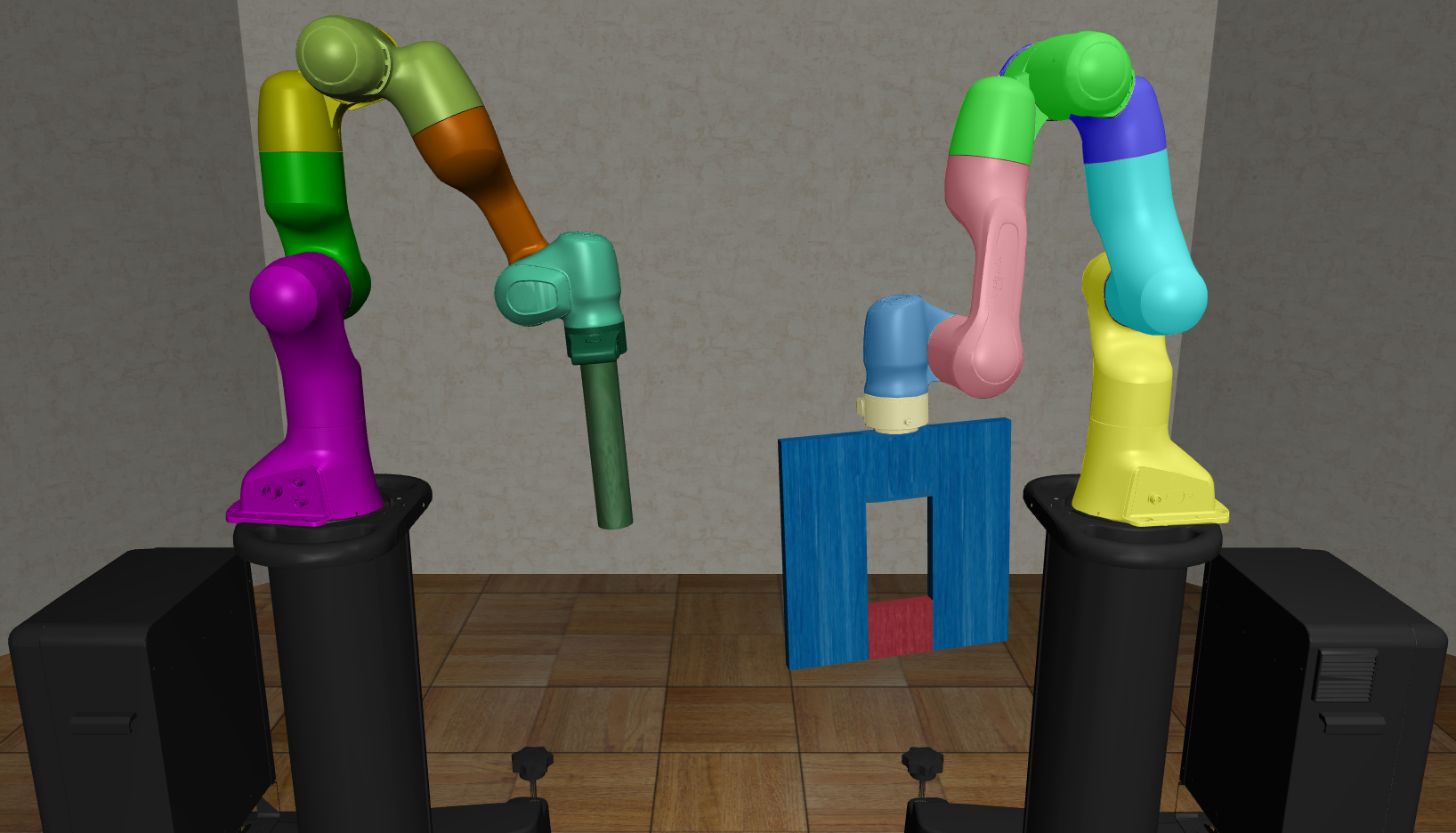}
}
\subcaptionbox{}
{
\includegraphics[width=0.376\linewidth]{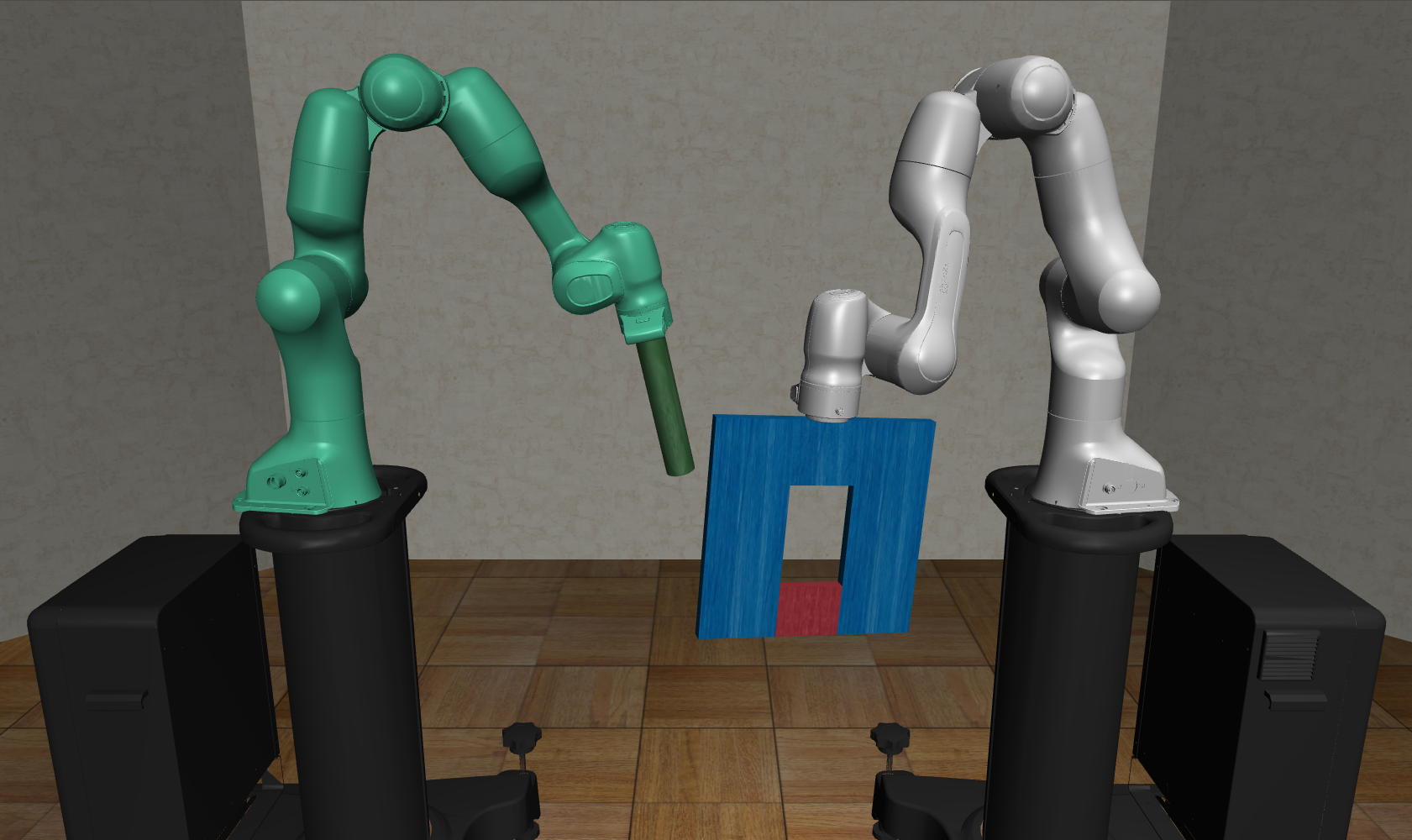}
}
 	\caption{\normalsize Example tasks in Safe MARobosuite Environments, e.g., two-agent Lift tasks, eight-agent Lift tasks and fourteen-agent TwoArmPegInHole tasks. Agents jointly learn to manipulate the robot, while avoiding crashing into unsafe  areas, for details, see~\citep{gu2021multi}(Adapt the figures with permission from~\citep{gu2021multi}). 
 	} 
 	\label{appendix-fig:Safety-robosuite-lift-twoarmpeginhole}
 \end{figure*}

Safe Multi-Agent Robotsuite (Safe MARobosuite)~\citep{gu2021multi}~\footnote{\scriptsize \url{https://github.com/chauncygu/Safe-Multi-Agent-Robosuite.git}}, shown in Figure \ref{appendix-fig:Safety-robosuite-lift-twoarmpeginhole}, has been developed on the basis of Robosuite~\citep{zhu2020robosuite} which is a popular robotic arm benchmark for single-agent reinforcement learning. In Safe MARobosuite, multiple agents are set up according to the robot joint settings, and each agent controls every joint or several joints. A Lift task, for example, can be divided up into 2 agents (2x4 Lift), 4 agents (4x2 Lift), 8 agents (8x1 Lift); for Stack tasks, similarly, tasks can be provided with 2 agents (2x4 Stack), 4 agents (4x2 Stack), 8 agents (8x1 Stack); for a two-robot TwoArmLift task, it can be divided up into: 2 agents (2x8 TwoArmLift), 4 agents (4x4 TwoArmLift), 8 agents (8x2 TwoArmLift), 16 agents (16x1 TwoArmLift).


Moreover, Safe MARobosuite is also a fully cooperative, continuous, and decentralized benchmark that considers the constraints of robot safety. Where  Franka robots are used to conduct each task, each agent can observe partial environmental information (such as the velocity and position). More importantly, Safe MARobosuite can be easily used for modular robots and make robots have good robustness and scalability. In many real-world applications, modular robots can be quickly paired and assembled for different tasks  \citep{althoff2019effortless}. For instance, when communication bandwidth is limited, or some joints of robotic arms are broken, leading to causes malfunctioning communication, modular robots can still work well. The reward setting is the same as Robosuite \cite{zhu2020robosuite}, and the cost design is used to prevent robots from crashing into unsafe areas.

\subsubsection{Safe Multi-Agent Isaac Gym}
\begin{figure*}[htbp!]
 \centering
{
 \subcaptionbox{ShadowHandOver 2x6}{
\includegraphics[width=0.48\linewidth]{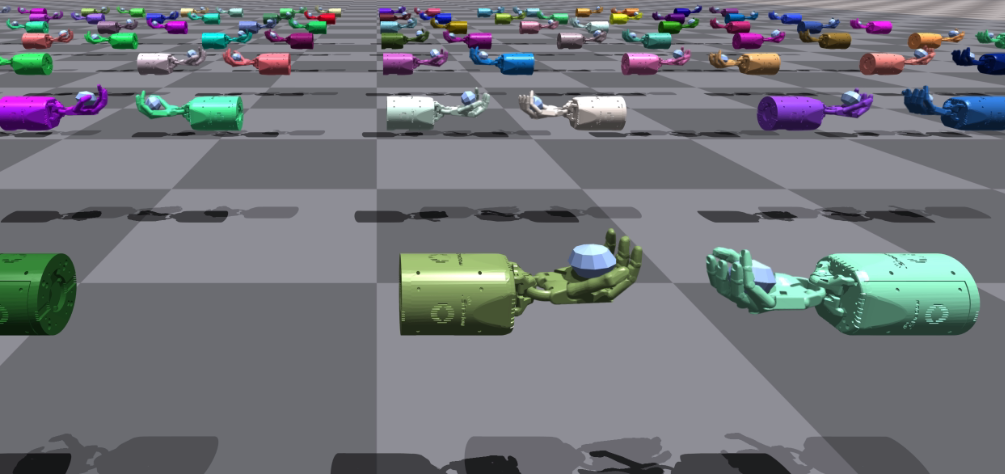}
} 
\subcaptionbox{ShadowHandCatchUnderarm 2x6}{\includegraphics[width=0.455\linewidth]{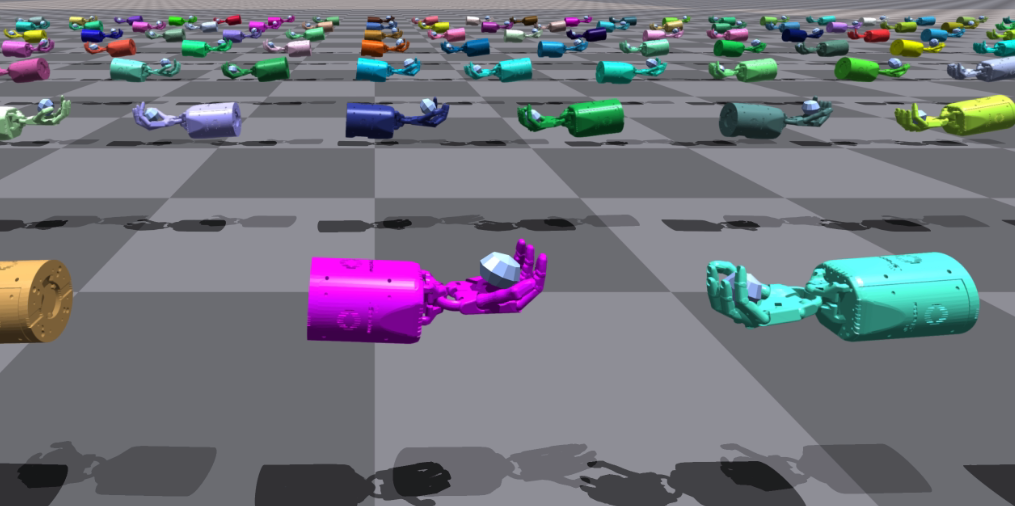}
}

 \subcaptionbox{ShadowHandReOrientation 4x3}
 {\hspace{-12pt}
\includegraphics[width=0.48\linewidth]{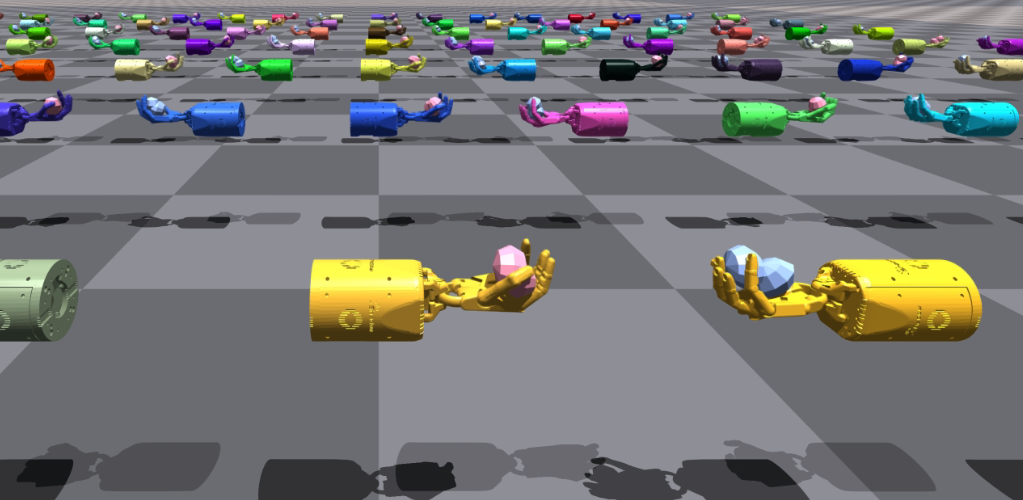} 
}
\subcaptionbox{A hand showing safety constraints}{\includegraphics[width=0.455\linewidth]{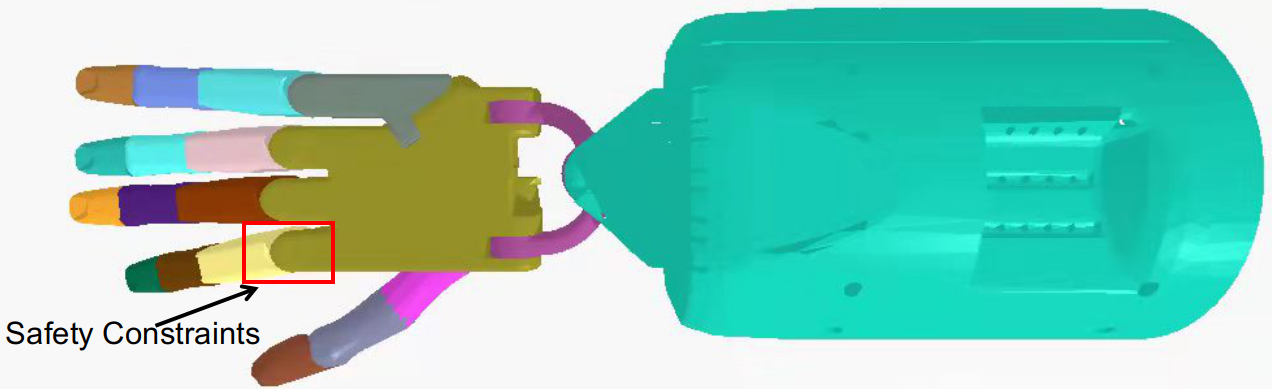} \hspace{-15pt}
}

}
    \vspace{-0pt}
 	\caption{\normalsize Safe multi-agent Isaac Gym environments. Different robot body parts of different colours  are manipulated
by different agents in  environments. Agents jointly learn how to control the robot, whilst the safety constraints are not violated (Adapt the figures with author permission).}

 \label{fig:safe-multi-agent-isaac-gym}
 \end{figure*}

Safe Multi-Agent Isaac Gym (Safe MAIG)~\footnote{\scriptsize \url{https://github.com/chauncygu/Safe-Multi-Agent-Isaac-Gym.git}} is a high-performance environment that uses GPU for trajectory sampling and logical computation based on Isaac Gym~\citep{makoviychuk2021isaac}, see Figure~\ref{fig:safe-multi-agent-isaac-gym}. The computation speed of Safe MAIG is almost ten times that of Safe MAuJoCo and Safe MARobosuite on the same server, in the same task, using the same algorithm. The communication speed is also better than Safe MAMuJoCo and Safe MARobosuite. But the memory might need to be optimized between CPU and GPU, since GPU memory is generally too small.

\section{Challenges and Outlook of Safe Reinforcement Learning}
\label{section:Challenges-safe-DRL}

\hl{When deploying RL in real-world applications, numerous challenges emerge throughout the implementation process.} In this section, the \textbf{``Safety Challenges"} problem is investigated by proposing several significant challenges we need to address \hl{moving forward. Additionally, we identify potential research directions for enhancing the safety of RL systems.} Except for the following challenges and research directions, we will investigate meta-safe RL \citep{khattar2022cmdp} and robust RL \citep{gao2022resilient, pang2021robust} in the future.

\subsection{Human-Compatible Safe Reinforcement Learning}

Modern safe RL algorithms rely on humans to state the objectives of safety.     
While humans understand the dangers, the potential risks are less noticed. Below we discuss two challenges in human preference statements and concerns on ethics and morality.
\begin{itemize}
    \item \textbf{Human preference statement.}
Many challenges are posed as AI agents are frequently evaluated in terms of performance measures, such as human-stated rewards. 
       On the one hand, \hl{while it is usually assumed that humans are acting honestly in specifying their preferences, such as by rewards or demonstrations, the consequence of humans misstating their objectives is commonly underestimated. Humans may maliciously or unintentionally misstate their preferences,} leading the safe RL agent to perform unexpected implementations. One example is the Tay chatbot from Microsoft; prankster users falsify their demonstrations and train Tay to mix racist comments into its dialogue \cite{fickinger2020multiprincipal}.
       On the other hand, multiple humans might be involved in training one safe RL agent. Thus, agents have to learn to strike a balance between the widely different human preferences. Earlier attempt \cite{hadfield2016cooperative} considers one agent vs one human scenario. \hl{However, many open questions remain, such as training robust agents against malicious users, personalizing assistance toward human preferences, etc.}

    \item \textbf{Ethics and morality concerns.} 
In modern society, the human interrelationship is built based on social or moral norms. 
While reinforcement learning agents are deployed to the real world, they start having impacts on each other, turning into a multi-agent system, in which norms act similarly in human society on agents. Therefore, the decisions made by agents always involve ethical issues.
For example, \hl{social dilemmas will emerge from the relation between individual goals and overall interests~\cite{hughes2018inequity,leibo2017multi}.} The conflicts are produced when each agent aims to maximize its benefit. 
For another example, consider a \textit{trolley problem} \cite{foot1967problem}. When the agent is faced with the choice of either harming multiple people on the current track or one person by diverting the train, what would you expect the safe agents to choose?
\hl{Or, more realistically, when the driving agent is about to bump into a lorry, it can swerve off the road to the left to save itself. However, there is a bike on the left.}
How could the driving agent make decisions? A human driver's knee-jerk reaction might be swerving left to save itself. \hl{However, the decision of the driving agent depends on its value systems.} 
How to leverage the different value systems to enable safe agents to make ethical decisions is an open question.

\end{itemize}

\subsection{Industrial Deployment Standards for Safe Reinforcement Learning}

Although safe RL has been developed with a wealth of well-understood methods and algorithms in recent years~\cite{garcia2015comprehensive, kim2020safe, liu2021policy}, to our knowledge, there is no RL deployment standard for industrial applications, including technical standards and morality standards, etc. \hl{More attention is required on deployment standards and the alignment between academia and industries}. Applications include robotics, autonomous driving, recommendation systems, finance, \textit{et al}. We should tailor specific deployment standards to specific applications.
\begin{itemize}
    \item \textbf{Technical standards.} In a technical standard, we need to think about how much efficiency RL can generate, how much time and money can be saved using RL methods, what environments can be handled with RL, how to design cost and reward functions considering the balance between RL reward, performance and safety, etc.
   
    \item \textbf{Law standards.} Human-machine interaction needs to be considered in legal judgments, for example, when robots hurt humans due to programming errors using RL methods. Furthermore, we need to determine how responsibilities are divided, e.g., do programmers of robots need to take more responsibility, or do robot users need to take more responsibility?
\end{itemize}

\subsection{Safety Guarantee Efficiently in Large Number of Agents' Environments}

Since the decision-making space is incredibly large when the number of agents increases in a safe MARL setting, it is not easy to optimize the multi-agent policy to finish tasks safely~\citep{yang2020overview, zhang2021multi}. Thus, efficiently guaranteeing safety in an environment with a large number of agents, e.g., 1 million agents, is still a challenging problem.

\begin{itemize}
    \item \textbf{Theory of safe  MARL.} \hl{Theory and experiments of safe MARL for massive swarm robots should be considered in the future, e.g., the convergence problem remains open for massive safe MARL methods without strong assumptions.}
    \hl{Furthermore, optimizing sample complexity and stability within safe MARL environments is essential. Specifically,} greater emphasis should be placed on the following key points regarding the theory of safe MARL.
    (1). Credit assignment both in cost and reward. 
In cooperative, competitive, and mixed game settings,  it is crucial to contemplate the trade-off between reward and safety performance, e.g., a policy should be searched to minimize each agent's cost value while improving reward. Furthermore, it is necessary to optimize the precise cost value for each agent, and consider each agent's cost credit.    
  (2). Nonstability. In a multi-agent system, when an agent takes actions, which will influence other agents' actions and may make other agents get worse reward value or unsafe.
  (3). Scalability. In safe MARL settings, when the number of agents becomes large, such as one billion agents, it will be challenging for computation and hard to ensure agents' safety, \hl{since it is almost impossible to estimate each agent's Q values or V value simultaneously.}

    \item \textbf{Multiple heterogeneous constraints.} It still needs to be determined how multiple heterogeneous constraints should be handled in multi-agent systems for safe MARL. To our knowledge, almost no study has investigated multiple heterogeneous constraints for MARL. For example, when different agents encounter different heterogeneous constraints, we need to study how to balance different heterogeneous constraints to optimize different agents' policies for safety while improving reward values.  
    \item \textbf{Carry out complex multi-agent tasks safely.}
     For instance, swarm robots perform the encirclement task, and then rescue the hostages from the dangerous scene safely. Currently, most safe MARL algorithms could have some challenging issues while conducting complex tasks, e.g., time-consuming issues or convergence issues.
    \item \textbf{Robust MARL.} One of the concerns in robust MARL settings: \hl{ensuring zero cost in different tasks without tuning parameters using the same algorithms is still open~\cite{hendrycks2021unsolved}.}  \hl{Another concern is when we transfer the safe MARL simulation results to real-world applications, it is still unclear regarding how to handle the mismatch between the nominal and real system,} since there may be the sensor noise~\cite{kamran2021minimizing} generated by fault sensors or disturbing sensor information transferred by adversary attacks.
    \item \textbf{Trade-off Balances.} The trade-off balance between exploration and exploitation is a dilemma in RL or MARL. \hl{Safe RL or safe MARL has the same problem.} More particularly, there is another dilemma that is the trade-off between reward and cost, which is different from the exploration and exploitation, since each action can result in the change of reward and cost simultaneously, it is a multi-objective problem. \hl{Thus, in safe MARL settings, addressing the two balances mentioned earlier is imperative.} In competitive game settings, efficiently modeling an opponent's decisions with limited information and implementing safe actions must be ascertained. In cooperative game settings, searching for a policy to guarantee the whole team's reward monotonic improvement while adhering to individual agent constraints is essential. In mixed-game settings, the optimization of both local and overall rewards while executing safe actions needs to be determined.

\end{itemize}

\subsection{Possible Directions for Future Safe Reinforcement Learning Research}

\begin{itemize}
    \item   Possible direction one: Safe MARL with game theory. How to solve the above challenges by leveraging game theory is a primary direction, since we can consider different games for real-world applications in different game settings, such as cooperative games or competitive games; how to optimize safety in an extensive form game is also helpful in real-world applications. For instance, In a fencing competition, we need to determine how to ensure that two agents complete the task while ensuring agents' safety in the game process. 
    \item  Possible direction two: Safe RL with information theory. Information theory may be useful to handle the uncertainty reward signal and cost estimation, and efficiently address the problem of large-scale MARL environments. For example, we can use information coding theory to construct the representation of different agent actions or reward signals. 
    \item  Possible direction three: Safe MARL with human-brain theory and biology inspiration theory. We can draw some inspiration from the laws of biology to design safe MARL algorithms. For example, we learn the flying rules of geese, understand how geese form a certain formation, and ensure the safety of each geese.
    \item Possible direction four: Learn safe and diverse behaviors from human feedback, like ChatGPT\footnote{\url{https://openai.com/blog/chatgpt/}}. During human-robot interaction, robots need to pay more effort into learning safe and diverse behaviors from humans rather than discriminatory and illegal behaviors. We envision that a robot needs to adapt to different tasks and learn to satisfy different people's preferences after a robot can learn safe behaviors from one person. In such a scenario, a robot needs to learn how to conduct safe behaviors with multiple persons. For example, feedback from humans can be used as training data to boost the relationship between robots and humans, and help to guarantee exploration safety~\cite{frye2019parenting}. Moreover, behavior diversity is a critical factor to successful multi-agent systems~\cite{yang2021diverse}, and diverse behaviors can be leveraged to search for safe policies for safe robot learning.

\item Possible direction five: {Human-robot interactions.}
Learning from interactions with non-expert users is essential for long-term SRRL. For example, an earlier attempt \cite{hadfield2016cooperative} considers learning the user's reward signal from human-machine cooperation. The prerequisite that robots work with humans safely and efficiently is mutualism. \hl{Robots have to learn human preferences and comprehend human intentions and actions for successful collaboration with humans.} By understanding what humans want and anticipate, robots can better adapt to their environment and receive enhanced support from their human counterparts. This mutual understanding fosters an effective and harmonious human-robot collaboration. Human behavior modeling~\cite{deng2022policy} and realistic interactive modeling~\cite{zhu2022rita} can be the potential solutions for robots safely inheriting human preferences and learning more about humans.

\end{itemize}

\section{Conclusion}
\label{section:Conclusion-safe-DRL}
We carefully review safe RL methods from the past 20 years, attempt to answer the key safe RL question around the investigation of safety research with \textbf{``2H3W"} problems, and provide a clear clue for further safe RL research. Firstly, five critical safe RL problems are posed, and the model-based and model-free RL methods are analyzed in a unified safety framework. Secondly, the sample complexity and convergence of each method are investigated briefly. Thirdly, applications of safe RL are analyzed, for example, in the fields of autonomous driving and robotics. Fourthly, the benchmarks for safe RL communities are revealed, which may help RL take further toward real-world applications. Finally, the challenging problems that confront us during RL applications in safe RL domains are pointed out for future research. 

\section*{Acknowledgments}
 This work was partially supported by the European Union’s Horizon 2020 Framework Programme for Research
			and Innovation under the Specific Grant Agreement No. 945539 (Human Brain Project
			SGA3).	We would like to express our gratitude to Ming Jin, Yaodong Yang, and Hanna Krasowski for their invaluable suggestions during our safe RL research.

\normalem





\clearpage

\bibliography{main}{}
\bibliographystyle{plain}

\end{document}